\documentclass[preprint,3p,12pt]{elsarticle}

\usepackage{layout}
\usepackage{amsmath,amssymb,amsfonts}
\usepackage{algorithmic}
\usepackage{graphicx}
\usepackage{xcolor}
\usepackage{textcomp}
\usepackage{booktabs}
\usepackage{caption}
\usepackage{comment}
\usepackage{lineno}
\usepackage{float}
\usepackage{multirow}
\usepackage{makecell}
\usepackage{pdflscape}
\usepackage{microtype}
\usepackage{url}
\usepackage[acronym]{glossaries}
\usepackage{subcaption}
\usepackage{geometry}
\usepackage[table]{xcolor}
\usepackage{xcolor}
\usepackage[most]{tcolorbox}
\usepackage{placeins} 
\usepackage{float}   
\usepackage{xcolor}
\usepackage[most]{tcolorbox}
\usepackage{fontawesome5}

\definecolor{sysColor}{HTML}{66979C}
\definecolor{userColor}{HTML}{E27C78}
\definecolor{placeholderColor}{HTML}{C1440E}
\definecolor{myacc}{HTML}{B3A0CD}  
\definecolor{myf1}{HTML}{F3B9B8}   

\newcommand{\ph}[1]{\textcolor{placeholderColor}{\texttt{\{#1\}}}}

\tcbset{
  promptbox/.style={
    enhanced,
    boxrule=0.7pt,
    arc=2mm,
    left=8pt, right=8pt, top=6pt, bottom=6pt,
    fontupper=\small,
    fonttitle=\bfseries\sffamily,
    coltitle=white,
    title={#1},
  }
}

\usepackage{bbm}

\journal{Computers and Electronics in Agriculture}
\usepackage{lineno} 

\begin{document}

\begin{frontmatter}



\title{A Multi-Modal Generative Model for Tomato Disease Leaves Understanding}




\author[1]{Khang Nguyen Quoc}
\ead{khangnq@korea.ac.kr}
\affiliation[1]{organization={School of Electrical Engineering, Korea University},
            city={Seoul},
            postcode={02841},
            country={South Korea}}

\author[2]{Minh-Phuoc Tran}
\ead{minhphuoc19122005@gmail.com}

\author[2]{Gia-Han Truong}
\ead{truonggiahan0702ct@gmail.com}

\author[2]{Luyl-Da Quach\corref{cor1}}
\ead{luyldaquach@gmail.com}
\affiliation[2]{organization={Department of Software Engineering, FPT University},
            city={Cantho city},
            postcode={CT 90000},
            country={Vietnam}}

\cortext[cor1]{Corresponding author}

\begin{abstract}
Artificial intelligence for plant disease analysis has advanced from task-specific classifiers to multi-modal models capable of jointly interpreting visual and textual information. However, practical deployment in precision agriculture remains limited because most existing approaches treat disease understanding as isolated prediction tasks, failing to capture the complementary relationships among symptom recognition, severity assessment, and question-driven diagnostic reasoning. In tomato pathology, accurate interpretation of diseased leaves requires more than label prediction; it demands integrating visual symptoms with semantic context to support a comprehensive and explainable understanding. Here, we present SOLAR, a multimodal generative model that understands tomato disease spanning six question-answering tasks. SOLAR learns to align visual features with task-aware language representations by Fusion Expert module based on mixture-of-expert, enabling it to generate contextually relevant answers across diverse diagnostic tasks. By formulating tomato disease analysis as a generative Visual Question Answering (VQA) task, SOLAR provides a flexible framework that supports multi-task inference within a single model while improving performance and cross-task knowledge sharing. We evaluate SOLAR on $41,677$ images, including $216,209$ Question-Answering (QA) pairs to understand tomato leaf disease under both closed and open-ended QA settings. Experimental results show that SOLAR consistently outperforms state-of-the-art vision-only, vision-language, and task-specific models across all tasks, demonstrating superior accuracy, robustness, and multimodal reasoning. These findings highlight the potential of generative multimodal modeling as an effective direction for understanding of plant disease. The code for this study is available at \url{https://github.com/EnalisUs/SOLAR}.
\end{abstract}

\begin{graphicalabstract}
\begin{figure}[H]
 \centering
\includegraphics[width=1\textwidth]{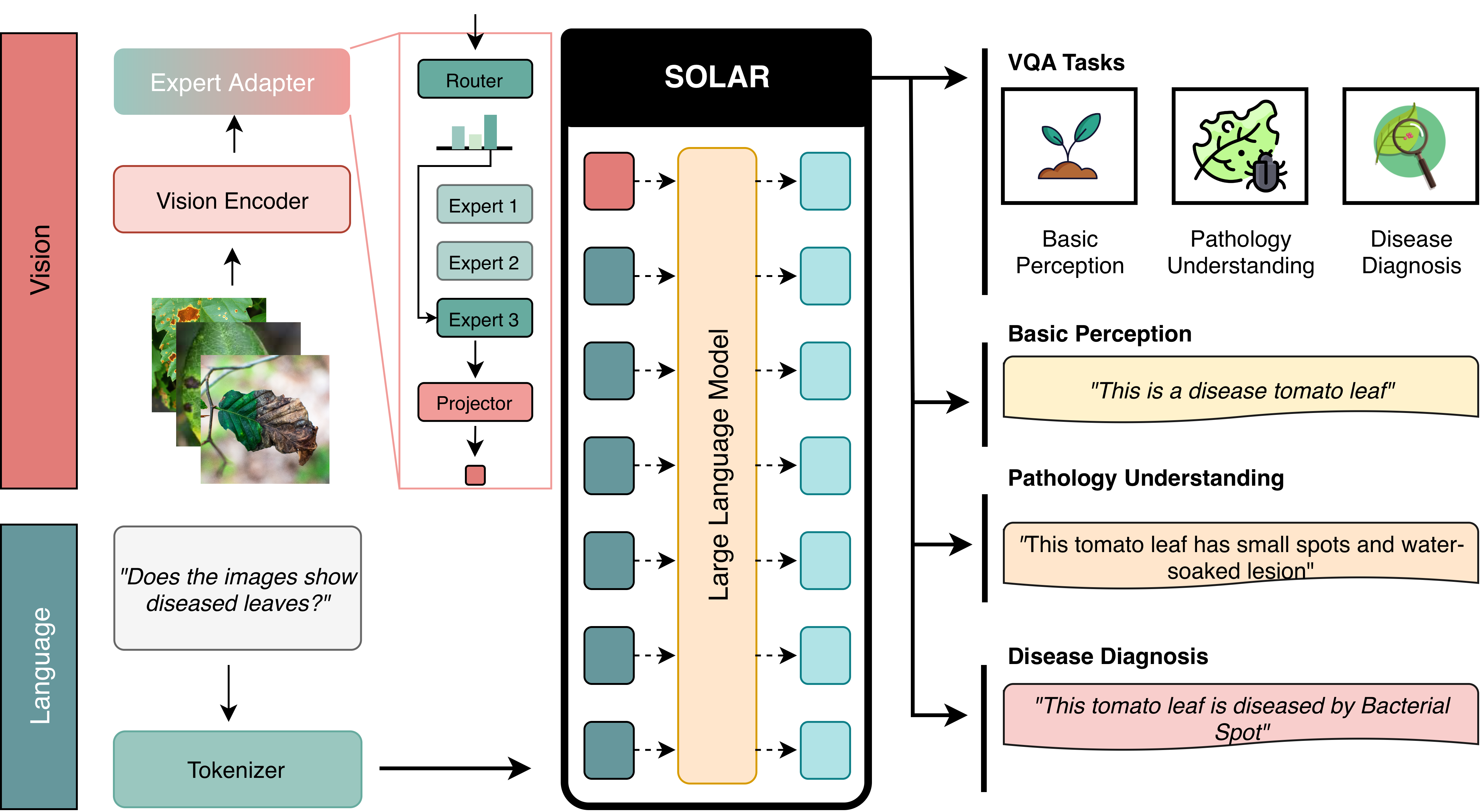}
    \caption*{\textbf{Overview of the SOLAR framework for tomato leaf disease diagnosis via Visual Question Answering (VQA).} The architecture comprises two input modalities: a Vision branch, in which input leaf images are processed through a Vision Encoder and subsequently refined by an Expert Fusion module consisting of a Router, multiple specialized Experts, and a Projector that maps visual features into the language embedding space; and a Language branch, in which natural language queries are tokenized and encoded as text embeddings. Both modalities are jointly processed by LLM to produce responses across three hierarchical diagnostic levels--Basic Perception, Pathology Understanding, and Disease Diagnosis---enabling structured, coarse-to-fine disease recognition across six VQA task types.}
\end{figure}
\end{graphicalabstract}

\begin{highlights}
    \item This study is the first to reformulate the tomato image classification problem as a generative task, specifically, Visual Question Answering.
    
    \item We propose SOLAR, a generative multimodal model for understanding tomato disease that mimics an agricultural expert's diagnostic workflow across multiple VQA tasks.
    
    \item We propose an Expert Fusion module to improve alignment between the visual feature space and multiple question embeddings.
    
    \item We evaluate SOLAR on a combined set of four distinct tomato disease datasets, which we formulate into six diagnostic tasks.
    
    \item Our model achieves state-of-the-art (SOTA) performance across all benchmarks, outperforming existing SOTA vision, vision-language, general, and domain-specific foundation models.
\end{highlights}


\end{frontmatter}


\section{Introduction}
\label{sec:intro}
According to statistics from the European Commission, tomato production reached over 8.2 million tons (2025). However, it showed a downward trend from 2024 (over 8.6 million tons), indicating that tomatoes are among the most important crops globally \cite{ECTomato}. This production tends to decrease due to strong seasonal and disease-related influences. To tackle this problem, many studies have focused on applying artificial intelligence (AI) and advances in deep learning (DL) to classify and name diseases, but have not yet optimized feature classification and the embedding space of large language models (LLMs) \cite{107054, 106997}. Therefore, in this problem, the research focuses on purely visual image classification in the form of VQA, with specific tasks, to serve as a foundation for subsequent studies that build datasets and models for each specific disease.

In AI applications for smart agriculture, research often focuses solely on disease identification, severity assessment, and symptom localization as separate tasks. In the disease identification task, research surveys leaf image classification approaches that leverage single-leaf characteristics using traditional machine learning and DL models. However, these can only answer "what disease is this leaf?" and fail to generalize to unseen content \cite{1356260}. Meanwhile, disease detection approaches also identify diseased areas using bounding boxes with Faster R-CNN \cite{3654570}, YOLO \cite{Nguyen2023}, and related models. While development has shifted from two-stage to single-stage models, blurred and irregular symptom borders affect severity assessment and make it difficult to identify small lesions, such as early-stage spots \cite{1041514}. Furthermore, semantic or instance segmentation techniques for symptom localization, severity measurement, and disease monitoring are expensive, require expert support, and do not provide sufficient information for disease identification \cite{1435016}. These challenges make it difficult to detect early abnormal symptoms in plants, compounded by semantic correlations between images and disease symptoms in the complex environment of smart agriculture and the level of farmers' understanding.

The problem of disease detection in tomato plants is no exception to this trend; currently, the focus is only on detecting, classifying, and isolating diseases, and there are common limitations in smart agriculture. Research focuses on detecting and recognizing small objects in complex real-world environments by moving from feature engineering to semantic representation to identify meaningful regions, making it easier to access new data in these scenarios, but it does not yet understand the causal relationships and disease progression \cite{ s13007}. Research using prescription data combined with stacking ensemble learning and an image semantics approach differs in that it shifts from pixel-level to context-level semantics, which is less dependent on images, but it does not directly address the full semantic understanding problem in computer vision, as it does not truly address the problem when combining multimodal fusion \cite{106997}.  The research shows improvements in data enhancement and optimization for disease, pest, and weed detection, with the potential to enhance the semantic representation of small, obscured objects by leveraging attention to increase semantic focus in YOLO. However, it does not yet understand the depth level \cite{125737}. This indicates that traditional classifier models act like a “black box,” only providing labels without explaining why, making it difficult for agricultural experts to trust them.

To address the limitations identified in previous studies, we introduce SOLAR, a generative multimodal model. SOLAR innovates by framing the problem as a VQA task informed by the diagnostic process of task-specific agricultural experts, and implemented across six integrated diagnosis tasks. At the same time, SOLAR improves the semantic relationship between images and text through proposed modules and evaluations based on different datasets. 

The main contributions of this work are summarized as follows:
\begin{itemize}
    \item This study is the first to reformulate the tomato image classification problem as a generative task, specifically, Visual Question Answering.
    
    \item We propose SOLAR, a generative multimodal model for understanding tomato disease, capable of mimicking an agricultural expert's diagnostic workflow across multiple VQA tasks.
    
    \item We propose an Expert Fusion module to improve alignment between the visual feature space and multiple question embeddings.
    
    \item We evaluate SOLAR on a combined set of five distinct tomato disease datasets, which we formulate into six diagnostic tasks.
    
    \item Our model achieves SOTA performance across all benchmarks, outperforming existing SOTA vision, vision-language, general, and domain-specific foundation models.
\end{itemize}

\section{Related Work}
\subsection{Deep Learning in Plant Disease Diagnosis}
The DL method based on Convolutional Neural Networks (CNNs) and Vision Transformers has made progress in classifying various plant disease characteristics. In particular, the use of basic DL models such as ResNet and EfficientNet, along with advances in parameter optimization and adjustment, has achieved high accuracy in identifying and classifying leaf diseases in maize, rice, wheat, and mango, with accuracies exceeding 90\%. However, it focuses only on classification and lacks explanatory indices for the results \cite{0131011, 3358333, 100476}. To improve model accuracy, many studies have combined advances in DL, such as transformer-based architectures and attention mechanisms, into YOLOv4 to delineate disease regions on leaf backgrounds and in the medium \cite{ s13007, 125737}. In other studies, utilizing the advancements of LLM models has made certain progress in interpreting results through visualizing features with color (Grab-CAM, Ablation-CAM), or using text to clarify features, as studies conducted on corn, tomatoes, and mangoes have achieved positive results, as it evaluates the relationship between features related to disease characteristics, physiology and contributes to expressing semantic information at the expert level but has not yet expressed the features or simulated the expert's diagnostic process \cite{101348,108155, Quach2024}. These issues challenge the ability to leverage advances in computer vision and natural language processing models to interpret and infer, thereby providing an expert perspective rather than analyzing images pixel-by-pixel or simply reading plain text.

\subsection{Vision-Language Models in Agriculture}
Addressing the limitations of traditional image classification, the VLM model is proposed to shift data from "image-only" to "image-contextual text" in agriculture. Various approaches are employed to enhance the effectiveness of Vision-Language Models (VLMs) using generative adversarial networks (GANs). In the applied approach, research focuses on proposing a pragmatic framework and exploring statistical language models, neural network models, and big language models, intending to integrate agricultural knowledge into VLMs using approaches such as instruction tuning, in-context learning, and generative retrieval architectures. However, these approaches also face difficulties in terminology and information discrepancies, such as distribution shift \cite{Li2025, Zhu2025}. Another approach involves building high-complexity standard datasets to test, evaluate, and identify the limits of existing VLMs across various tasks, such as AgroCoT (5 tasks) and AgEval (12 tasks), to assess in-context learning capabilities. However, these models are limited in accuracy, as evidenced by their complete reliance on GANs such as GPT, Gemini, and LlaVA, where performance decreases significantly when encountering types less common in the general dataset (AgEval) or when generating redundant text (AgroCoT) \cite{23253, 19617, leafnet}. A completely different approach, studies propose new models capable of generating in-depth VLMs for agriculture based on different approaches (based on LLaVA/Mipha to combine with SigLIP/CLIP ViT with LLM or based on contrastive learning to combine image and text encoders into a common embedding space), which creates effective plant disease identification models in various scenarios, but is limited in its ability to infer open conversational patterns and suffers from logical errors when relying on GAN models \cite{00555, scold}. Overall, research studies have shown a shift away from rigid classification labels toward detailed symptom analysis, data smoothing, expert support, and step-by-step explanations. However, this also presents challenges due to the need for expert knowledge, which can be overcome by adopting the flexible, natural-language-based approach of generative VQA.

\subsection{Generative VQA with interpretive capabilities}

The VQA model is an artificial intelligence system that takes images and natural-language questions as input, then provides accurate answers through three structured stages: a vision encoder, a language encoder, and a fusion \& decoder. This helps to answer specific, clear questions, demonstrating true semantic and visual inference capabilities \cite{Antol2015}. In the traditional VQA approach, the method involves selecting answers from a fixed list of labels or filling in the blanks through knowledge processing by directly embedding into network layers, such as co-attention and multimodal tucker fusion with free responses \cite{Zhao2024}, combining CNN with transformer models to extract pathological image features and link information with questions from farmers  \cite{Lan2023}, and combining VQA with federated learning models to update weights with recognition questions \cite{Nanavaty2024}. Its advantage is its ease of implementation on edge devices, but it cannot answer questions outside the predefined set of labels.

In particular, VQA relies on a large-scale/dialogue model to facilitate free-text generation, answer open-ended questions, and provide highly detailed technical solutions through LLM reasoning, such as mapping 5 core information aspects to 5 core information aspects, but it fails to overcome complex situational reasoning questions \cite{leafnet,leafmd}. PlantVillageVQA applies a language reconstruction system and diverse natural-language questions with expert review, but it faces significant accuracy challenges when real-world images have complex backgrounds \cite{17117}. Research using authentic grower dialogues, through tests and open-ended questions, helps capture farmers' writing styles and the real-world issues they care about \cite{agmmu}. Research using contrastive learning combined with lesion-aware learning and LLM models to conclude is limited to the sketch level\cite{TomaAD}. However, these models suffer from technical and computationally expensive hallucinations.

Overall, research shows that the VQA approach overcomes one-hot labels (e.g., forcing the model to respond as “disease” or “healthy”) by consolidating tasks through prompt changes, but still relies on question-embedding sensitivity and true interpretability. Hence, our study developed an expert fusion module to align the space and act as an expert to analyze the data and apply true pathology logic.

\section{Method}
\subsection{Task Formulation}

To transcend the limitations of conventional single-label image classification, this study reformulates tomato leaf disease recognition as a structured VQA problem organized around a coarse-to-fine diagnostic reasoning hierarchy. Formally, given an input image $I$ and a natural language question $\mathcal{Q}$, the model $\mathcal{F}$ is trained to generate a textual answer $\mathcal{A} = f(I, \mathcal{Q})$, where $f$ denotes the vision-language model. Under this formulation, we define six VQA tasks grouped into three progressive levels of diagnostic reasoning, encompassing basic visual perception, pathology understanding, and disease diagnosis, following:

\begin{itemize}
    \item \textbf{Level I: Basic Perception.} The first group establishes foundational visual awareness prior to any disease-level reasoning. \textit{Leaf Count (LC)} tasks require the model to identify and enumerate the leaves present in a given image, probing spatial and object-level perceptual capability. \textit{Healthy or Disease Classification (HDC)} tasks subsequently demand a coarse binary judgment of whether the observed leaf is healthy or exhibiting visible signs of pathological change, serving as an initial diagnostic screening gate.
    \item \textbf{Level II: Pathology Understanding.} Building upon perceptual foundations, the second group introduces a higher degree of biologically grounded reasoning. \textit{Pathogen Classification (PC)} tasks require the model to infer the nature of the causative agent---distinguishing among fungal, bacterial, and viral origins---based solely on observable visual evidence. \textit{Symptom Identification (SI)} tasks further demand that the model recognize and articulate specific visible disease manifestations, including lesions, chlorosis, necrosis, and irregular spotting patterns, thereby grounding its responses in directly observable leaf morphology.
    \item \textbf{Level III: Disease Diagnosis.}
    The third group demands precise, clinically meaningful outputs that reflect deep domain knowledge. \textit{Disease Classification (DC)} tasks require the model to assign the correct disease label from among 15 predefined tomato disease categories, necessitating fine-grained visual discrimination. \textit{Scientific Name Classification (SNC)} tasks present the most demanding challenge, requiring the model to identify the scientific nomenclature of the diagnosed condition---a degree of pathological specificity that extends well beyond colloquial naming conventions.
\end{itemize}
\vspace{0.5em}
 Collectively, these three hierarchical levels constitute a coherent diagnostic reasoning pipeline that progresses systematically from low-level visual perception, through biological pathology interpretation, to high-precision disease identification, mirroring the reasoning process of domain experts in plant pathology.
 
\subsection{SOLAR Architecture}
SOLAR is a multimodal generative model designed to emulate the diagnostic reasoning workflow of an agricultural expert in the context of tomato leaf disease understanding. Given a leaf image as input, SOLAR first encodes the visual content into a rich set of feature representations that capture the biological characteristics of the observed leaf. A structured two-stage reasoning pipeline then processes these features to produce accurate, contextually grounded responses. In the first stage, expert-aware visual encoding, the extracted visual features are refined by a Mixture-of-Experts (MoE)~\cite{moe} module that selectively activates relevant feature pathways conditioned on the input question. This mechanism allows the model to adaptively emphasize diagnostically relevant visual regions, such as lesion patterns, discoloration, or necrotic spots, depending on the nature of the query. The filtered visual features are subsequently fused with the question embedding to form robust multimodal representations, jointly conditioned on both the visual input and the question's linguistic context. In the second stage, answer generation, a large language model (LLM) receives the question embedding alongside the expert-aware visual features to generate a free-form textual response tailored to the current task. This design enables SOLAR to flexibly handle the full spectrum of diagnostic queries across the defined coarse-to-fine task hierarchy, from basic leaf perception to detailed pathological diagnosis. By aggregating evidence across all reasoning stages, SOLAR produces a comprehensive and interpretable understanding of tomato leaf disease, bridging the gap between visual observation and expert-level diagnostic judgment. 

\begin{figure}[t]
    \centering
    \includegraphics[width=1\linewidth]{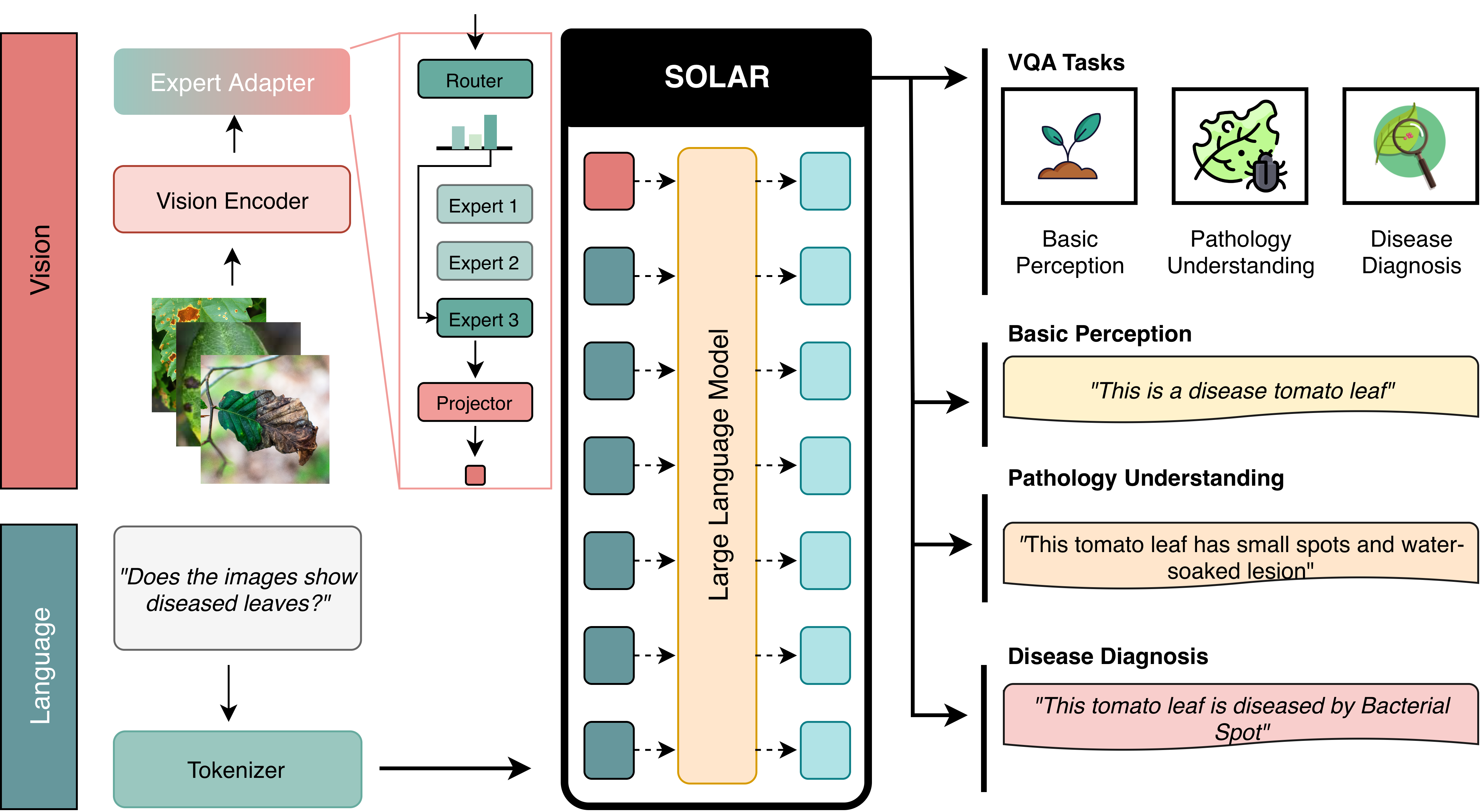}
   \caption{\textbf{Overview of the SOLAR framework for tomato leaf disease diagnosis via VQA.} The architecture comprises two input modalities: a Vision branch, in which input leaf images are processed through a Vision Encoder and subsequently refined by an Expert Fusion module consisting of a Router, multiple specialized Experts, and a Projector that maps visual features into the language embedding space; and a Language branch, in which natural language queries are tokenized and encoded as text embeddings. Both modalities are jointly processed by LLM to produce responses across three hierarchical diagnostic levels--Basic Perception, Pathology Understanding, and Disease Diagnosis--enabling structured, coarse-to-fine disease recognition across six VQA task types.}
    \label{fig:solar}
\end{figure}

The framework consists of three primary components: 1) an image encoder $f_{img}$, 2) an Expert Fusion module $\mathcal{E}$, and 3) a LLM decoder $\mathcal{T}$. The overall architecture is illustrated in Figure~\ref{fig:solar}.

Given a image $\mathcal{I}$, we extract visual embedding $\mathcal{V}$ by the visual encoder $f_{img}$, where $\mathcal{V} \in \mathbb{R}^{D}$. We utilize SCOLD~\cite{scold}, a vision-language foundation model pretrained on over 157K plant image-caption pairs, as the visual encoder. Since the dimension of the image encoder is different from the text embeddings, we use a projection layer to map $\mathcal{P}$ into the same space as the embeddings $\mathcal{V} = \mathcal{P}(\mathcal{V})$.

The question-aware Expert Fusion adapter $\mathcal{E}$ bridges the visual embedding space and the LLM embedding space. Unlike traditional static projectors (e.g., LLaVA-OV~\cite{llava-ov}), which directly map visual embedding to LLM space, $\mathcal{E}$ explicitly incorporates the question embedding $\mathcal{Q}$ as an expert filter, obtained by encoding the question with a text encoder, to produce task-relevant experts. To obtain $\mathcal{Q}$, we leverage LFM-VL 2.5~\cite{lfm2}, a model designed for on-device deployment. It builds on the LFM2 architecture with extended pre-training and reinforcement learning. Building on this encoder-decoder model, we ensure that the semantic feature space of the encoded question is well-designed for the generative space.

With the features extracted from the question and the image, the Expert Adapter module  $\mathcal{E}$ first refines the visual features via a Sparse-MoE to filter the important information relevant to each specific question.A remarkable aspect of our model is that the gating network takes the question $\mathcal{Q}$ as input to determine which experts to activate. Consequently, the same image paired with different questions will trigger different experts, it make visual features for each question type. 

\begin{equation}
    V_e = \mathcal{V} + \sum_{i=1}^{N} R_i(\mathcal{Q}) \cdot E_i(\mathcal{V})
\end{equation}
With $N$ is the number of experts, $E_i$ denotes the $i$-th expert layer, and $R_i$ is the $i$-th routing weight computed from the question $\mathcal{Q}$ from the gating network. Specifically, only the $K$ experts with the highest routing weights are evaluated.

Subsequently, the expert-aware feature fuses with the question feature to compute the final visual token with flexibility for each question. 
\begin{equation}
    x_v = \text{Dropout}\Big(\text{GELU}\big(\text{LayerNorm}(W_f \cdot [\mathcal{Q} ||\, V_e])\big)\Big)
\end{equation}
Where $[\cdot||\cdot]$ represents the concatenation operation and $W_f$ are learnable projection weights.

The LLM decoder, denoted $\mathcal{T}$, executes the VQA task by formulating diagnostic responses in natural language. In contrast to conventional classification models that only predict classes, $\mathcal{T}$ accepts the specific visual token $X_v$ alongside the question tokens $X_q$. After concatenating these inputs, it generates the response autoregressively, predicting each subsequent token conditioned on both the visual features and the textual context.

\subsection{Question-Answer Dataset Construction}

We develop the dataset in a question–answer prompt format, where each image is paired with a corresponding question and answer. For each type of disease, a sample consists of a leaf image, a diagnostic question, and its associated ground-truth label. These pairs are represented as $ {({X}_{Q,i}, {X}_{A,i})}_{i=1}^N$ where ${X}_{Q,i}$ denotes the input question related to the image, and
${X}_{A,i}$ denotes the corresponding ground-truth output. This formulation enables the model to learn task-specific diagnostic knowledge, focus on visual information related to the question, follow a fixed output format, and generate correct labels.

We construct the input sequence by concatenating the question ${X}_{Q}$, the string  \text{`` Answer: ''}, and the ground-truth answer ${X}_{A}$ to make a unified prompt:

\begin{equation}
X_{\text{p}} = X_{Q} + \text{`` Answer: ''} + {X}_{A}
\end{equation}

\subsection{SOLAR Training and Inference}

The full SOLAR training target is presented below:

\begin{equation}
{L} = {L}_{\text{TASK}} + \lambda_1 {L}_{\text{AUX}} + \lambda_2 {L}_{\text{DIV}}
\end{equation}
where ${L}_{\text{TASK}}$ is the loss of autoregressive language modeling, ${L}_{\text{AUX}}$ is the loss of load balance that regularizes expert utilization, and ${L}_{\text{DIV}}$ is the loss of expert diversity that encourages specialized representations. $\lambda_1$ and $\lambda_2$ denote weighting coefficients.

The primary loss ${L}_{\text{TASK}}$ maximizes the probability of the target answer ${X}_{A}$ consisting of M tokens conditioned on the input prompt ${X}_{\text{p}}$, which is given by:
\begin{equation}
{L}_{\text{TASK}} = -\sum_{i=1}^{M} \log p_{\theta}(x_i \mid {X}_p, x_{<i})
\end{equation}
where, $\theta$ denotes the trainable parameters, $x_i$ is the token at position $i$ in the answer, and $x_{<i}$ refers to all tokens that appear before position $i$.

In the Mixture-of-Experts architecture, a common issue is expert collapse, where the gating network repeatedly routes most inputs to only a few experts, while the remaining experts are barely used. To prevent this degenerate behavior, we apply the auxiliary load-balancing loss $L_{\text{AUX}}$ following the formulation of the Switch Transformer \cite{210103961F}:

\begin{equation}
\mathcal{L}_{\text{AUX}} = N \cdot \sum_{i=1}^{N} f_i \cdot P_i
\end{equation}
where $f_i$ is the fraction of samples in batch $B$ sent to expert $i$.

\begin{equation}
f_i = \frac{1}{B} \sum_{b \in \mathcal{B}} \mathbbm{1}\{\text{expert } i \in \text{TopK}(b)\}
\end{equation}
and $P\textsubscript{i}$ is the fraction of the probability that the router is assigned to the expert $i$.

\begin{equation}
P_i = \frac{1}{B} \sum_{b \in \mathcal{B}} p_i(b)
\end{equation}
with $p_i(b)$ being the probability of gating assigned to expert $i$ in the sample $b$ in the batch. This loss reaches its minimum when all experts are utilized uniformly, $f_i = P_i = \frac{1}{N} $ for all $i$.

Although load balancing helps distribute inputs relatively evenly between experts, it does not ensure that different experts learn different features. We observe that, without explicit regularization, experts may learn overlapping features, which reduces the effectiveness of the model. To address this issue, based on the Mean Squared Overlap \cite{kim2026}, we propose an expert diversification loss $L_{\text{DIV}}$ that penalizes high cosine similarity between the weight matrices of different experts. Let $W_i \in \mathbb{R}^{d_h \times d_v}$ denote the weight matrix of the first linear layer in expert $i$, where $d_h$ is the hidden dimension and $d_v$ is the visual dimension. We then flatten and normalize each weight matrix in L2 into a unit vector $\hat{w}_i$, and define:

\begin{equation}
\mathcal{L}_{\text{DIV}} = \frac{1}{N(N-1)} \sum_{i<j} (\hat{\mathbf{w}}_i^\top \hat{\mathbf{w}}_j)^2
\end{equation}
This loss helps experts specialize better, ensuring that they extract different aspects of visual features.

In the training phase, we fine-tune the model on the TomaMMU dataset~\cite{tomammu}, which includes more than 27,000 unique images and 94,932 QA pairs. These QA pairs are divided into six specific groups, covering three levels from simple to complex: Basic Perception, Pathology Understanding, and Disease Diagnosis. This design helps the model to understand the question more effectively. This allows us to pay attention to the right visual evidence in the image and make the best diagnoses. Furthermore, in the LLM layer, we apply LoRA\cite{lora} to optimize performance, and the learned weights are merged directly into the original LLM weights. Specifically, for a frozen weight matrix $W_0 \in \mathbb{R}^{d \times k}$ of the attention layer, the weight update $\Delta W$ is decomposed into the product of two low-rank matrices $A \in \mathbb{R}^{r \times k}$ and $B \in \mathbb{R}^{d \times r}$ with rank $r \ll \min(d, k)$:

\begin{equation}
W = W_0 + \Delta W = W_0 + \frac{\alpha}{r} B A
\end{equation}

The forward pass calculation for a given hidden activation vector $h$ through the adapter-equipped attention layer is formulated as:

\begin{equation}
h' = W_0 h + \frac{\alpha}{r} B A h
\end{equation}

where $\alpha$ is a constant scaling hyperparameter. Throughout training, the original pre-trained weight matrix $W_0$ remains completely frozen, and gradients are only computed and backpropagated to update the low-rank adapter parameters $A$ and $B$, which significantly minimizes memory consumption and training overhead. After training phase, we evaluate the model on the validation set at the end of each epoch. We select the model weights from the epoch that achieves the highest validation accuracy as the final checkpoint for testing phases.

In the inference phases, the model uses the components described in the Architecture section. Given the input image $I$ and the question $\mathcal{Q}$, the question-conditioned visual token $x_v$ is computed using the Expert Fusion adapter $\mathcal{E}$:
\begin{equation}
x_v = \mathcal{E}(\mathcal{V}, \mathcal{Q})
\end{equation}
Then, the target diagnostic answer $Y = (y_1, y_2, \dots, y_L)$ is generated in autoregressive form based on the visual token $x_v$ and the question tokens, based on the following conditional probability equation
\begin{equation}
P(Y \mid x_v, \mathcal{Q}) = \prod_{t=1}^{L} P(y_t \mid y_{<t}, x_v, \mathcal{Q})
\end{equation}
where $y_{<t}$ denotes the sequence of answer tokens generated prior to step $t$. 

\section{Experimental Setup}
\label{sec:experimentalsetup}
\subsection{Datasets}

\begin{figure}[H]
    \centering    \includegraphics[width=\linewidth]{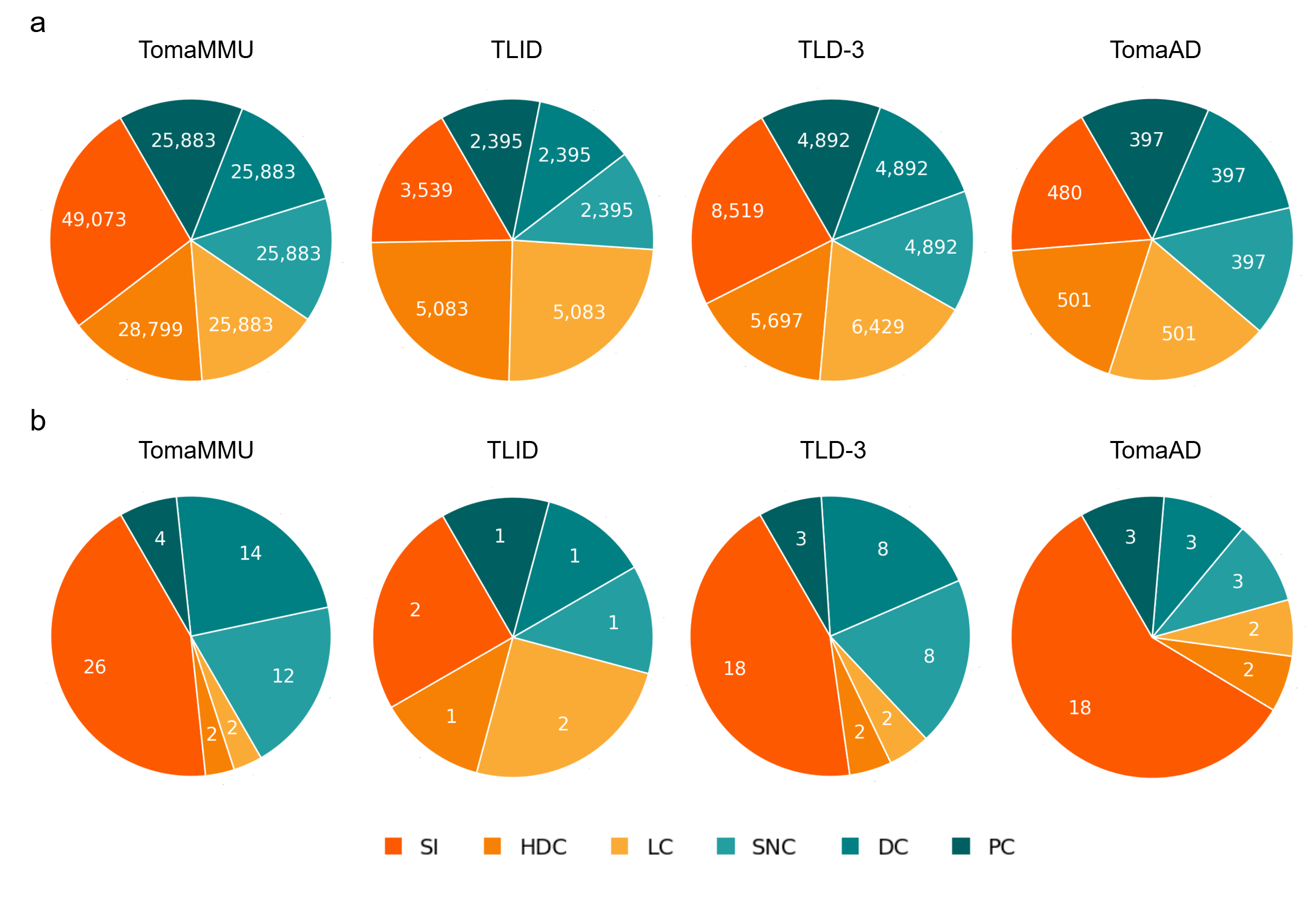}
    \caption{\textbf{Distribution of QA samples(a) and Number of classes(b) across the six VQA tasks for the five datasets}. In TomaMMU, number of class SNC and DC are differ because two DC classes, \textit{Mite} and \textit{Spider mites Two-spotted Spider Mite}, are the same scientific name \textit{Tetranychus urticae}. In the same way two DC classes, \textit{Mosaic Virus} and \textit{Virosis}, are same \textit{Tomato mosaic virus (ToMV)} in SNC. }
    \label{fig:OverViewDataset}
\end{figure}

We evaluated SOLAR in our internal and three external tomato  disease datasets, including TomaMMU~\cite{tomammu}, TLID \cite{TLID}, TLD-3 \cite{TLD-3} and TomaAD \cite{TomaAD}. 
All input images were resized to a fixed resolution of $224\times224$ pixels prior to model training and testing. The overview of the datasets is shown in Figure~\ref{fig:OverViewDataset}.

In this study, we utilized our TomaMMU dataset \cite{tomammu}. The images were filtered according to disease, symptom, leaves, pathogen, scientific name, and image descriptions. Lastly, the research team classified the images into disease types. The Dataset consists of more than 21,000 images of 15 diseases with 6 tasks, 158,214 QA pairs, and covering $60$ classes. TomaMMU was divided into 6:2:2, which corresponds to 94,932 QA pairs in the training phase, 31,641 QA pairs in the validation phase, and 31,641 QA pairs in the testing phase. We ensure that the images belonging to each disease category are proportionally allocated within the dataset, despite the inherently unbalanced distribution between classes.  The statistics of these dataset classes are presented in  
Figure~\ref{fig:statistic_image}.

The three external test datasets are reserved solely for the testing phase. Before entering this phase, these images are filtered and preprocessed following TomaMMU~\cite{tomammu} procedure.


TLID \cite{TLID} was constructed through multiple field surveys in various greenhouse farms, allowing the documentation of different diseases at various stages of development. This dataset comprises seven distinct classes categorized into 4 types of disease , two co-infection scenarios, and one healthy state. However, to align with our model, we utilized only 2 disease with 25,973 QA pairs.

TLD-3 \cite{TLD-3} is a dataset specifically focused on images of tomato leaf collected under various environmental conditions. This dataset comprises 10 classes with 49,213 QA pairs representing various leaf diseases. 

TomaAD \cite{TomaAD} is a dataset consisting of 502 tomato leaf images categorized into 4 diseases with 3,174 QA pairs. Data were collected from a commercial greenhouse in the arid climate of Abu Dhabi, United Arab Emirates. Disease labels were assigned based on visual observation and validated by agricultural experts on-site. In particular, these datasets reflect real-world imaging conditions, including illumination variance, leaf orientation, and partial occlusions.

\subsection{Comparison Setup}
\label{ComparisonSetup}
\noindent \textbf{State-of-the-art competing models.} To benchmark SOLAR, we evaluated competing SOTA architectures across three distinct paradigms. First, to assess out-of-the-box generalization, we tested a cohort of FMs and domain-specific VLMs without task-specific fine-tuning, comprising InternVL3 (1B and 2B)~\cite{internvl3}, Qwen2.5-3B~\cite{qwen25}, Qwen3-2B~\cite{qwen3}, LLaVA-OV~\cite{llava-ov}, LFM-VL 2.5~\cite{lfm2}, BioCLIP~\cite{bioclip}, SigLIP2~\cite{siglip2}, and SCOLD~\cite{scold}. Second, to establish standard image classification baselines, we fine-tuned representative vision-only architectures, comprising ResNet50~\cite{resnet50}, EfficientNetV2S~\cite{efficientnetv2}, DenseNet121~\cite{densenet}, and InceptionResNetV2~\cite{inception}. Lastly, we compared SOLAR directly against fine-tuned VLMs—namely CLIP~\cite{clip}, BioCLIP~\cite{bioclip}, SigLIP2~\cite{siglip2}, AgriCLIP~\cite{agriclip}, SCOLD~\cite{scold}, and LLaVA 1.5~\cite{llava-ov}—trained under identical conditions. All evaluations were conducted within a strictly consistent experimental framework to ensure fair comparison.

\noindent \textbf{Evaluation Metrics.} We evaluate model performance using five complementary metrics. Accuracy (Acc) measures the proportion of correct predictions over the total number of test samples, reflecting the general reliability of the classification. F1-Score (F1) combines precision and recall into a single balanced measure to account for class-level performance. For generation quality, we employ $ROUGE_L$~\cite{rouge} complements these by measuring the longest common subsequence between generated and reference texts, thereby reflecting fluency and structural coherence. To evaluate the response quality of open-ended VQA output, we use $GPT$ that the LLM-as-a-judge \cite{judge} method with Gemini Flash 2.5 \cite{gemini25}. This evaluator compares the generated response with ground truth and rates it on a five-point scale based on  semantic alignment, correctness, and completeness. Under this scheme, a score of 1 indicates a completely incorrect or irrelevant answer; 2 represents a partially relevant response with major inaccuracies; 3 denotes a generally correct output that lacks specific details; 4 signifies a fully correct and semantically equivalent answer; and 5 corresponds to an exceptionally precise, clear diagnosis that demonstrates a comprehensive understanding of both visual and textual contexts. The model is instructed to output only a single integer to facilitate structured data parsing. Together, these metrics provide a comprehensive assessment of both global correctness and per-class diagnostic effectiveness for classification tasks, as well as lexical fidelity and generation quality for generative task types. The detail prompt in the Table \ref{fig:llm_judge_prompt}. 

\textbf{Statistical Methods} We compare between the best model and the second best by using a two-sided Wilcoxon signed-rank test. Performance scores are presented as the mean with $95\%$ confidence intervals (CIs) estimated via bootstrapping ($n = 1,000$ replicates).

\subsection{Experimental Setup}
In this section, we describe the hardware and hyperparameter settings and evaluation method as below.
used in our experiments.

\noindent \textbf{Hardware Platform}: All experiments are performed on a local workstation with the following hardware and software specifications: 
CPU: Intel(R) Xeon(R) CPU E5-2686 v4 @ 2.30GHz, GPU: NVIDIA GeForce RTX 4060 Ti 16 GB, Memory: 128 GB RAM, System: Windows

\noindent  \textbf{Hyperparameter Settings}: For training, we run the model for 20 epochs using the AdamW optimizer with a learning rate of $1 \times 10^{-4}$ and a weight decay of 0.01. The batch size is set to 32. For LoRA tuning, we apply the adapters to the query, key, value and output layers ($W_q, W_k, W_v, W_o$) of the language model with a rank $r = 16$, alpha $\alpha = 32$, and a dropout rate of 0.1. For the MoE layer, the model has $N = 4$ experts and uses top-$K = 2$ routing. Each expert has a hidden size of 4096. The weights for the loss balance load and the loss of diversity are set to $\lambda_1 = 0.01$ and $\lambda_2 = 0.001$. 

\noindent  \textbf{Foundation Models}: Both general-purpose and domain-specific vision-language foundation models are evaluated in a zero-shot multiple-choice setup. The prompt presents the query with four options. For single-label tasks (LC, HDC, PC, DC, SNC), each option corresponds to a distinct category. For the multi-label Symptom Identification (SI) task, since a single image may have multiple distinct symptoms, we ensure that each option contains only a single symptom associated with that image.

\noindent  \textbf{Vision Models}: We add a classification layer to each model for each question type. The number of classification heads corresponds to the number of classes for each type of question.

\noindent  \textbf{Fine-tuned VLMs}: Contrastive vision-language models are fine-tuned on the training set. During inference, option text prompts corresponding to all class names within a task are encoded through the text encoder. For a given input image, the cosine similarity between the image embedding and all option text embeddings is computed. The option that yields the maximal similarity score is selected as the predicted label.

Unlike traditional image classification models, SOLAR directly generates natural language answers. For single-label tasks (LC, HDC, PC, DC, SNC), predicted labels are extracted from the generated text via exact keyword matching against the ground truth. For the multi-label Symptom Identification (SI) task, predicted symptom terms are parsed into a discrete set and evaluated against the ground-truth symptom set.

\section{Experimental Result}

\subsection{Closed Visual Question Answering Performance}
\label{sec:result}

To evaluate the effectiveness of SOLAR, we conduct comprehensive comparative experiments against three categories of models on our internal dataset and three external datasets. 

\begin{figure}[H]
    \centering
    \includegraphics[width=\linewidth]{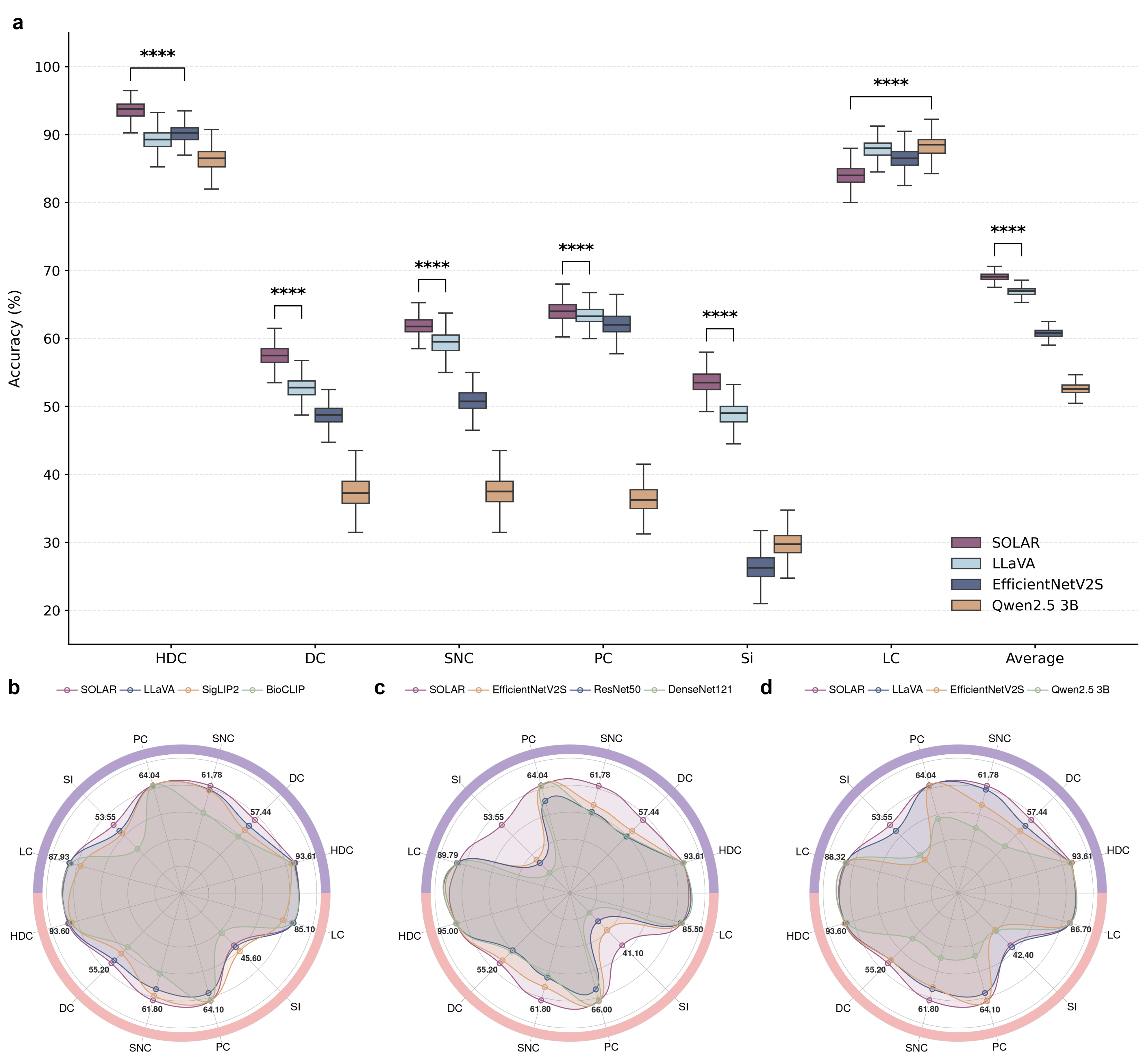}
    \caption{\textbf{Performance comparison across the six VQA tasks and the overall average and performance comparison of the top-performing models in each category across the six tasks}. \textbf{a,} The box plots show the distribution of accuracy for SOLAR with the best models in each category(LLaVA, EfficientNetV2S, and Qwen2.5 3B) across the six tasks (HDC, DC, SNC, PC, SI and LC). \textbf{b,} Comparison of SOLAR with the fine-tuned VLMs achieving the highest average performance(LLaVA,
    SigLIP2, BioCLIP).
    \textbf{c,} Comparison of SOLAR with the CNN architectures achieving the highest average performance(EfficientNetV2S, ResNet50, DenseNet121). \textbf{d,} Comparison of the SOLAR model with the
    best-performing models in each evaluation domain (LLaVA, EfficientNetV2S, Qwen2.5 3B). Error bar represented as the mean with 95\% CIs estimated using the bootstrap method ($n=1,000$ replicates). The two-sided Wilcoxon signed-rank test was used to assess the statistical differences between SOLAR and the second-best model. $^{**}P\leq0.01$, $^{***}P\leq0.001$, $^{****}P\leq0.0001$. (\textcolor{myacc}{purple} indicates Acc, and \textcolor{myf1}{pink} denotes F1)}
    \label{fig:box_plot}
\end{figure}

\noindent \textbf{SOLAR demonstrated superior performance on all VQA tasks.} On average, SOLAR consistently outperformed all SOTA models (Figure~\ref{fig:box_plot}a). In particular, SOLAR significantly outperformed all competing models in basic perception tasks such as HDC ($93.7\%$, $95\%$ CI: $91.3\%$--$96.0\%$), pathology understanding tasks such as PC ($64.0\%$, $95\%$ CI: $61.5\%$--$66.5\%$) and SI ($53.6\%$, $95\%$ CI: $50.0\%$--$57.0\%$), and disease diagnosis tasks including DC ($57.4\%$, $95\%$ CI: $54.5\%$--$60.3\%$) and SNC ($61.8\%$, $95\%$ CI: $59.0\%$--$64.8\%$). SOLAR achieved an average increase of $+2.15$\% ($P\leq0.0001$), with respective performance differences of $+3.49$\% (HDC), $+0.72$\% (PC), $+4.63$\% (SI), $+4.58$\% (DC), $+2.37$\% (SNC), and $-4.36$\% (LC) compared with the second-best model ($P\leq0.0001$). In addition, we compare our model with the top-3 models from each model category across all six tasks (Figure \ref{fig:box_plot}b-d, and Tables~\ref{tab:detailed_tomammu}-\ref{tab:detailed_tomaad}). Specifically, when comparing with the top-3 fine-tuned VLMs, SOLAR outperforms on almost all tasks with $+16.79\%$ and $+0.196$ (LLaVA), $+37.68\%$ and $+0.263$ (SigLIP2), and $+60.70\%$ and $+0.513$ (BioCLIP), but underperforms on LC with a gap of $-3.95\%$ and $-0.011$ (LLaVA). Similarly, SOLAR also outperforms vision-only models in almost all tasks with $+52.37\%$ and $+0.208$ (EfficientNetV2S), $+69.34\%$ and $+0.496$ (ResNet50), and $+70.13\%$ and $+0.503$ (DenseNet121), but slightly lower on LC with gaps of $-2.39\%$ (EfficientNetV2S), $-5.81\%$ (ResNet50), and $-2.28\%$ (DenseNet121). Specifically, with top-1 representatives from each category, SOLAR also exceeded other models in almost all tasks, but still underperformed on LC with $-3.95\%$ and $-0.011$ (LLaVA) and $-4.34\%$ and $-0.027$ (Qwen2.5 3B).

\begin{figure}[H]
    \centering
    \includegraphics[width=\linewidth]{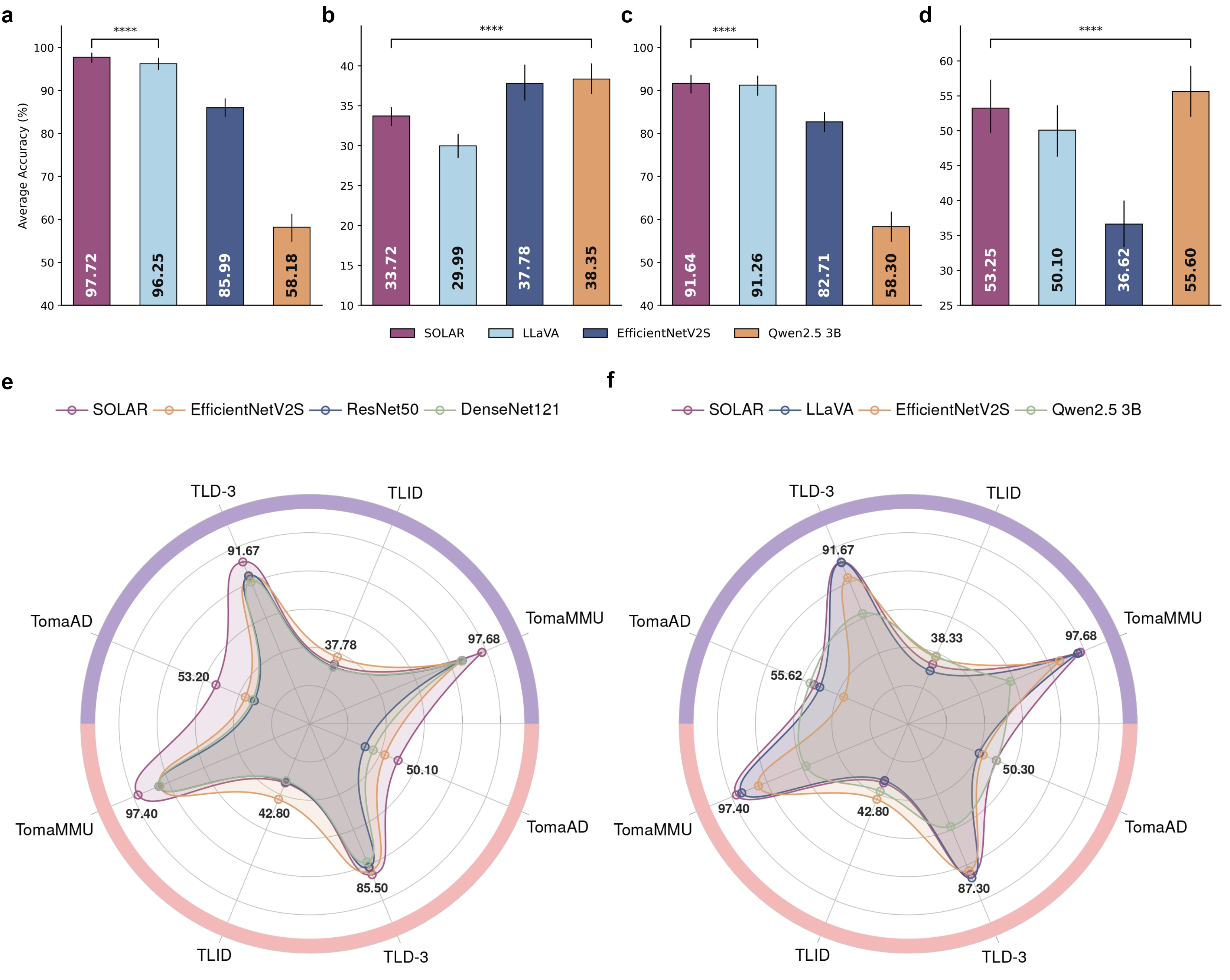}
    \caption{\textbf{Average accuracy comparison of SOLAR and baseline models across the four evaluation datasets including (a)TomaMMU, (b) TLID (c) TLD-3 and (d) TomaAD datasets}. The bars represent the mean accuracy of SOLAR with the best models in each category (LLaVA, EfficientNetV2S, and Qwen2.5 3B). Performance comparison on tomato disease VQA datasets. \textbf{e,} Comparison of SOLAR with the
    fine-tuned VLMs achieving the highest average performance(LLaVA, SigLIP2, BioCLIP) \textbf{f,} Comparison of
    SOLAR with the CNN architectures achieving the highest average performance (EfficientNetV2S, ResNet50,
    DenseNet121). Error bar represented as the mean with 95\% CIs estimated using the bootstrap method ($n=1,000$ replicates). The two-sided Wilcoxon signed-rank test was used to assess the statistical differences between SOLAR and the second-best model. $^{**}P\leq0.01$, $^{***}P\leq0.001$, $^{****}P\leq0.0001$. (\textcolor{myacc}{purple} indicates Acc, and \textcolor{myf1}{pink} denotes F1)}
    \label{fig:bootstrap}
\end{figure}

\noindent \textbf{SOLAR significantly outperformed baselines on internal and external datasets.} Figure~\ref{fig:bootstrap} demonstrated that SOLAR consistently outperformed all competing models, leading with the highest average accuracy of $69.08\%$ ($95\%$ CI: $67.88\%$--$70.21\%$). This represents an accuracy improvement of $+8.32\%$ over EfficientNetV2S, $+10.58\%$ over ResNet50, and $+10.59\%$ over SigLIP2 (Table~\ref{tab:Overall}). Specifically, on our internal dataset and TLD-3 dataset, SOLAR maintained superior performance, achieving accuracies of $97.68\%$ ($95\%$ CI: $96.50\%$--$98.83\%$) and $91.67\%$ ($95\%$ CI: $89.33\%$--$93.67\%$), respectively. In these specific evaluations, SOLAR outperformed LLaVA by $+1.78\%$, EfficientNet by $+10.24\%$, and Qwen2.5 3B by $+10.45\%$ (Figure~\ref{fig:bootstrap}a-d). Specifically, when comparing with top-3 fine-tuned VLMs(Figure~\ref{fig:bootstrap}e), SOLAR covers almost all other VLMs $+8.55\%$ and $+0.123$ (LLaVA), $+42.35\%$ and $+0.348$ (SigLIP2), and
$+44.72\%$ and $+0.385$ (BioCLIP). Meanwhile, SOLAR still outperformed vision models in almost dataset (Figure~\ref{fig:bootstrap}f)  such as $+37.38\%$ and $+0.222$ (EfficientNetV2S), $+40.91\%$ and $+0.349$ (ResNet50), and $+43.52\%$ and $+0.334$  (DenseNet121), but underperformed in TLID, with the gaps are $-4.08\%$ and $-0.093$ (EfficientNetV2S).  (See detail in Tables~\ref{tab:detailed_tomammu}-\ref{tab:detailed_tomaad}).

\noindent \textbf{SOLAR demonstrates strong generative capabilities for agricultural VQA.} Because agricultural VQA requires not only accurate classification but also informative context, we further evaluated the models' capacities to generate free-form responses. As shown on Table~\ref{tab:generative}, on average, it outperformed Qwen3-2B by $+0.024$ ($\text{ROUGE}_\text{L}$) and $+0.759$ (GPT), with consistent improvements over InternVL3-2B ($+0.028$, $+0.674$) and InternVL3-1B ($+0.034$, $+1.013$). Specifically, on our internal dataset, SOLAR achieved the highest performance ($\text{ROUGE}_\text{L}$: $0.995$; GPT: $4.741$), outperforming the strongest competitors, including LFM-VL 2.5 (by $+0.06$ and $+1.000$, respectively), Qwen3-2B ($+0.063$, $+1.312$), and InternVL3-2B ($+0.024$, $+1.371$). Conversely, on the TLID dataset, SOLAR trailed the two leading models, yielding a $\text{ROUGE}_\text{L}$ of $0.894$ and a GPT score of $2.649$. This represented a deficit of $0.036$ ($\text{ROUGE}_\text{L}$) and $0.462$ (GPT) compared to Qwen3-2B, alongside gaps of $0.025$ and $0.67$ relative to InternVL3-2B. However, SOLAR regained its lead on the TLD-3 and TomaAD datasets, scoring $0.973$ and $0.920$ in $\text{ROUGE}_\text{L}$, and $4.549$ and $3.334$ in GPT, respectively. Across these datasets, SOLAR maintained substantial margins over Qwen3-2B (by $+0.068$ $\text{ROUGE}_\text{L}$ and $+2.186$ GPT), InternVL3-2B ($+0.072$ and $1.997$), and InternVL3-1B ($+0.079$ and $+2.431$). Overall, SOLAR achieved the highest average scores across all evaluated datasets ($\text{ROUGE}_\text{L}$: $0.946$; GPT: $3.818$). (See detail in Tables~\ref{tab:detailed_gen_tomammu}-\ref{tab:detailed_gen_tomaad})

\begin{table}[H]
\caption{Generative evaluation of SOLAR and baseline models across four tomato disease VQA datasets, measured by $ROUGE_L$ and $GPT$ judged by Gemini Flash 2.5.}
\label{tab:generative}
\centering
\resizebox{\textwidth}{!}{
\begin{tabular}{lcccccccccc}
\toprule
\multirow{2}{*}{\textbf{Model}} & \multicolumn{2}{c}{\textbf{TomaMMU}} & \multicolumn{2}{c}{\textbf{TLID}} & \multicolumn{2}{c}{\textbf{TLD-3}} & \multicolumn{2}{c}{\textbf{TomaAD}} & \multicolumn{2}{c}{\textbf{Average}} \\ \cmidrule(lr){2-3} \cmidrule(lr){4-5} \cmidrule(lr){6-7} \cmidrule(lr){8-9} \cmidrule(lr){10-11}
 & $ROUGE_L$ & $GPT$ & $ROUGE_L$ & $GPT$ & $ROUGE_L$ & $GPT$ & $ROUGE_L$ & $GPT$ & $ROUGE_L$ & $GPT$ \\ 
\midrule
InternVL3 1B~\cite{internvl3}  & 0.926 & 3.311 & 0.908 & 2.459 & 0.926 & 3.311 & 0.888 & 2.141 & 0.912 & 2.805 \\
InternVL3 2B~\cite{internvl3}  & 0.931 & 3.370 & 0.919 & \textbf{3.319} & 0.931 & 3.370 & 0.890 & 2.516 & 0.918 & 3.144 \\
Qwen2.5 3B~\cite{qwen25}       & 0.931 & 3.382 & 0.909 & 2.750 & 0.931 & 3.382 & 0.882 & 2.252 & 0.913 & 2.941 \\
Qwen3 2B~\cite{qwen3}          & 0.932 & 3.429 & \textbf{0.930} & 3.111 & 0.932 & 3.429 & 0.893 & 2.268 & 0.922 & 3.059 \\
LlaVA-OV~\cite{llava-ov}       & 0.921 & 3.153 & 0.906 & 2.522 & 0.921 & 3.153 & 0.885 & 2.347 & 0.908 & 2.794 \\
LFM-VL 2.5~\cite{lfm2}         & 0.935 & 3.471 & 0.884 & 2.272 & 0.935 & 3.471 & 0.862 & 1.835 & 0.904 & 2.762 \\
\midrule
\rowcolor[HTML]{E8F4FC} 
SOLAR                          & \textbf{0.995} & \textbf{4.741} & 0.894 & 2.649 & \textbf{0.973} & \textbf{4.549} & \textbf{0.920} & \textbf{3.334} & \textbf{0.946} & \textbf{3.818} \\
\bottomrule
\end{tabular}
}
\end{table}

\subsection{Interpretability}

\begin{figure}[H]
    \centering
    \includegraphics[width=\linewidth]{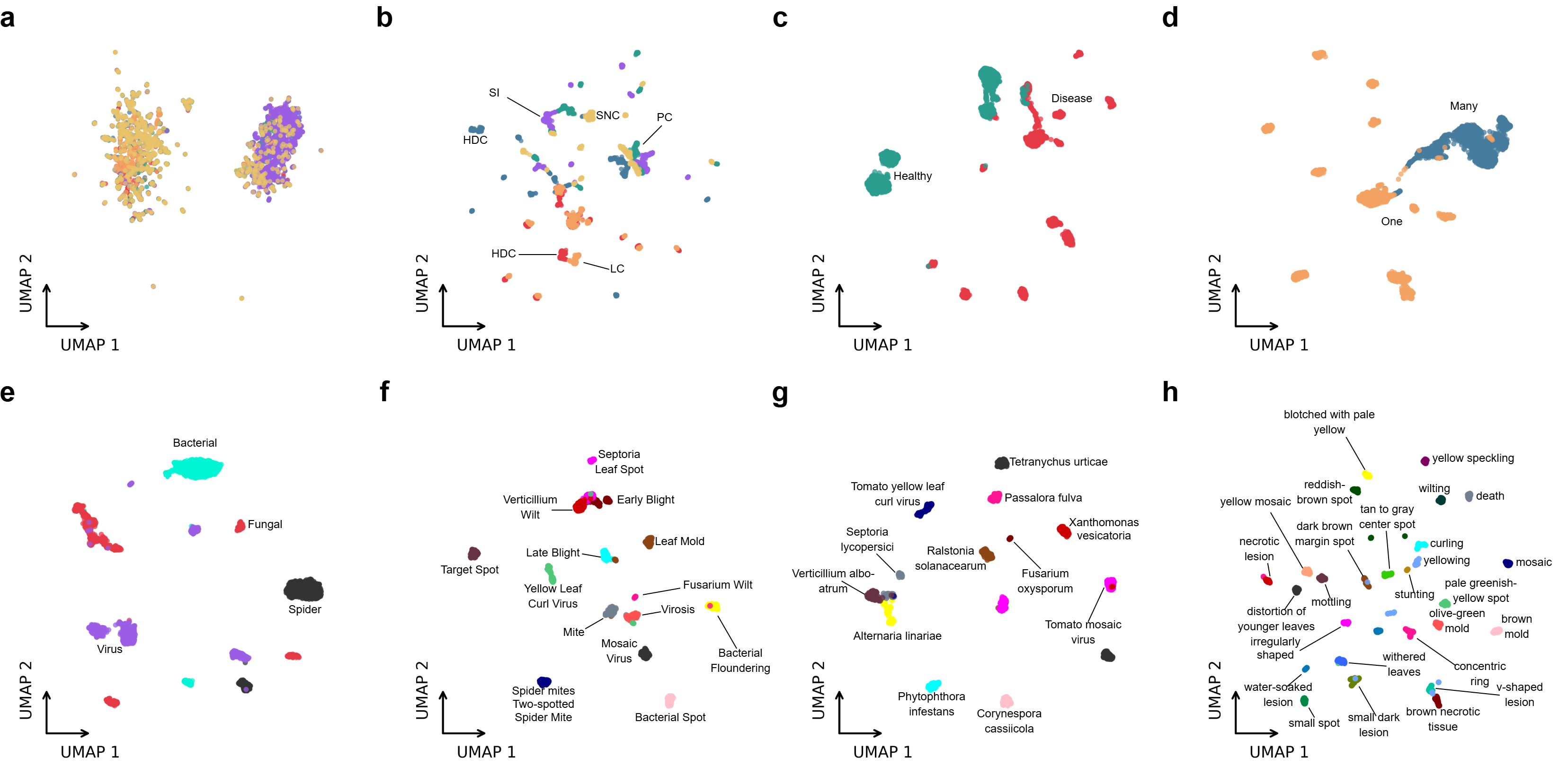}
    \caption{\textbf{UMAP visualizations of learned visual features}. \textbf{a,} UMAP of image embeddings before the MoE layer. \textbf{b,} UMAP by the 6 VQA tasks. \textbf{c,} UMAP of the HDC task. \textbf{d,} UMAP of the LC task. \textbf{e,} UMAP of the PC task. \textbf{f,} UMAP of the DC task. \textbf{g,} UMAP of the SNC task. \textbf{h,} UMAP of the SI task.}
    \label{fig:umap}
\end{figure}

\noindent \textbf{Learned representations from SOLAR. }To qualitatively evaluate the representation learning capability of SOLAR, we analyzed the UMAP visualizations of its learned visual feature embeddings (Figure~\ref{fig:umap}). Figure~\ref{fig:umap}a establishes original image embeddings, which then resolve into highly distinct, task-specific manifolds when categorized across the six VQA tasks by the MoE layer (Figure~\ref{fig:umap}b). In particular, basic perceptual tasks such as HDC and LC formed well-separated binary decision boundaries by SOLAR, demonstrating robust linear separability (Figure~\ref{fig:umap}c-d). When processing biological features, the PC task yielded compact clusters corresponding to distinct pathogen types (Figure~\ref{fig:umap}e). For high-precision diagnostic outputs, both DC and SNC exhibited tight intra-class compactness and clear inter-class margins across 15 tomato disease categories (Figure~\ref{fig:umap}f) and 12 scientific nomenclatures (Figure~\ref{fig:umap}g). This indicates that SOLAR effectively grounds expert pathology logic into distinct feature spaces. Lastly, on the most challenging Symptom Identification (SI) task, SOLAR successfully partitioned 26 fine-grained symptoms (Figure~\ref{fig:umap}h), evidencing precise, localized feature extraction despite high semantic overlap among visual symptom presentations.

\noindent \textbf{Expert Routing for multiple VQA tasks. }We next hypothesized that configuring an MoE architecture with $n=6$ experts and a Top-1 hard routing mechanism would optimally map the six distinct question types in our dataset (HDC, DC, LC, SI, PC, and SNC) to the experts, thereby allowing each expert to specialize exclusively in a single task. However, this configuration yielded the lowest overall performance, achieving an average accuracy of $43.46\%$ and an F1 of $0.468$ (Table~\ref{tab:TableExpertCapacity}). This severe performance degradation stems from the semantic interdependence inherent to domain-specific multi-task VQA. For instance, answering a scientific nomenclature (SNC) question intrinsically relies on recognizing both the visual disease category (DC) and related localized symptoms (SI). Enforcing a fixed routing path completely isolated the specialized experts, precluding essential cross-task knowledge transfer. Consequently, each expert was forced to independently learn visual features and textual reasoning from an artificially narrowed data partition. This architectural bottleneck prevented the model from capturing generalized foundational representations and ultimately precipitated severe overfitting.

\begin{figure}[t]
    \centering
    \includegraphics[width=\linewidth]{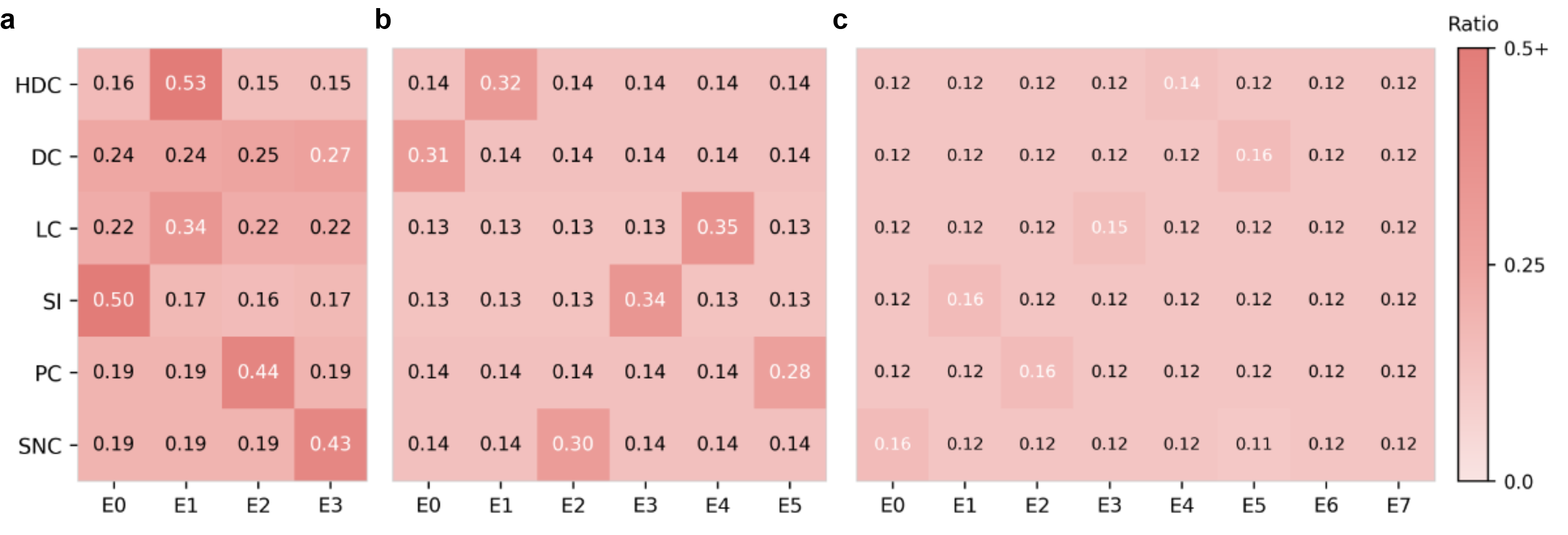}
    \caption{\textbf{Expert routing distribution heatmaps for image-level MoE with different configurations of expert numbers: $n=4$ (a), $n=6$ (b), and $n=8$ (c)}. Each cell indicates the average gating probability of a specific question type routed to the corresponding expert}
    \label{fig:triple_heatmap}
\end{figure}

\begin{figure}[H]
    \centering
    \includegraphics[width=\linewidth]{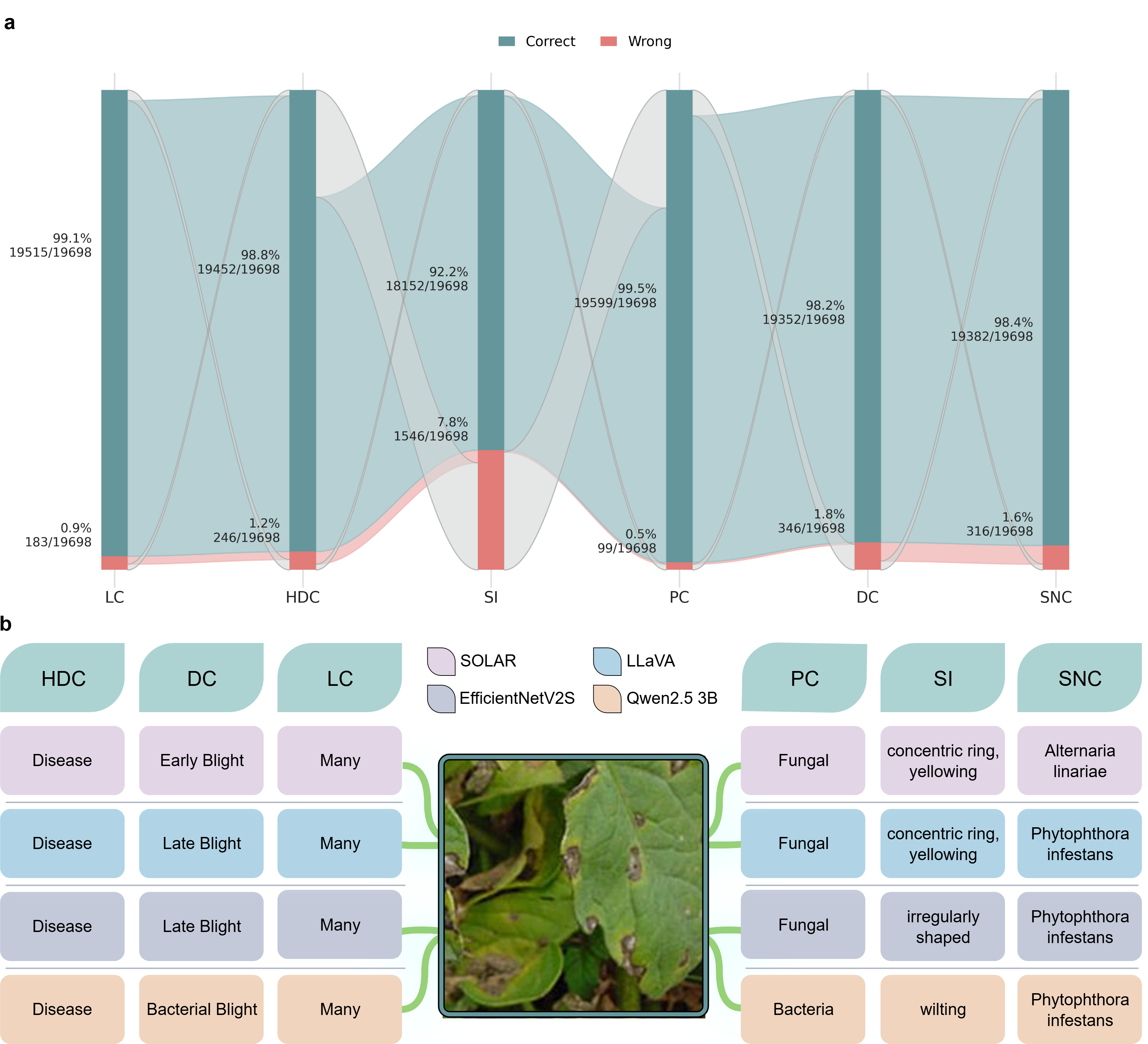}
    \caption{\textbf{Diagnostic error propagation and qualitative performance of SOLAR across VQA tasks.} \textbf{a,} Alluvial diagram tracking prediction accuracy and error propagation flows across the six sequential VQA tasks (LC $\to$ SNC) for on TomaMMU. Teal bands represent correct predictions, while red bands indicate incorrect predictions. The flow thicknesses correspond to the proportion of samples, highlighting SOLAR's near-perfect accuracy on Pathogen Classification (PC, $99.5\%$ correct) and demonstrating that Symptom Identification (SI) remains the most challenging bottleneck, harboring the highest error rate ($7.8\%$). \textbf{b,} Qualitative comparative analysis of model predictions for a representative pathological leaf sample. Responses from SOLAR (purple) are contrasted against LLaVA (light blue), EfficientNetV2S (gray-blue), and Qwen2.5-3B (orange) across all six tasks. While all models correctly identify the perceptual Leaf Count (LC: "One") and basic Healthy/Disease Classification (HDC: "Disease"), only SOLAR successfully correlates the fine-grained visual symptoms ("concentric ring, yellowing") to the correct Disease Classification (DC: "Early Blight") and corresponding Scientific Nomenclature Classification (SNC: \textit{Alternaria linariae}). Conversely, the baseline models exhibit cascading errors, misclassifying the sample as Late Blight (\textit{Phytophthora infestans}) or Bacterial Blight.}
    \label{fig:example}
\end{figure}

\noindent \textbf{Qualitative analysis and hierarchical error propagation.} To interpret the diagnostic reasoning of SOLAR, we analyzed the hierarchical flow of its predictions across the sequential VQA tasks (Figure~\ref{fig:example}a). The alluvial diagram illustrates the propagation of diagnostic logic, revealing that SOLAR maintains exceptional accuracy on foundational perceptual and biological tasks, such as Leaf Counting ($99.1\%$), Healthy/Disease Classification ($98.8\%$), and Pathogen Classification ($99.5\%$). However, Symptom Identification (SI) emerges as the primary bottleneck, containing the highest proportion of misclassifications ($7.8\%$). Because the agricultural VQA tasks are semantically interdependent, errors at the intermediate symptom level naturally cascade into downstream failures in Disease Classification (DC) and Scientific Nomenclature Classification (SNC). This hierarchical dependence is further exemplified through qualitative comparison (Figure~\ref{fig:example}b). When processing a representative pathological sample, all evaluated models successfully resolved the basic perceptual tasks (LC: ``One''; HDC: ``Disease''). However, the baseline architectures (LLaVA, EfficientNetV2S, and Qwen2.5-3B) failed to extract the correct localized features, leading to incorrect symptom and pathogen identification and subsequent misdiagnoses (e.g., incorrectly predicting Late Blight or Bacterial Blight). In contrast, SOLAR explicitly identified the correct fine-grained symptoms (``concentric ring, yellowing'') and pathogen type (``Fungal''). By correctly anchoring its reasoning in these intermediate biological features, SOLAR successfully deduced the specific disease (``Early Blight'') and its corresponding scientific nomenclature (\textit{Alternaria linariae}). This demonstrates a transparent, step-by-step diagnostic progression that mirrors the hierarchical logic of an expert pathologist.

\subsection{Ablation Study}

To isolate the contributions of key architectural components within SOLAR, we conducted comprehensive ablation studies across five dimensions: (1) the individual impact of core modules, (2) the effectiveness of different vision encoders, (3) the scaling of the large language model (LLM) layers, (4) the positioning of the MoE layer, and (5) the optimal number of experts. To ensure a rigorous and fair comparison, all experiments were executed under identical configurations and evaluation protocols, as detailed in Section~\ref{sec:experimentalsetup}.

\begin{figure}[H]
    \centering
    \includegraphics[width=\linewidth]{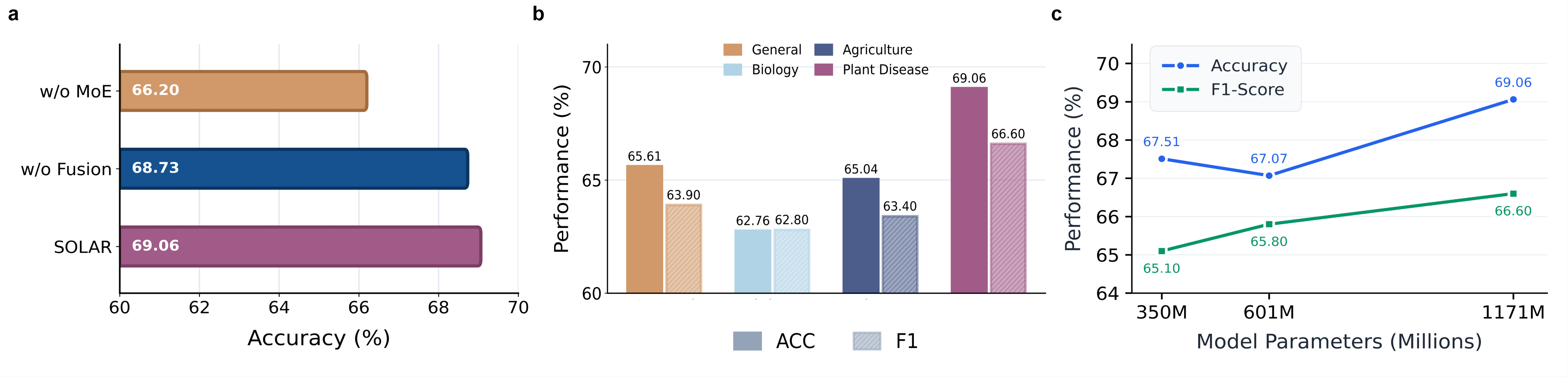}
    \caption{\textbf{Ablation studies and analysis of the SOLAR architecture}. \textbf{a,} Performance of SOLAR when removing individual key components. \textbf{b,} Performance of SOLAR when employing different vision encoder backbones (CLIP, BioCLIP, AgriCLIP, and SCOLD). \textbf{c,} Performance of SOLAR when employing different LLM decoders (BioGPT, Qwen3, and LFM-VL 2.5).}
    \label{fig:Abl_triple}
\end{figure}

\noindent \textbf{Effectiveness of core modules in SOLAR.} To delineate the individual contributions of SOLAR's architectural components to overall diagnostic performance, we evaluated three configurations: the isolated removal of the Mixture of Experts layer (w/o MoE), the removal of the fusion adapter (w/o Fusion Adapter), and the complete SOLAR architecture (Figure~\ref{fig:Abl_triple}a and detail numbers in Table~\ref{tab:TableComponets} for each dataset). 
Specifically, across all four datasets, the fully integrated SOLAR model secured the highest average performance ($69.06\%$ Acc, $0.666$ F1), exceeding the w/o Fusion Adapter by $+0.33\%$ and $+0.002$, and the w/o MoE by $+2.86\%$ and $+0.024$. This confirms that the synergistic integration of both the MoE and fusion layers is essential for maximizing generalized diagnostic capabilities. (See detail in Tables~\ref{tab:ablation_moe_tomammu}-\ref{tab:ablation_moe_tomaad})

\noindent \textbf{Performance of Different Vision Encoders.} To assess the impact of the visual feature extractor on diagnostic capability, we substituted SOLAR's default SCOLD encoder with three widely used vision backbones: CLIP, BioCLIP, and AgriCLIP (Figure~\ref{fig:Abl_triple}b and Table~\ref{tab:TableVisionEncoder}). As demonstrated in Figure~\ref{fig:Abl_triple}b, substituting SCOLD with less specialized encoders resulted in a marked decline in overall diagnostic performance. The specialized plant disease encoder (SCOLD) achieved the highest overall results, reaching $69.06\%$ Acc and a $66.60\%$ F1. In comparison, the general-domain model (CLIP) achieved $65.61\%$ Acc and $63.90\%$ F1, while the macro-agricultural model (AgriCLIP) scored $65.04\%$ Acc and $63.40\%$ F1. The taxonomy-focused biology model (BioCLIP) exhibited the lowest performance ($62.76\%$ Acc, $62.80\%$ F1). These results indicate that general or broadly agricultural models lack the fine-grained perceptual capabilities required for precise symptom identification. SCOLD's domain-specific pre-training effectively captures the key details of foliar diseases, yielding superior foundational representations for downstream pathology tasks. (See detail in Tables~\ref{tab:detailed_vision_encoder_tomammu}-\ref{tab:detailed_vision_encoder_tomaad})

\noindent \textbf{Performance of different LLMs.} To elucidate the impact of the underlying large language model (LLM) backbone on diagnostic reasoning, we compared SOLAR's default textual decoder, LFM-VL 2.5, against two SOTA generative models: Qwen3 \cite{qwen3} and BioGPT\cite{biogpt} (Figure~\ref{fig:Abl_triple}c and Table~\ref{tab:TableLLM}). As illustrated in Figure~\ref{fig:Abl_triple}c, substituting the text decoder significantly impacts diagnostic efficacy across different parameter scales.  SOLAR with LFM-VL 2.5 (1,171M parameters) achieved the highest overall performance, reaching $69.06\%$ Acc and a $66.60\%$ F1. This configuration yielded substantial performance gains of $+1.55\%$ in Acc and $+1.50\%$ in F1 compared to the BioGPT backbone (350M parameters), alongside respective improvements of $+1.99\%$ and $+0.80\%$ over the Qwen3 backbone (601M parameters). This demonstrates that LFM-VL 2.5's targeted multimodal alignment and architectural efficiency provide superior diagnostic reasoning capabilities compared to purely general-domain or tangentially related medical models. (See detail in Tables~\ref{tab:detailed_llm_tomammu}-\ref{tab:detailed_llm_tomaad})

\noindent \textbf{MoE layer positioning.} To determine the optimal integration flow for expert selection, we evaluated three placements for the Mixture of Experts (MoE) module (Table~\ref{tab:TablePositionMoe}). \textit{Dual MoE} employs independent MoE blocks on the image and text branches prior to combination. \textit{Fusion MoE} applies a single MoE module late in the network, processing the already-mixed multimodal features. Finally, our proposed \textit{Image MoE} places the module exclusively on the visual branch, uniquely utilizing the textual question as the gating signal to route visual features. 
While the Dual MoE configuration marginally outperformed Image MoE on the TomaMMU (by $+0.57\%$ Acc, $+0.005$ F1) and TLD-3 (by $+0.34\%$ Acc, $+0.004$ F1) datasets, Image MoE proved substantially more robust across diverse and challenging diagnostic scenarios. On the TLID dataset, Image MoE ($33.70\%$ Acc, $0.335$ F1) exceeded Dual MoE by $+0.87\%$ (Acc) and $+0.008$ (F1), and Fusion MoE by $+0.90\%$ and $+0.009$. This advantage widened significantly on the TomaAD dataset ($53.20\%$ Acc, $0.501$ F1), where Image MoE outperformed Fusion MoE by $+3.81\%$ and $+0.030$, and Dual MoE by a substantial $+7.85\%$ and $+0.061$. 
Overall, the question-gated Image MoE strategy achieved the highest average performance ($69.06\%$ Acc, $0.666$ F1), yielding net improvements of $+1.36\%$ and $+0.011$ over Fusion MoE, and $+1.95\%$ and $+0.015$ over Dual MoE. This indicates that using linguistic cues to dynamically route foundational visual features is more effective than fully independent or post-fusion routing. (See detail in Tables~\ref{tab:detailed_moe_placement_tomammu}-\ref{tab:detailed_moe_placement_tomaad})

\begin{table}[H]
\caption{Performance of different MoE placement strategies across four datasets in the SOLAR architecture}
\label{tab:TablePositionMoe}
\centering
\resizebox{\textwidth}{!}{%
\begin{tabular}{lcccccccccc}
\hline
& \multicolumn{2}{c}{TomaMMU} & \multicolumn{2}{c}{TLID} & \multicolumn{2}{c}{TLD-3} & \multicolumn{2}{c}{TomaAD} & \multicolumn{2}{c}{Average} \\
\cmidrule(lr){2-3} \cmidrule(lr){4-5} \cmidrule(lr){6-7} \cmidrule(lr){8-9} \cmidrule(lr){10-11}
\multirow{-2.5}{*}{\textbf{Type}} & Acc (\%) & F1 & Acc (\%) & F1 & Acc (\%) & F1 & Acc (\%) & F1 & Acc (\%) & F1 \\
\hline
Dual MoE & \textbf{98.25} & \textbf{0.979} & 32.83          & 0.327          & \textbf{92.01} & \textbf{0.859} & 45.35          & 0.440          & 67.11          & 0.651          \\
Fusion MoE        & 97.02          & 0.966          & 32.80          & 0.326          & 91.60          & 0.855          & 49.39          & 0.471          & 67.70          & 0.655          \\
\rowcolor[HTML]{E8F4FC}
Image MoE & 97.68          & 0.974          & \textbf{33.70} & \textbf{0.335}          & 91.67          & 0.855          & \textbf{53.20} & \textbf{0.501}          & \textbf{69.06} & \textbf{0.666}          \\
\hline
\end{tabular}%
}

\end{table}

\noindent \textbf{Number of experts.} To determine the optimal capacity of the Image MoE layer, we evaluated the impact of varying the total number of experts ($n \in \{4, 6, 8\}$) while maintaining a constant Top-$2$ routing mechanism ($k = 2$) (Table~\ref{tab:TableExpertCapacity}). Top-$2$ routing restricts the gating network to select and aggregate outputs from only the two highest-scoring experts per input image. This ensures a stable, bounded computational cost while promoting collaborative feature sharing between the activated experts.
Focusing on these Top-$2$ configurations, our default setting of $n=4$ experts achieved the highest overall performance, with $69.06\%$ Acc and $0.666$ F1 ($+1.56\%$ Acc and $-0.031$ F1 relative to $n=6$). Specifically, the $n=4$ configuration maximized performance on the TomaMMU ($97.68\%$ Acc; $+0.28\%$ over $n=6$) and TomaAD ($53.20\%$ Acc; $+6.69\%$ over $n=6$) datasets. Conversely, on the TLID and TLD-3 datasets, $n=4$ yielded slightly lower accuracies of $33.70\%$ and $91.67\%$, trailing the $n=6$ configuration by margins of $-0.53\%$ and $-0.37\%$, respectively. 
The general performance degradation observed with larger expert pools ($n \geq 6$) suggests that excessive capacity exacerbates overfitting on limited training data and increases the gating network's optimization complexity. As visualized in Figure~\ref{fig:triple_heatmap}, a moderate number of experts ($n=4$) encourages distinct functional specialization. In contrast, an over-parameterized MoE layer results in dispersed routing patterns, load imbalances, and a failure of individual experts to develop localized diagnostic specializations. (See detail in Tables~\ref{tab:moe_ablation_tomammu}-\ref{tab:moe_ablation_tomaad})

\begin{table}[H]
\caption{Ablation study on the number of experts in the Image MoE layer of the SOLAR architecture}
\label{tab:TableExpertCapacity}
\centering
\resizebox{\textwidth}{!}{%
\begin{tabular}{cccccccccccc}
\hline
& & \multicolumn{2}{c}{TomaMMU} & \multicolumn{2}{c}{TLID} & \multicolumn{2}{c}{TLD-3} & \multicolumn{2}{c}{TomaAD} & \multicolumn{2}{c}{Average} \\
\cmidrule(lr){3-4} \cmidrule(lr){5-6} \cmidrule(lr){7-8} \cmidrule(lr){9-10} \cmidrule(lr){11-12}
\multirow{-2.5}{*}{\textbf{Expert}} & \multirow{-2.5}{*}{\textbf{Top-k}} & Acc (\%) & F1 & Acc (\%) & F1 & Acc (\%) & F1 & Acc (\%) & F1 & Acc (\%) & F1 \\
\hline
\rowcolor[HTML]{E8F4FC}
\textbf{4} & \textbf{2} & \textbf{97.68} & \textbf{0.974} & 33.70          & 0.335          & 91.67          & 0.855          & \textbf{53.20}          & 0.501          & \textbf{69.06}          & 0.666          \\
6          & 1          & 96.81          & 0.965          & 7.85           & 0.079          & 52.63          & 0.493          & 33.65          & 0.337          & 47.73          & 0.468          \\
6          & 2          & 97.23          & 0.968          & \textbf{34.23} & \textbf{0.369} & \textbf{92.04} & \textbf{0.902} & 46.51          & \textbf{0.548} & 67.50          & \textbf{0.697} \\
8          & 2          & 97.40          & 0.972          & 33.03          & 0.330          & 91.93          & 0.860          & 45.24          & 0.440          & 66.90          & 0.650          \\
\hline
\end{tabular}%
}
\end{table}

\section{Discussion}
In this study, we present SOLAR, a generative multimodal foundation model for a comprehensive understanding of tomato diseases. Through extensive benchmark evaluation across six hierarchical VQA tasks, we show that SOLAR achieves superior performance over existing vision and vision-language models for basic perception, pathology understanding, and precision disease diagnosis. Importantly, in contrast to previous studies that relied on isolated, single-label image classification, we leveraged a generative framework to capture the complementary semantic relationships between visual symptoms and diagnostic reasoning. This multimodal approach successfully mirrors the cognitive, coarse-to-fine diagnostic workflow of an agricultural expert.

The performance gain achieved by SOLAR is largely attributable to its Expert Fusion module, which dynamically routes and integrates visual features based on the semantic context of the input question. Existing agricultural studies have predominantly used static, off-the-shelf foundation models or simple projectors that impose a rigid mapping between image and text spaces \cite{llava-ov,agri_llava}. By contrast, SOLAR is custom-designed with a question-conditioned Image MoE layer. This module allows the network to adaptively emphasize diagnostically relevant visual regions---such as distinct lesion patterns or necrosis---depending on the specific nature of the query. This mechanism effectively prevents the overlap of learned features and mitigates severe overfitting on limited domain data.

The semantic interdependence among agricultural tasks poses a major challenge for training reliable deep learning models. For instance, answering a highly specific SNC question that uses scientific nomenclature inherently relies on accurately recognizing both the visual disease category and localized leaf symptoms. Our approach provides an effective solution to this problem by using a Top-2 MoE routing mechanism across four experts, which maintains stable computational cost while encouraging distinct functional specialization among the experts. This training paradigm can be extended and applied in building interpretable multimodal foundation models for other complex agricultural domains.

Accurate and explainable disease diagnosis has significant practical implications for precision agriculture. There are important conceptual and practical distinctions between basic binary disease detection and the comprehensive understanding of pathology that is the primary focus of most existing agricultural classifiers. Given that simple classifiers only provide a black-box label without contextual grounding, their impact in real-world farming is limited to a basic assistive role. However, fine-grained symptom identification is a much more challenging problem due to the high semantic overlap and substantial intra-class visual variability in disease manifestations. By explicitly anchoring its reasoning in intermediate biological features (such as concentric rings or yellowing), the multimodal SOLAR model significantly improved upon traditional classification pipelines, demonstrating a transparent, step-by-step diagnostic progression. 

Despite its strong overall performance, symptom identification task remains the most difficult bottleneck in the diagnostic pipeline, harboring the highest error rate ($7.8\%$) in our hierarchical analysis. Furthermore, SOLAR exhibited vulnerabilities when generalizing to highly complex, in-the-wild datasets such as TLID. While the model excels in carefully curated environments like TomaMMU and TomaAD, it underperformed compared to large general-domain models (e.g., Qwen3-2B) on the generative tasks in the TLID dataset. This discrepancy highlights the inherent trade-off of using a hyper-specialized plant disease encoder such as SCOLD. In real-world greenhouse environments characterized by chaotic backgrounds, multiple co-infections, and nutrient deficiencies, highly specialized encoders can become overly sensitive to localized pathological artifacts, whereas massive general models maintain a more robust macroscopic understanding of overall plant health. 

Although the results of SOLAR's generative VQA framework are highly promising, they were evaluated primarily on isolated leaf images from specific datasets. Before the model can be considered for widespread implementation in commercial greenhouses, several steps are needed to ensure rigorous evaluation of its robustness and field utility. First, these findings should be validated in future studies with larger, multi-regional datasets that capture extreme environmental variations and complex multi-disease interactions. Second, expanding multimodal inputs to integrate external metadata, such as climate conditions and temporal growth stages, will be required to support precision agricultural management fully. Lastly, while SOLAR excels at identifying diseases and symptoms, translating these diagnostic insights into actionable, localized treatment protocols remains an open challenge. Future iterations must bridge the gap between diagnostic and prescriptive agriculture by generating safe, evidence-based intervention strategies tailored to the farmer's specific environmental constraints.

\section{Conclusion}

In this study, we introduced SOLAR, a generative multimodal architecture that fundamentally reformulates tomato leaf disease diagnosis as a hierarchical Visual Question Answering task. By integrating a question-conditioned fusion Mixture-of-Experts layer with a residual expert fusion mechanism, SOLAR dynamically routes and aligns fine-grained visual features based on the specific semantic context of the diagnostic query. Extensive evaluations across four diverse datasets demonstrate that SOLAR consistently achieves state-of-the-art performance, surpassing existing vision-only architectures, general vision-language models, and domain-specific foundation models in both classification accuracy and generative response quality. Crucially, by mirroring the step-by-step diagnostic reasoning of agricultural experts, SOLAR overcomes the interpretability limitations of traditional "black-box" classifiers, providing a transparent and contextually grounded understanding of pathology. These findings validate the efficacy of dynamic multimodal soft-routing for complex agricultural VQA tasks. Ultimately, SOLAR establishes a highly scalable and explainable framework for precision agriculture, paving the way for next-generation, expert-level disease management systems deployed in real-world farming environments.

\section*{Acknowledgment}
This research was supported by the Hyundai Motor Chung Mong-Koo Foundation Global Scholarship (GSS-25-02120).

\section*{Data Availability}
 TomaMMU, TomaBench, TLID, TomaAD, and TLD-3 are publicly available and can be accessed from the following: TomaMMU and TomaBench (\url{https://huggingface.co/datasets/enalis/TomaMMU}), TLID (\url{https://data.mendeley.com/datasets/kt64b2kh89/2}), TLD-3 (\url{https://data.mendeley.com/datasets/zfv4jj7855/1}) and TomaAD (\url{https://doi.org/10.6084/m9.figshare.31145998}).


\section*{CRediT authorship contribution statement}
\textbf{Khang Nguyen Quoc} contributed to conceptualization, data curation, methodology, software, model development, data analysis and visualization, and writing – original draft. \textbf{Minh-Phuoc Tran} contributed to methodology, data curation, software, model development, data analysis, visualization, and writing – original draft. 
\textbf{Gia-Han Truong} contributed to data curation, validation, and writing – review \& editing.  \textbf{Luyl-Da Quach} contributed to methodology, project administration, supervision, validation, and writing – review \& editing.

\section*{Declaration of Generative AI Use}
The authors acknowledge the use of Claude Sonnet 5 and Grammarly to enhance the grammar and clarity of the manuscript. After using them, the authors reviewed and edited the content as needed and take full responsibility for the content of the published article.

\section*{Declaration of Competing Interest}
The authors declare no known competing conflicts of interest that could have influenced the work reported in this paper.




\bibliographystyle{elsarticle-num} 
\bibliography{cas-refs}

@Article{moe,
  title={Switch transformers: Scaling to trillion parameter models with simple and efficient sparsity},
  author={Fedus, William and Zoph, Barret and Shazeer, Noam},
  journal={Journal of Machine Learning Research},
  volume={23},
  number={120},
  pages={1--39},
  year={2022}
}

@misc{tomammu,
      title={TomaMMU: A Comprehensive Multimodal Understanding Benchmark for Tomato Leaf Diseases}, 
      author={Gia-Han Truong and Khang Nguyen Quoc and Luyl-Da Quach},
      year={2026},
      eprint={2608.08727},
      archivePrefix={arXiv},
      primaryClass={cs.CV},
      url={https://arxiv.org/abs/2608.08727}, 
}

@article{internvl3,
  title={Internvl3: Exploring advanced training and test-time recipes for open-source multimodal models},
  author={Zhu, Jinguo and Wang, Weiyun and Chen, Zhe and Liu, Zhaoyang and Ye, Shenglong and Gu, Lixin and Tian, Hao and Duan, Yuchen and Su, Weijie and Shao, Jie and others},
  journal={arXiv preprint arXiv:2504.10479},
  year={2025}
}

@article{qwen3,
  title={Qwen3-vl technical report},
  author={Bai, Shuai and Cai, Yuxuan and Chen, Ruizhe and Chen, Keqin and Chen, Xionghui and Cheng, Zesen and Deng, Lianghao and Ding, Wei and Gao, Chang and Ge, Chunjiang and others},
  journal={arXiv preprint arXiv:2511.21631},
  year={2025}
}

@misc{qwen25,
      title={Qwen2.5-VL Technical Report}, 
      author={Shuai Bai and Keqin Chen and Xuejing Liu and Jialin Wang and Wenbin Ge and Sibo Song and Kai Dang and Peng Wang and Shijie Wang and Jun Tang and Humen Zhong and Yuanzhi Zhu and Mingkun Yang and Zhaohai Li and Jianqiang Wan and Pengfei Wang and Wei Ding and Zheren Fu and Yiheng Xu and Jiabo Ye and Xi Zhang and Tianbao Xie and Zesen Cheng and Hang Zhang and Zhibo Yang and Haiyang Xu and Junyang Lin},
      year={2025},
      eprint={2502.13923},
      archivePrefix={arXiv},
      primaryClass={cs.CV},
      url={https://arxiv.org/abs/2502.13923}, 
}

@article{llava-ov,
  title={Llava-onevision: Easy visual task transfer},
  author={Li, Bo and Zhang, Yuanhan and Guo, Dong and Zhang, Renrui and Li, Feng and Zhang, Hao and Zhang, Kaichen and Zhang, Peiyuan and Li, Yanwei and Liu, Ziwei and others},
  journal={arXiv preprint arXiv:2408.03326},
  year={2024}
}

@article{lfm2,
 title={LFM2 Technical Report},
 author={Liquid AI},
 journal={arXiv preprint arXiv:2511.23404},
 year={2025}
}

@inproceedings{bioclip,
  title={Bioclip: A vision foundation model for the tree of life},
  author={Stevens, Samuel and Wu, Jiaman and Thompson, Matthew J and Campolongo, Elizabeth G and Song, Chan Hee and Carlyn, David Edward and Dong, Li and Dahdul, Wasila M and Stewart, Charles and Berger-Wolf, Tanya and others},
  booktitle={Proceedings of the IEEE/CVF conference on computer vision and pattern recognition},
  pages={19412--19424},
  year={2024}
}

@article{biogpt,
   title={BioGPT: generative pre-trained transformer for biomedical text generation and mining},
   volume={23},
   ISSN={1477-4054},
   url={http://dx.doi.org/10.1093/bib/bbac409},
   DOI={10.1093/bib/bbac409},
   number={6},
   journal={Briefings in Bioinformatics},
   publisher={Oxford University Press (OUP)},
   author={Luo, Renqian and Sun, Liai and Xia, Yingce and Qin, Tao and Zhang, Sheng and Poon, Hoifung and Liu, Tie-Yan},
   year={2022},
   month=Sept }

@article{agri_llava,
  title={Agri-llava: Knowledge-infused large multimodal assistant on agricultural pests and diseases},
  author={Wang, Liqiong and Jin, Teng and Yang, Jinyu and Leonardis, Ales and Wang, Fangyi and Zheng, Feng},
  journal={arXiv preprint arXiv:2412.02158},
  year={2024}
}

@article{leafmd,
  title={A multiregional image--text dataset and benchmark for vision-language modeling of plant diseases},
  author={Nguyen, Trang V and Nguyen Quoc, Khang and Harwath, David and Quach, Luyl-Da and Dao, Phuong D},
  journal={bioRxiv},
  pages={2026--07},
  year={2026},
  publisher={Cold Spring Harbor Laboratory}
}

@article{siglip2,
  title={Siglip 2: Multilingual vision-language encoders with improved semantic understanding, localization, and dense features},
  author={Tschannen, Michael and Gritsenko, Alexey and Wang, Xiao and Naeem, Muhammad Ferjad and Alabdulmohsin, Ibrahim and Parthasarathy, Nikhil and Evans, Talfan and Beyer, Lucas and Xia, Ye and Mustafa, Basil and others},
  journal={arXiv preprint arXiv:2502.14786},
  year={2025}
}

@article{scold,
  title={A vision-language foundation model for leaf disease identification},
  author={Quoc, Khang Nguyen and Thu, Lan Le Thi and Quach, Luyl-Da},
  journal={Expert Systems with Applications},
  pages={130084},
  year={2025},
  publisher={Elsevier}
}

@inproceedings{clip,
  title={Learning transferable visual models from natural language supervision},
  author={Radford, Alec and Kim, Jong Wook and Hallacy, Chris and Ramesh, Aditya and Goh, Gabriel and Agarwal, Sandhini and Sastry, Girish and Askell, Amanda and Mishkin, Pamela and Clark, Jack and others},
  booktitle={International conference on machine learning},
  pages={8748--8763},
  year={2021},
  organization={PmLR}
}

@inproceedings{resnet50,
  title={Deep residual learning for image recognition},
  author={He, Kaiming and Zhang, Xiangyu and Ren, Shaoqing and Sun, Jian},
  booktitle={Proceedings of the IEEE conference on computer vision and pattern recognition},
  pages={770--778},
  year={2016}
}

@inproceedings{efficientnetv2,
  title={Efficientnetv2: Smaller models and faster training},
  author={Tan, Mingxing and Le, Quoc},
  booktitle={International conference on machine learning},
  pages={10096--10106},
  year={2021},
  organization={PMLR}
}

@article{densenet,
  title={Densenet: Implementing efficient convnet descriptor pyramids},
  author={Iandola, Forrest and Moskewicz, Matt and Karayev, Sergey and Girshick, Ross and Darrell, Trevor and Keutzer, Kurt},
  journal={arXiv preprint arXiv:1404.1869},
  year={2014}
}

@inproceedings{inception,
  title={Inception-v4, inception-resnet and the impact of residual connections on learning},
  author={Szegedy, Christian and Ioffe, Sergey and Vanhoucke, Vincent and Alemi, Alexander},
  booktitle={Proceedings of the AAAI conference on artificial intelligence},
  volume={31},
  number={1},
  year={2017}
}

@inproceedings{rouge,
  title={Rouge: A package for automatic evaluation of summaries},
  author={Lin, Chin-Yew},
  booktitle={Text summarization branches out},
  pages={74--81},
  year={2004}
}

@inproceedings{agriclip,
  title={AgriCLIP: Adapting CLIP for agriculture and livestock via domain-specialized cross-model alignment},
  author={Nawaz, Umair and Muhammad, Awais and Gani, Hanan and Naseer, Muzammal and Khan, Fahad Shahbaz and Khan, Salman and Anwer, Rao},
  booktitle={Proceedings of the 31st International Conference on Computational Linguistics},
  pages={9630--9639},
  year={2025}
}

@online{ECTomato,
  author       = {{Directorate-General for Agriculture and Rural Development}},
  title        = {Tomato dashboard – information on the Agri-food data portal},
  year         = {2026},
  url          = {https://agridata.ec.europa.eu/extensions/DashboardTomato/Dashboard.html},
  organization = {European Commission}
}

@article{107054,
  title={A diverse ensemble classifier for tomato disease recognition},
  author={Astani, Mounes and Hasheminejad, Mohammad and Vaghefi, Mahsa},
  journal={Computers and Electronics in Agriculture},
  volume={198},
  doi={https://doi.org/10.1016/j.compag.2022.107054},
  pages={107054},
  year={2022},
  publisher={Elsevier}
}

@article{106997,
  title={Tomato disease and pest diagnosis method based on the Stacking of prescription data},
  author={Xu, Chang and Ding, Junqi and Qiao, Yan and Zhang, Lingxian},
  journal={Computers and Electronics in Agriculture},
  volume={197},
  pages={106997},
  doi={10.1016/j.compag.2022.106997},
  year={2022},
  publisher={Elsevier}
}

@article{1356260,
  title={Revolutionizing agriculture with artificial intelligence: plant disease detection methods, applications, and their limitations},
  author={Jafar, Abbas and Bibi, Nabila and Naqvi, Rizwan Ali and Sadeghi-Niaraki, Abolghasem and Jeong, Daesik},
  journal={Frontiers in Plant Science},
  volume={15},
  pages={1356260},
  doi={10.3389/fpls.2024.1356260},
  year={2024},
  publisher={Frontiers Media SA}
}

@article{1041514,
  title={AI-based object detection latest trends in remote sensing, multimedia and agriculture applications},
  author={Nawaz, Saqib Ali and Li, Jingbing and Bhatti, Uzair Aslam and Shoukat, Muhammad Usman and Ahmad, Raza Muhammad},
  journal={Frontiers in Plant Science},
  volume={13},
  pages={1041514},
  doi={10.3389/fpls.2022.1041514},
  year={2022},
  publisher={Frontiers Media SA}
}

@article{1435016,
  title={Intelligent agriculture: Deep learning in UAV-based remote sensing imagery for crop diseases and pests detection},
  author={Zhu, Hongyan and Lin, Chengzhi and Liu, Gengqi and Wang, Dani and Qin, Shuai and Li, Anjie and Xu, Jun-Li and He, Yong},
  journal={Frontiers in Plant Science},
  volume={15},
  pages={1435016},
  doi={10.3389/fpls.2024.1435016},
  year={2024},
  publisher={Frontiers}
}

@article{s13007,
  title={An efficient deep learning model for tomato disease detection},
  author={Wang, Xuewei and Liu, Jun},
  journal={Plant Methods},
  volume={20},
  number={1},
  pages={61},
  doi={10.1186/s13007-024-01188-1},
  year={2024},
  publisher={Springer}
}

@article{125737,
  title={Mixed data augmentation and osprey search strategy for enhancing YOLO in tomato disease, pest, and weed detection},
  author={Lin, Jiewen and Hu, Gui and Chen, Jian},
  journal={Expert Systems with Applications},
  volume={264},
  pages={125737},
  doi={10.1016/j.eswa.2024.125737},
  year={2025},
  publisher={Elsevier}
}

@article{0131011,
  title = {Evaluation of the Efficiency of the Optimization Algorithms for Transfer Learning on the Rice Leaf Disease Dataset},
  volume = {13},
  ISSN = {2158-107X},
  url = {http://dx.doi.org/10.14569/IJACSA.2022.0131011},
  DOI = {10.14569/ijacsa.2022.0131011},
  number = {10},
  journal = {International Journal of Advanced Computer Science and Applications},
  publisher = {The Science and Information Organization},
  author = {Quach,  Luyl-Da and Quoc,  Khang Nguyen and Quynh,  Anh Nguyen and Ngoc,  Hoang Tran},
  year = {2022}
}

@article{3358333,
  title = {Real-Time Plant Disease Dataset Development and Detection of Plant Disease Using Deep Learning},
  volume = {12},
  ISSN = {2169-3536},
  url = {http://dx.doi.org/10.1109/ACCESS.2024.3358333},
  DOI = {10.1109/access.2024.3358333},
  journal = {IEEE Access},
  publisher = {Institute of Electrical and Electronics Engineers (IEEE)},
  author = {Joseph,  Diana Susan and Pawar,  Pranav M. and Chakradeo,  Kaustubh},
  year = {2024},
  pages = {16310–16333}
}

@article{100476,
  title = {Classification of mango disease using ensemble convolutional neural network},
  volume = {8},
  ISSN = {2772-3755},
  url = {http://dx.doi.org/10.1016/j.atech.2024.100476},
  DOI = {10.1016/j.atech.2024.100476},
  journal = {Smart Agricultural Technology},
  publisher = {Elsevier BV},
  author = {Bezabh,  Yohannes Agegnehu and Ayalew,  Aleka Melese and Abuhayi,  Biniyam Mulugeta and Demlie,  Tensay Nigussie and Awoke,  Eshete Ayenew and Mengistu,  Taye Endeshaw},
  year = {2024},
  month = Aug,
  pages = {100476}
}

@article{101348,
  title = {An investigation on advances in transfer learning and explainable AI for mango leaf disease detection},
  volume = {12},
  ISSN = {2772-3755},
  url = {http://dx.doi.org/10.1016/j.atech.2025.101348},
  DOI = {10.1016/j.atech.2025.101348},
  journal = {Smart Agricultural Technology},
  publisher = {Elsevier BV},
  author = {Prabhu,  Omkar and T,  Manoj and Shetty,  Sucharitha},
  year = {2025},
  month = Dec,
  pages = {101348}
}

@inbook{Quach2024,
  title = {Explainable AI for Plant Disease Detection: Assessing Explainability in Classifying Maize Leaves Diseases with Focus Score and Ablation-CAM},
  ISBN = {9789819796137},
  ISSN = {1865-0937},
  url = {http://dx.doi.org/10.1007/978-981-97-9613-7\_2},
  DOI = {10.1007/978-981-97-9613-7\_2},
  booktitle = {Intelligent Systems and Data Science},
  publisher = {Springer Nature Singapore},
  author = {Quach,  Luyl-Da and Quoc,  Khang Nguyen and Nguyen,  Chi-Ngon and Thai-Nghe,  Nguyen},
  year = {2024},
  month = Nov,
  pages = {19–32}
}

@article{108155,
  title = {LLM-led vision-spectral fusion: A zero-shot approach to temporal fruit image classification},
  volume = {194},
  ISSN = {0893-6080},
  url = {http://dx.doi.org/10.1016/j.neunet.2025.108155},
  DOI = {10.1016/j.neunet.2025.108155},
  journal = {Neural Networks},
  publisher = {Elsevier BV},
  author = {Wu,  Huyu and Jia,  Bowen and Yuan,  Xue–Ming},
  year = {2026},
  month = Feb,
  pages = {108155}
}

@article{Li2025,
  title = {A review on enhancing agricultural intelligence with large language models},
  volume = {15},
  ISSN = {2589-7217},
  url = {http://dx.doi.org/10.1016/j.aiia.2025.05.006},
  DOI = {10.1016/j.aiia.2025.05.006},
  number = {4},
  journal = {Artificial Intelligence in Agriculture},
  publisher = {Elsevier BV},
  author = {Li,  Hongda and Wu,  Huarui and Li,  Qingxue and Zhao,  Chunjiang},
  year = {2025},
  month = Dec,
  pages = {671–685}
}

@article{Zhu2025,
  title = {Harnessing large vision and language models in agriculture: a review},
  volume = {16},
  ISSN = {1664-462X},
  url = {http://dx.doi.org/10.3389/fpls.2025.1579355},
  DOI = {10.3389/fpls.2025.1579355},
  journal = {Frontiers in Plant Science},
  publisher = {Frontiers Media SA},
  author = {Zhu,  Hongyan and Qin,  Shuai and Su,  Min and Lin,  Chengzhi and Li,  Anjie and Gao,  Junfeng},
  year = {2025},
  month = {Sept} 
}

@misc{23253,
  doi = {10.48550/ARXIV.2511.23253},
  url = {https://arxiv.org/abs/2511.23253},
  author = {Wen,  Yibin and Li,  Qingmei and Ye,  Zi and Zhang,  Jiarui and Fan,  Xiaoya and Mai,  Zurong and Wu,  Jing and Lou,  Shuohong and Chen,  Yuhang and Huang,  Henglian and Zhang,  Yang and Gu,  Defeng and Zhao,  Lingyuan and Lu,  Yutong and Fu,  Haohuan and Huang,  Jianxi and Zheng,  Juepeng},
  title = {AgroCoT: A Chain-of-Thought Benchmark for Evaluating Reasoning in Vision-Language Models for Agriculture},
  publisher = {arXiv},
  year = {2025},
  copyright = {Creative Commons Attribution Share Alike 4.0 International}
}

@misc{19617,
  doi = {10.48550/ARXIV.2407.19617},
  url = {https://arxiv.org/abs/2407.19617},
  author = {Arshad,  Muhammad Arbab and Jubery,  Talukder Zaki and Roy,  Tirtho and Nassiri,  Rim and Singh,  Asheesh K. and Singh,  Arti and Hegde,  Chinmay and Ganapathysubramanian,  Baskar and Balu,  Aditya and Krishnamurthy,  Adarsh and Sarkar,  Soumik},
  title = {Leveraging Vision Language Models for Specialized Agricultural Tasks},
  publisher = {arXiv},
  year = {2024},
  copyright = {arXiv.org perpetual,  non-exclusive license}
}

@inproceedings{00555,
  title = {AgroGPT : Efficient Agricultural Vision-Language Model with Expert Tuning},
  url = {http://dx.doi.org/10.1109/WACV61041.2025.00555},
  DOI = {10.1109/wacv61041.2025.00555},
  booktitle = {2025 IEEE/CVF Winter Conference on Applications of Computer Vision (WACV)},
  publisher = {IEEE},
  author = {Awais,  Muhammad and Salem Abdulla Alharthi,  Ali Husain and Kumar,  Amandeep and Cholakkal,  Hisham and Anwer,  Rao Muhammad},
  year = {2025},
  month = Feb,
  pages = {5687–5696}
}

@inproceedings{Antol2015,
  title = {VQA: Visual Question Answering},
  url = {http://dx.doi.org/10.1109/ICCV.2015.279},
  DOI = {10.1109/iccv.2015.279},
  booktitle = {2015 IEEE International Conference on Computer Vision (ICCV)},
  publisher = {IEEE},
  author = {Antol,  Stanislaw and Agrawal,  Aishwarya and Lu,  Jiasen and Mitchell,  Margaret and Batra,  Dhruv and Zitnick,  C. Lawrence and Parikh,  Devi},
  year = {2015},
  month = Dec,
  pages = {2425–2433}
}

@article{Zhao2024,
  title = {Informed-Learning-Guided Visual Question Answering Model of Crop Disease},
  volume = {6},
  ISSN = {2643-6515},
  url = {http://dx.doi.org/10.34133/plantphenomics.0277},
  DOI = {10.34133/plantphenomics.0277},
  journal = {Plant Phenomics},
  publisher = {Elsevier BV},
  author = {Zhao,  Yunpeng and Wang,  Shansong and Zeng,  Qingtian and Ni,  Weijian and Duan,  Hua and Xie,  Nengfu and Xiao,  Fengjin},
  year = {2024},
  pages = {0277}
}

@article{Nanavaty2024,
  title = {Integrating deep learning for visual question answering in Agricultural Disease Diagnostics: Case Study of Wheat Rust},
  volume = {14},
  ISSN = {2045-2322},
  url = {http://dx.doi.org/10.1038/s41598-024-79793-2},
  DOI = {10.1038/s41598-024-79793-2},
  number = {1},
  journal = {Scientific Reports},
  publisher = {Springer Science and Business Media LLC},
  author = {Nanavaty,  Akash and Sharma,  Rishikesh and Pandita,  Bhuman and Goyal,  Ojasva and Rallapalli,  Srinivas and Mandal,  Murari and Singh,  Vaibhav Kumar and Narang,  Pratik and Chamola,  Vinay},
  year = {2024},
  month = Nov 
}

@article{Lan2023,
  title = {Visual question answering model for fruit tree disease decision-making based on multimodal deep learning},
  volume = {13},
  ISSN = {1664-462X},
  url = {http://dx.doi.org/10.3389/fpls.2022.1064399},
  DOI = {10.3389/fpls.2022.1064399},
  journal = {Frontiers in Plant Science},
  publisher = {Frontiers Media SA},
  author = {Lan,  Yubin and Guo,  Yaqi and Chen,  Qizhen and Lin,  Shaoming and Chen,  Yuntong and Deng,  Xiaoling},
  year = {2023},
  month = Jan 
}

@misc{17117,
  doi = {10.48550/ARXIV.2508.17117},
  url = {https://arxiv.org/abs/2508.17117},
  author = {Sakib,  Syed Nazmus and Haque,  Nafiul and Hossain,  Mohammad Zabed and Arman,  Shifat E.},
  title = {PlantVillageVQA: A Visual Question Answering Dataset for Benchmarking Vision-Language Models in Plant Science},
  publisher = {arXiv},
  year = {2025},
  copyright = {Creative Commons Attribution Share Alike 4.0 International}
}

@article{agmmu,
  title={Agmmu: A comprehensive agricultural multimodal understanding benchmark},
  author={Gauba, Aruna and Pi, Irene and Man, Yunze and Pang, Ziqi and Adve, Vikram and Wang, Yu-Xiong},
  journal={Advances in Neural Information Processing Systems},
  volume={38},
  year={2026}
}

@article{210103961F,
       author = {{Fedus}, William and {Zoph}, Barret and {Shazeer}, Noam},
        title = "{Switch Transformers: Scaling to Trillion Parameter Models with Simple and Efficient Sparsity}",
      journal = {arXiv e-prints},
         year = 2021,
        month = jan,
          eid = {arXiv:2101.03961},
        pages = {arXiv:2101.03961},
          doi = {10.48550/arXiv.2101.03961},
archivePrefix = {arXiv},
       eprint = {2101.03961},
       adsurl = {https://ui.adsabs.harvard.edu/abs/2021arXiv210103961F}
}

@misc{kim2026,
      title={Geometric Regularization in Mixture-of-Experts: The Disconnect Between Weights and Activations}, 
      author={Hyunjun Kim},
      year={2026},
      eprint={2601.00457},
      archivePrefix={arXiv},
      primaryClass={cs.LG},
      url={https://arxiv.org/abs/2601.00457}, 
}

@misc{lora,
      title={LoRA: Low-Rank Adaptation of Large Language Models}, 
      author={Edward J. Hu and Yelong Shen and Phillip Wallis and Zeyuan Allen-Zhu and Yuanzhi Li and Shean Wang and Lu Wang and Weizhu Chen},
      year={2021},
      eprint={2106.09685},
      archivePrefix={arXiv},
      primaryClass={cs.CL},
      url={https://arxiv.org/abs/2106.09685}, 
}

@misc{leafnet,
      title={LeafNet: A Large-Scale Dataset and Comprehensive Benchmark for Foundational Vision-Language Understanding of Plant Diseases}, 
      author={Khang Nguyen Quoc and Phuong D. Dao and Luyl-Da Quach},
      year={2026},
      eprint={2602.13662},
      archivePrefix={arXiv},
      primaryClass={cs.CV},
      url={https://arxiv.org/abs/2602.13662}, 
}

@misc{judge,
      title={LLMs-as-Judges: A Comprehensive Survey on LLM-based Evaluation Methods}, 
      author={Haitao Li and Qian Dong and Junjie Chen and Huixue Su and Yujia Zhou and Qingyao Ai and Ziyi Ye and Yiqun Liu},
      year={2024},
      eprint={2412.05579},
      archivePrefix={arXiv},
      primaryClass={cs.CL},
      url={https://arxiv.org/abs/2412.05579}, 
}

@misc{gemini25,
      title={Gemini 2.5: Pushing the Frontier with Advanced Reasoning, Multimodality, Long Context, and Next Generation Agentic Capabilities}, 
      author={Gheorghe Comanici and Eric Bieber and Mike Schaekermann and Ice Pasupat and Noveen Sachdeva and Inderjit Dhillon and Marcel Blistein and Ori Ram and orthers},
      year={2025},
      eprint={2507.06261},
      archivePrefix={arXiv},
      primaryClass={cs.CL},
      url={https://arxiv.org/abs/2507.06261}, 
}

@article{TLID,
  title = {Enhancing Disease and Pest Detection in Greenhouse Tomato Cultivation Using Advanced Machine Learning on New Dataset of Images},
  volume = {31},
  ISSN = {1678-4804},
  url = {http://dx.doi.org/10.5753/jbcs.2025.4581},
  DOI = {10.5753/jbcs.2025.4581},
  number = {1},
  journal = {Journal of the Brazilian Computer Society},
  publisher = {Sociedade Brasileira de Computacao - SB},
  author = {Zimmermann,  Grasielli B. and Pellenz,  Marcelo E. and Costa,  Yandre M. G. and Britto Jr.,  Alceu de S.},
  year = {2025},
  month = Mar,
  pages = {187–202}
}

@misc{TLD-3,
  doi = {10.17632/ZFV4JJ7855.1},
  url = {https://data.mendeley.com/datasets/zfv4jj7855/1},
  author = {Solapure,  Vaibhav and DY,  SmartAgroTech and JAWALE,  ANISH},
  title = {Tomato Leaf Disease Dataset},
  publisher = {Mendeley Data},
  year = {2024}
}

@article{TomaAD,
  title = {An end-to-end vision–language framework with LLM-based reasoning for decision-oriented tomato leaf disease management in greenhouse environments},
  volume = {247},
  ISSN = {0168-1699},
  url = {http://dx.doi.org/10.1016/j.compag.2026.111692},
  DOI = {10.1016/j.compag.2026.111692},
  journal = {Computers and Electronics in Agriculture},
  publisher = {Elsevier BV},
  author = {Shafay,  Muhammad and Velayudhan,  Divya and Owais,  Muhammad and Hassan,  Taimur and Seneviratne,  Lakmal and Hussain,  Irfan and Werghi,  Naoufel},
  year = {2026},
  month = June,
  pages = {111692}
}

@inproceedings{3654570,
author = {Nguyen, Khang Quoc and Nguyen, Huy Cao Gia and Le, Trinh Ngo Diem and Thai, Viet Binh Quoc and Tran, Binh Phong and Le, Lan Thi Thu and Quach, Luyl-Da},
title = {Tiny-CNN: Structuring Convolutional Neural Networks for Accurate Classification of Rice Leaf Diseases in Resource-Constrained Environments},
year = {2024},
isbn = {9798400716713},
publisher = {Association for Computing Machinery},
address = {New York, NY, USA},
url = {https://doi.org/10.1145/3654522.3654570},
doi = {10.1145/3654522.3654570},
booktitle = {Proceedings of the 2024 9th International Conference on Intelligent Information Technology},
pages = {171–178},
numpages = {8},
location = {Ho Chi Minh City, Vietnam},
series = {ICIIT '24}
}

@inproceedings{Nguyen2023,
  series = {ICIIT 2023},
  title = {Combining Autoencoder and Yolov6 Model for Classification and Disease Detection in Chickens},
  url = {http://dx.doi.org/10.1145/3591569.3591591},
  DOI = {10.1145/3591569.3591591},
  booktitle = {Proceedings of the 2023 8th International Conference on Intelligent Information Technology},
  publisher = {ACM},
  author = {Nguyen,  Khang Hoang and Nguyen,  Huynh Vu Nhu and Tran,  Hoang Ngoc and Quach,  Luyl-Da},
  year = {2023},
  month = Feb,
  pages = {132–138},
  collection = {ICIIT 2023}
}



\setcounter{table}{0}
\setcounter{figure}{0}
\appendix
\section{Appendix Section}

\begin{figure}[H]
    \centering
    \includegraphics[width=\linewidth]{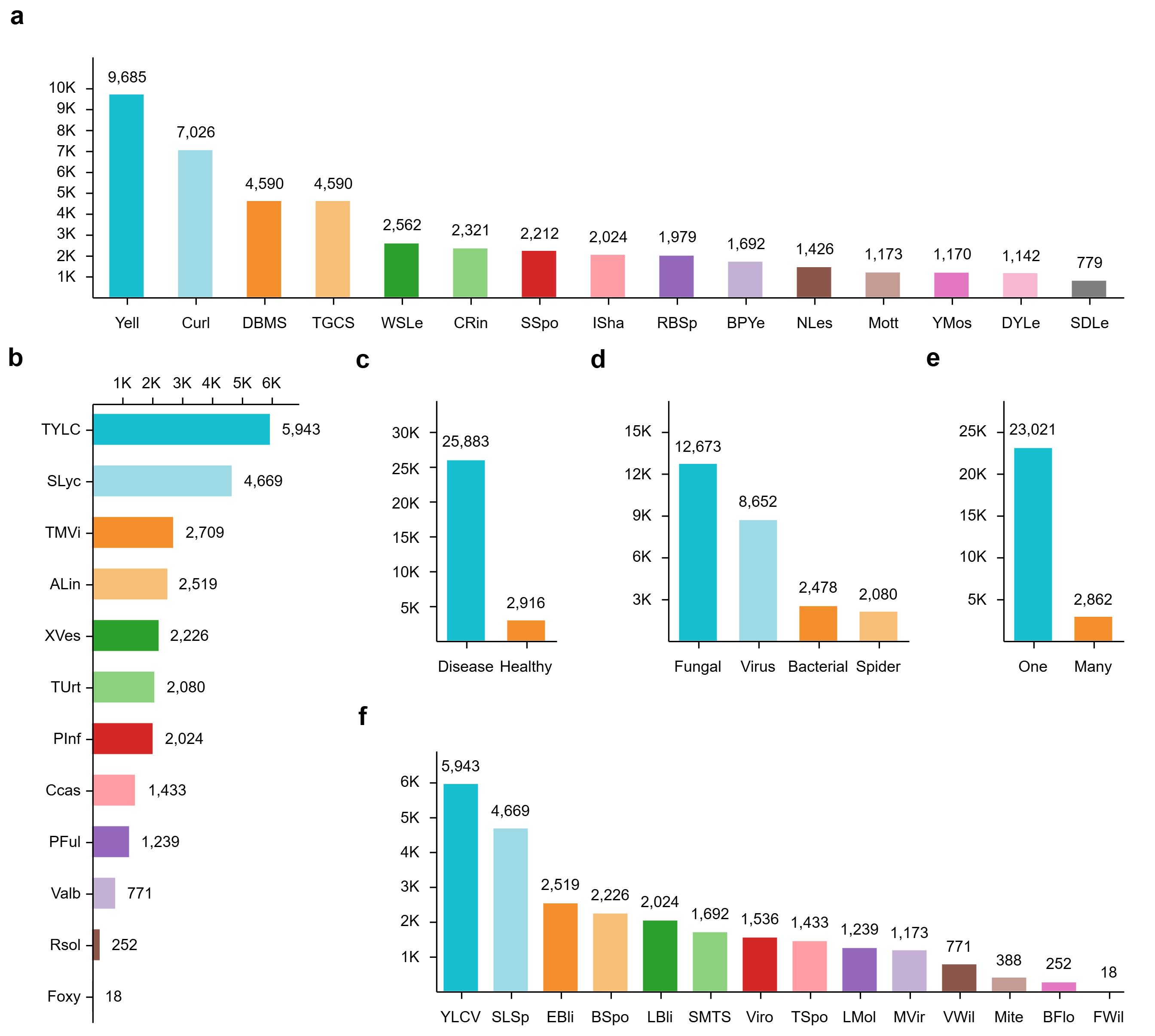}
    \caption{Class distribution of the dataset across the six VQA tasks. 
    (a) For the SI task, we show the top 15 classes with the highest image counts, where abbreviations are: Yell (yellowing), Curl (curling), DBMS (dark brown margin spot), TGCS (tan to gray center spot), WSLe (water-soaked lesion), CRin (concentric ring), SSpo (small spot), ISha (irregularly shaped), RBSp (reddish-brown spot), BPYe (blotched with pale yellow), NLes (necrotic lesion), Mott (mottling), YMos (yellow mosaic), DYLe (distortion of younger leaves), and SDLe (small dark lesion).
    (b) For the SNC task, abbreviations are: TYLC (Tomato yellow leaf curl virus), SLyc (Septoria lycopersici), TMVi (Tomato mosaic virus), ALin (Alternaria linariae), XVes (Xanthomonas vesicatoria), TUrt (Tetranychus urticae), PInf (Phytophthora infestans), Ccas (Corynespora cassiicola), PFul (Passalora fulva), Valb (Verticillium albo-atrum), Rsol (Ralstonia solanacearum), and Foxy (Fusarium oxysporum).
    (c) For the HDC task, classes are Disease and Healthy.
    (d) For the PC task, classes are Fungal, Virus, Bacterial, and Spider.
    (e) For the LC task, classes are One and Many.
    (f) For the DC task, abbreviations are: YLCV (Yellow Leaf Curl Virus), SLSp (Septoria Leaf Spot), EBli (Early Blight), BSpo (Bacterial Spot), LBli (Late Blight), SMTS (Spider mites Two-spotted Spider Mite), Viro (Virosis), TSpo (Target Spot), LMol (Leaf Mold), MVir (Mosaic Virus), VWil (Verticillium Wilt), Mite (Mite), BFlo (Bacterial Floundering), and FWil (Fusarium Wilt).
    }
    \label{fig:statistic_image}
\end{figure}


\begin{figure}[H]
\centering

\begin{tcolorbox}[promptbox={\faRobot\ \ SYSTEM\_PROMPT},
  colback=sysColor!12!white, colframe=sysColor, colbacktitle=sysColor,
  width=\linewidth]
You are an expert multimodal AI evaluator. Your task is to evaluate the accuracy of a candidate answer to a visual question, based on the provided ground truth answer set.
\end{tcolorbox}

\vspace{6pt}

\begin{tcolorbox}[promptbox={\faUser\ \ PROMPT\_TEMPLATE},
  colback=userColor!12!white, colframe=userColor, colbacktitle=userColor,
  width=\linewidth]

Ground Truth Reference Answers: \ph{ground\_truth\_answers}
Prediction Answer: \ph{prediction\_answer}

\medskip
Evaluate the candidate answer based on the following scale (1 to 5):
\begin{enumerate}\itemsep1.5pt
  \item The answer is completely incorrect, hallucinatory, or irrelevant to the question.
  \item The answer is partially relevant but contains major inaccuracies or misses the core question.
  \item The answer is generally correct but lacks specific details, or contains minor factual errors.
  \item The answer is correct, directly addresses the question, and is semantically equivalent to the ground truth.
  \item The answer is exceptionally accurate, clear, and demonstrates a strong understanding of the visual and textual context.
\end{enumerate}

\medskip
Respond with a score (\texttt{1}, \texttt{2}, \texttt{3}, \texttt{4}, or \texttt{5}).
\end{tcolorbox}

\caption{Prompt template used for LLM-as-a-Judge evaluation of visual question answering (VQA) responses on a five-point scale. The \textcolor{sysColor}{\textbf{system prompt}} (top) specifies the evaluator role and behavioral constraints; the \textcolor{userColor}{\textbf{user prompt template}} (bottom) supplies input placeholders (\ph{ground\_truth\_answers} and \ph{prediction\_answer}), the 5-point evaluation criteria, and restricts the output strictly to a single integer for automated parsing.}
\label{fig:llm_judge_prompt}
\end{figure}


\begin{table}[H]
\caption{Performance comparison of SOLAR against foundation models, vision models, and fine-tuned vision-language models across four tomato disease VQA datasets.}
\label{tab:Overall}
\centering
\resizebox{\textwidth}{!}{
\begin{tabular}{lcccccccccc}
\toprule
\multirow{2}{*}{\textbf{Model}} & \multicolumn{2}{c}{\textbf{TomaMMU}} & \multicolumn{2}{c}{\textbf{TLID}} & \multicolumn{2}{c}{\textbf{TLD-3}} & \multicolumn{2}{c}{\textbf{TomaAD}} & \multicolumn{2}{c}{\textbf{Average}} \\ \cmidrule(lr){2-3} \cmidrule(lr){4-5} \cmidrule(lr){6-7} \cmidrule(lr){8-9} \cmidrule(lr){10-11}
& Acc & F1 & Acc & F1 & Acc & F1 & Acc & F1 & Acc & F1 \\
\midrule
\rowcolor[HTML]{F4CCCC} \multicolumn{11}{c}{\textit{\textbf{Foundation Models}}} \\
\midrule
InternVL3 1B & 55.21 & 0.552 & 27.04 & 0.270 & 53.80 & 0.538 & 50.20 & 0.441 & 46.56 & 0.450 \\
InternVL3 2B & 56.20 & 0.562 & \textbf{38.49} & 0.385 & 55.01 & 0.550 & \textbf{59.99} & \textbf{0.541} & 52.42 & 0.510 \\
Qwen2.5 3B & 58.00 & 0.580 & 38.33 & 0.383 & 62.46 & 0.582 & 55.62 & 0.503 & 53.60 & 0.512 \\
Qwen3 2B & 58.87 & 0.589 & 33.02 & 0.330 & 58.01 & 0.579 & 43.87 & 0.389 & 48.44 & 0.472 \\
LLaVA-OV & 52.42 & 0.524 & 33.04 & 0.330 & 53.80 & 0.538 & 42.63 & 0.314 & 45.47 & 0.427 \\
\bottomrule
\end{tabular}
}
\end{table}

\begin{table}[H]
\centering
\caption{Performance comparison of SOLAR against foundation models, vision models, and fine-tuned vision-language models. The values represent the average results across four tomato disease VQA datasets categorized by question-based tasks.}
\label{tab:Overall3}
\resizebox{\textwidth}{!}{
\begin{tabular}{lcccccccccccccc}
\toprule
\multirow{2}{*}{\textbf{Model}} & \multicolumn{2}{c}{\textbf{HDC}} & \multicolumn{2}{c}{\textbf{DC}} & \multicolumn{2}{c}{\textbf{SNC}} & \multicolumn{2}{c}{\textbf{PC}} & \multicolumn{2}{c}{\textbf{SI}} & \multicolumn{2}{c}{\textbf{LC}} & \multicolumn{2}{c}{\textbf{Average}} \\ \cmidrule(lr){2-3} \cmidrule(lr){4-5} \cmidrule(lr){6-7} \cmidrule(lr){8-9} \cmidrule(lr){10-11} \cmidrule(lr){12-13} \cmidrule(lr){14-15}
& Acc & F1 & Acc & F1 & Acc & F1 & Acc & F1 & Acc & F1 & Acc & F1 & Acc & F1 \\
\midrule
\rowcolor[HTML]{F4CCCC} \multicolumn{15}{c}{\textit{\textbf{Foundation Models}}} \\
\midrule
InternVL3 1B & 63.76 & 0.550 & 24.71 & 0.247 & 34.85 & 0.349 & 37.38 & 0.373 & 28.25 & 0.280 & \textbf{90.45} & \textbf{0.904} & 46.56 & 0.450 \\
InternVL3 2B & 88.25 & 0.798 & 39.22 & 0.391 & 32.16 & 0.321 & 39.36 & 0.393 & 25.34 & 0.253 & 90.22 & 0.901 & 52.42 & 0.510 \\
Qwen2.5 3B & 86.41 & 0.817 & 36.80 & 0.358 & 37.52 & 0.373 & 42.71 & 0.361 & 29.86 & 0.298 & 88.32 & 0.867 & 53.60 & 0.512 \\
Qwen3 2B & 86.23 & 0.789 & 27.57 & 0.275 & 17.53 & 0.175 & 38.42 & 0.384 & 31.66 & 0.318 & 89.23 & 0.890 & 48.44 & 0.472 \\
LLaVA-OV & 75.54 & 0.668 & 26.10 & 0.243 & 24.49 & 0.210 & 32.12 & 0.310 & 27.44 & 0.262 & 87.17 & 0.868 & 45.47 & 0.427 \\
\bottomrule
\end{tabular}
}
\end{table}

\begin{table}[H]
\caption{Performance comparison of different components in the SOLAR architecture. The results show that SOLAR outperforms other configurations, showing that all components are important.}
\label{tab:TableComponets}
\centering
\resizebox{\textwidth}{!}{%
\begin{tabular}{lcccccccccc}
\hline
& \multicolumn{2}{c}{TomaMMU} & \multicolumn{2}{c}{TLID} & \multicolumn{2}{c}{TLD-3} & \multicolumn{2}{c}{TomaAD} & \multicolumn{2}{c}{Average} \\
\cmidrule(lr){2-3} \cmidrule(lr){4-5} \cmidrule(lr){6-7} \cmidrule(lr){8-9} \cmidrule(lr){10-11}
\multirow{-2.5}{*}{\textbf{Model}} & Acc & F1 & Acc & F1 & Acc & F1 & Acc & F1 & Acc & F1 \\
\hline
$W/o$ Fusion Adapter & 97.20          & 0.969          & \textbf{34.05} & \textbf{0.337} & \textbf{92.03} & \textbf{0.859} & 51.65          & 0.490          & 68.73          & 0.664          \\
$W/o$ MoE            & \textbf{98.04} & \textbf{0.977} & 33.44          & 0.330          & 90.56          & 0.846          & 42.75          & 0.416          & 66.20          & 0.642          \\
\rowcolor[HTML]{E8F4FC}
SOLAR      & 97.68          & 0.974          & 33.70          & 0.335          & 91.67          & 0.855          & \textbf{53.20} & \textbf{0.501} & \textbf{69.06} & \textbf{0.666} \\
\hline
\end{tabular}%
}
\end{table}

\begin{table}[H]
\caption{Performance comparison of different vision encoders across four datasets in the SOLAR architecture}
\label{tab:TableVisionEncoder}

\centering
\resizebox{\textwidth}{!}{%
\begin{tabular}{lcccccccccc}
\hline
& \multicolumn{2}{c}{TomaMMU} & \multicolumn{2}{c}{TLID} & \multicolumn{2}{c}{TLD-3} & \multicolumn{2}{c}{TomaAD} & \multicolumn{2}{c}{Average} \\
\cmidrule(lr){2-3} \cmidrule(lr){4-5} \cmidrule(lr){6-7} \cmidrule(lr){8-9} \cmidrule(lr){10-11}
\multirow{-2.5}{*}{\textbf{Model}} & Acc & F1 & Acc & F1 & Acc & F1 & Acc & F1 & Acc & F1 \\
\hline
CLIP     & 96.53          & 0.958          & 38.17          & 0.373          & 86.80          & 0.822          & 40.95          & 0.403          & 65.61          & 0.639          \\
BioCLIP  & 97.21          & \textbf{0.976} & 32.96          & 0.329          & 85.21          & 0.852          & 35.65          & 0.356          & 62.76          & 0.628          \\
AgriCLIP & 96.94          & 0.967          & \textbf{46.13} & \textbf{0.451} & 85.09          & 0.798          & 31.99          & 0.318          & 65.04          & 0.634          \\
\rowcolor[HTML]{E8F4FC}
SCOLD    & \textbf{97.68} & 0.974          & 33.70          & 0.335          & \textbf{91.67} & \textbf{0.855} & \textbf{53.20} & \textbf{0.501} & \textbf{69.06} & \textbf{0.666} \\
\hline
\end{tabular}%
}

\end{table}

\begin{table}[H]
\caption{Performance comparison of different LLM backbones across four datasets in the SOLAR architecture}
\label{tab:TableLLM}
\centering
\resizebox{\textwidth}{!}{%
\begin{tabular}{lcccccccccc}
\hline
& \multicolumn{2}{c}{TomaMMU} & \multicolumn{2}{c}{TLID} & \multicolumn{2}{c}{TLD-3} & \multicolumn{2}{c}{TomaAD} & \multicolumn{2}{c}{Average} \\
\cmidrule(lr){2-3} \cmidrule(lr){4-5} \cmidrule(lr){6-7} \cmidrule(lr){8-9} \cmidrule(lr){10-11}
\multirow{-2.5}{*}{\textbf{Model}} & Acc & F1 & Acc  & F1 & Acc & F1 & Acc & F1 & Acc  & F1 \\
\hline
Qwen 3            & 97.24          & \textbf{0.978} & \textbf{34.25} & \textbf{0.343} & 89.42          & 0.855          & 47.38          & 0.454          & 67.07          & 0.658          \\
BioGPT            & \textbf{97.76} & 0.974          & 32.00          & 0.318          & \textbf{92.09} & \textbf{0.861} & 48.20          & 0.452          & 67.51          & 0.651          \\
\rowcolor[HTML]{E8F4FC}
LFM-VL 2.5 & 97.68          & 0.974          & 33.70          & 0.335          & 91.67          & 0.855          & \textbf{53.20} & \textbf{0.501} & \textbf{69.06} & \textbf{0.666} \\
\hline
\end{tabular}%
}
\end{table}

\begin{table}[H]
\caption{Performance comparison across six VQA tasks on the \textbf{TomaMMU} dataset.}
\label{tab:detailed_tomammu}
\centering
\resizebox{\textwidth}{!}{
\begin{tabular}{lcccccccccccccc}
\toprule
\multirow{2}{*}{\textbf{Model}} & \multicolumn{2}{c}{\textbf{HDC}} & \multicolumn{2}{c}{\textbf{DC}} & \multicolumn{2}{c}{\textbf{SNC}} & \multicolumn{2}{c}{\textbf{PC}} & \multicolumn{2}{c}{\textbf{SI}} & \multicolumn{2}{c}{\textbf{LC}} & \multicolumn{2}{c}{\textbf{Average}} \\ \cmidrule(lr){2-3} \cmidrule(lr){4-5} \cmidrule(lr){6-7} \cmidrule(lr){8-9} \cmidrule(lr){10-11} \cmidrule(lr){12-13} \cmidrule(lr){14-15}
 & Acc & F1 & Acc & F1 & Acc & F1 & Acc & F1 & Acc & F1 & Acc & F1 & Acc & F1 \\ \midrule
\rowcolor[HTML]{F4CCCC} \multicolumn{15}{c}{\textit{\textbf{Foundation Models}}} \\
\midrule
InternVL3 1B & 89.88 & 0.899 & 28.61 & 0.286 & 30.69 & 0.307 & 50.82 & 0.508 & 37.54 & 0.375 & 93.73 & 0.937 & 55.21 & 0.552 \\
InternVL3 2B & 82.08 & 0.821 & 34.23 & 0.342 & 42.65 & 0.426 & 51.44 & 0.514 & 32.42 & 0.324 & 94.38 & 0.944 & 56.20 & 0.562 \\
Qwen2.5 3B & 83.84 & 0.838 & 40.60 & 0.406 & 47.46 & 0.475 & 51.01 & 0.510 & 32.46 & 0.325 & 92.64 & 0.926 & 58.00 & 0.580 \\
Qwen3 2B & 91.55 & 0.915 & 45.55 & 0.456 & 32.63 & 0.326 & 55.88 & 0.559 & 33.94 & 0.339 & 93.68 & 0.937 & 58.87 & 0.589 \\
LLaVA-OV & 89.88 & 0.899 & 34.75 & 0.347 & 24.82 & 0.248 & 45.16 & 0.452 & 26.69 & 0.267 & 93.23 & 0.932 & 52.42 & 0.524 \\
\bottomrule
\end{tabular}
}
\end{table}

\begin{table}[H]
\caption{Performance comparison across six VQA tasks on the \textbf{TLID} dataset.}
\label{tab:detailed_tlid}
\centering
\resizebox{\textwidth}{!}{
\begin{tabular}{lcccccccccccccc}
\toprule
\multirow{2}{*}{\textbf{Model}} & \multicolumn{2}{c}{\textbf{HDC}} & \multicolumn{2}{c}{\textbf{DC}} & \multicolumn{2}{c}{\textbf{SNC}} & \multicolumn{2}{c}{\textbf{PC}} & \multicolumn{2}{c}{\textbf{SI}} & \multicolumn{2}{c}{\textbf{LC}} & \multicolumn{2}{c}{\textbf{Average}} \\ \cmidrule(lr){2-3} \cmidrule(lr){4-5} \cmidrule(lr){6-7} \cmidrule(lr){8-9} \cmidrule(lr){10-11} \cmidrule(lr){12-13} \cmidrule(lr){14-15}
 & Acc & F1 & Acc & F1 & Acc & F1 & Acc & F1 & Acc & F1 & Acc & F1 & Acc & F1 \\ \midrule
\rowcolor[HTML]{F4CCCC} \multicolumn{15}{c}{\textit{\textbf{Foundation Models}}} \\
\midrule
InternVL3 1B & 0.00 & 0.000 & 0.00 & 0.000 & \textbf{42.67} & \textbf{0.427} & 1.84 & 0.018 & 17.97 & 0.180 & 99.78 & 0.998 & 27.04 & 0.270 \\
InternVL3 2B & \textbf{95.44} & \textbf{0.954} & 25.85 & 0.259 & 4.30 & 0.043 & 0.04 & 0.000 & 5.34 & 0.053 & 99.98 & \textbf{1.000} & \textbf{38.49} & 0.385 \\
Qwen2.5 3B & 93.04 & 0.930 & 27.35 & 0.274 & 2.63 & 0.026 & 0.21 & 0.002 & 6.98 & 0.070 & 99.78 & 0.998 & 38.33 & 0.383 \\
Qwen3 2B & 81.41 & 0.814 & 10.44 & 0.104 & 0.71 & 0.007 & 0.08 & 0.001 & 5.57 & 0.056 & 99.94 & 0.999 & 33.02 & 0.330 \\
LLaVA-OV & 47.12 & 0.471 & 14.95 & 0.149 & 24.30 & 0.243 & 0.00 & 0.000 & 12.01 & 0.120 & 99.88 & 0.999 & 33.04 & 0.330 \\
\bottomrule
\end{tabular}
}
\end{table}

\begin{table}[H]
\caption{Performance comparison across six VQA tasks on the \textbf{TLD-3} dataset.}
\label{tab:detailed_tld_3}
\centering
\resizebox{\textwidth}{!}{
\begin{tabular}{lcccccccccccccc}
\toprule
\multirow{2}{*}{\textbf{Model}} & \multicolumn{2}{c}{\textbf{HDC}} & \multicolumn{2}{c}{\textbf{DC}} & \multicolumn{2}{c}{\textbf{SNC}} & \multicolumn{2}{c}{\textbf{PC}} & \multicolumn{2}{c}{\textbf{SI}} & \multicolumn{2}{c}{\textbf{LC}} & \multicolumn{2}{c}{\textbf{Average}} \\ \cmidrule(lr){2-3} \cmidrule(lr){4-5} \cmidrule(lr){6-7} \cmidrule(lr){8-9} \cmidrule(lr){10-11} \cmidrule(lr){12-13} \cmidrule(lr){14-15}
 & Acc & F1 & Acc & F1 & Acc & F1 & Acc & F1 & Acc & F1 & Acc & F1 & Acc & F1 \\ \midrule
\rowcolor[HTML]{F4CCCC} \multicolumn{15}{c}{\textit{\textbf{Foundation Models}}} \\
\midrule
InternVL3 1B & 85.89 & 0.859 & 27.25 & 0.273 & 20.07 & 0.201 & 67.46 & 0.675 & 25.74 & 0.257 & \textbf{96.41} & \textbf{0.964} & 53.80 & 0.538 \\
InternVL3 2B & 84.83 & 0.848 & 23.16 & 0.232 & 31.70 & 0.317 & 68.01 & 0.680 & 26.47 & 0.265 & 95.91 & 0.959 & 55.01 & 0.550 \\
Qwen2.5 3B & 85.73 & 0.857 & 36.16 & 0.362 & 38.53 & 0.385 & 95.18 & 0.697 & 23.96 & 0.240 & 95.18 & 0.952 & 62.46 & 0.582 \\
Qwen3 2B & 91.29 & 0.912 & 34.20 & 0.336 & 32.52 & 0.325 & 69.58 & 0.696 & 24.22 & 0.242 & 96.22 & 0.962 & 58.01 & 0.579 \\
LLaVA-OV & 85.87 & 0.858 & 35.59 & 0.356 & 25.22 & 0.252 & 56.93 & 0.569 & 25.19 & 0.252 & 94.00 & 0.940 & 53.80 & 0.538 \\
\bottomrule
\end{tabular}
}
\end{table}

\begin{table}[H]
\caption{Performance comparison across six VQA tasks on the \textbf{TomaAD} dataset.}
\label{tab:detailed_tomaad}
\centering
\resizebox{\textwidth}{!}{
\begin{tabular}{lcccccccccccccc}
\toprule
\multirow{2}{*}{\textbf{Model}} & \multicolumn{2}{c}{\textbf{HDC}} & \multicolumn{2}{c}{\textbf{DC}} & \multicolumn{2}{c}{\textbf{SNC}} & \multicolumn{2}{c}{\textbf{PC}} & \multicolumn{2}{c}{\textbf{SI}} & \multicolumn{2}{c}{\textbf{LC}} & \multicolumn{2}{c}{\textbf{Average}} \\ \cmidrule(lr){2-3} \cmidrule(lr){4-5} \cmidrule(lr){6-7} \cmidrule(lr){8-9} \cmidrule(lr){10-11} \cmidrule(lr){12-13} \cmidrule(lr){14-15}
 & Acc & F1 & Acc & F1 & Acc & F1 & Acc & F1 & Acc & F1 & Acc & F1 & Acc & F1 \\ \midrule
\rowcolor[HTML]{F4CCCC} \multicolumn{15}{c}{\textit{\textbf{Foundation Models}}} \\
\midrule
InternVL3 1B & 79.28 & 0.442 & 42.96 & 0.429 & 45.98 & 0.459 & 29.40 & 0.290 & 31.74 & 0.306 & \textbf{71.86} & \textbf{0.718} & 50.20 & 0.441 \\
InternVL3 2B & \textbf{90.64} & 0.570 & \textbf{73.62} & \textbf{0.732} & 50.00 & 0.499 & 37.94 & 0.377 & 37.14 & 0.368 & 70.60 & 0.701 & \textbf{59.99} & \textbf{0.541} \\
Qwen2.5 3B & 83.03 & 0.641 & 43.07 & 0.390 & 61.46 & 0.605 & 24.43 & 0.235 & 56.04 & 0.557 & 65.67 & 0.592 & 55.62 & 0.503 \\
Qwen3 2B & 80.68 & 0.516 & 20.10 & 0.202 & 4.27 & 0.043 & 28.14 & 0.280 & 62.92 & 0.633 & 67.09 & 0.660 & 43.87 & 0.389 \\
LLaVA-OV & 79.28 & 0.442 & 19.10 & 0.120 & 23.62 & 0.096 & 26.38 & 0.218 & 45.85 & 0.410 & 61.56 & 0.600 & 42.63 & 0.314 \\
\bottomrule
\end{tabular}
}
\end{table}

\begin{table}[H]
\caption{Performance comparison ($ROUGE_L$ and $GPT$) across six VQA tasks on the \textbf{TomaMMU} dataset.}
\label{tab:detailed_gen_tomammu}
\centering
\resizebox{\textwidth}{!}{
\begin{tabular}{lcccccccccccccc}
\toprule
\multirow{2}{*}{\textbf{Model}} & \multicolumn{2}{c}{\textbf{HDC}} & \multicolumn{2}{c}{\textbf{DC}} & \multicolumn{2}{c}{\textbf{SNC}} & \multicolumn{2}{c}{\textbf{PC}} & \multicolumn{2}{c}{\textbf{SI}} & \multicolumn{2}{c}{\textbf{LC}} & \multicolumn{2}{c}{\textbf{Average}} \\ \cmidrule(lr){2-3} \cmidrule(lr){4-5} \cmidrule(lr){6-7} \cmidrule(lr){8-9} \cmidrule(lr){10-11} \cmidrule(lr){12-13} \cmidrule(lr){14-15}
 & $ROUGE_L$ & $GPT$ & $ROUGE_L$ & $GPT$ & $ROUGE_L$ & $GPT$ & $ROUGE_L$ & $GPT$ & $ROUGE_L$ & $GPT$ & $ROUGE_L$ & $GPT$ & $ROUGE_L$ & $GPT$ \\ \midrule
InternVL3 1B & 0.987 & 4.595 & 0.871 & 2.474 & 0.840 & 2.230 & 0.965 & 3.033 & 0.897 & 2.782 & 0.993 & 4.749 & 0.926 & 3.311 \\
InternVL3 2B & 0.978 & 4.285 & 0.883 & 2.739 & 0.876 & 2.726 & 0.965 & 3.058 & 0.889 & 2.634 & 0.994 & 4.775 & 0.931 & 3.370 \\
Qwen2.5 3B   & 0.980 & 4.354 & 0.875 & 2.630 & 0.885 & 2.937 & 0.965 & 3.041 & 0.888 & 2.620 & 0.992 & 4.709 & 0.931 & 3.382 \\
Qwen3 2B     & 0.990 & \textbf{4.684} & 0.895 & 2.923 & 0.852 & 2.311 & 0.969 & 3.235 & 0.891 & 2.676 & 0.993 & 4.747 & 0.932 & 3.429 \\
LLaVA-OV     & 0.987 & 4.595 & 0.866 & 2.413 & 0.837 & 2.014 & 0.961 & 2.807 & 0.883 & 2.358 & 0.993 & 4.729 & 0.921 & 3.153 \\
LFM-VL 2.5   & 0.989 & 4.637 & 0.905 & 3.089 & 0.862 & 2.530 & 0.967 & 3.162 & 0.891 & 2.621 & 0.994 & 4.789 & 0.935 & 3.471 \\
\midrule
\rowcolor[HTML]{E8F4FC} \textbf{SOLAR} & \textbf{0.998} & 4.054 & \textbf{0.997} & \textbf{4.858} & \textbf{0.997} & \textbf{4.936} & \textbf{1.000} & \textbf{4.980} & \textbf{0.976} & \textbf{4.653} & \textbf{0.999} & \textbf{4.963} & \textbf{0.995} & \textbf{4.741} \\
\bottomrule
\end{tabular}
}
\end{table}

\begin{table}[H]
\caption{Performance comparison ($ROUGE_L$ and $GPT$) across six VQA tasks on the \textbf{TLID} dataset.}
\label{tab:detailed_gen_tlid}
\centering
\resizebox{\textwidth}{!}{
\begin{tabular}{lcccccccccccccc}
\toprule
\multirow{2}{*}{\textbf{Model}} & \multicolumn{2}{c}{\textbf{HDC}} & \multicolumn{2}{c}{\textbf{DC}} & \multicolumn{2}{c}{\textbf{SNC}} & \multicolumn{2}{c}{\textbf{PC}} & \multicolumn{2}{c}{\textbf{SI}} & \multicolumn{2}{c}{\textbf{LC}} & \multicolumn{2}{c}{\textbf{Average}} \\ \cmidrule(lr){2-3} \cmidrule(lr){4-5} \cmidrule(lr){6-7} \cmidrule(lr){8-9} \cmidrule(lr){10-11} \cmidrule(lr){12-13} \cmidrule(lr){14-15}
 & $ROUGE_L$ & $GPT$ & $ROUGE_L$ & $GPT$ & $ROUGE_L$ & $GPT$ & $ROUGE_L$ & $GPT$ & $ROUGE_L$ & $GPT$ & $ROUGE_L$ & $GPT$ & $ROUGE_L$ & $GPT$ \\ \midrule
InternVL3 1B & \textbf{0.996} & \textbf{4.877} & 0.819 & 1.634 & \textbf{0.841} & 1.005 & 0.927 & 1.042 & \textbf{0.872} & 1.377 & 0.995 & 4.816 & 0.908 & 2.459 \\
InternVL3 2B & 0.995 & 4.844 & 0.890 & 2.155 & 0.821 & 1.712 & 0.962 & 3.235 & 0.847 & \textbf{3.008} & \textbf{0.999} & 4.980 & 0.919 & \textbf{3.319} \\
Qwen2.5 3B   & 0.993 & 4.767 & 0.873 & 2.064 & 0.803 & 1.890 & 0.929 & 1.000 & 0.862 & 1.980 & 0.994 & 4.798 & 0.909 & 2.750 \\
Qwen3 2B     & 0.995 & 4.829 & \textbf{0.917} & \textbf{2.393} & 0.832 & 1.096 & \textbf{0.976} & \textbf{3.777} & 0.865 & 1.742 & 0.995 & 4.828 & \textbf{0.930} & 3.111 \\
LLaVA-OV     & 0.941 & 3.116 & 0.891 & 1.896 & 0.838 & \textbf{1.901} & 0.929 & 1.027 & 0.852 & 2.194 & 0.989 & \textbf{5.000} & 0.906 & 2.522 \\
LFM-VL 2.5   & 0.941 & 3.116 & 0.822 & 2.006 & 0.828 & 1.060 & 0.926 & 1.095 & 0.839 & 1.973 & 0.949 & 4.381 & 0.884 & 2.272 \\
\midrule
\rowcolor[HTML]{E8F4FC} \textbf{SOLAR} & 0.994 & 4.807 & 0.842 & 1.581 & 0.825 & 1.027 & 0.930 & 1.087 & 0.772 & 2.433 & \textbf{0.999} & 4.957 & 0.894 & 2.649 \\
\bottomrule
\end{tabular}
}
\end{table}

\begin{table}[H]
\caption{Performance comparison ($ROUGE_L$ and $GPT$) across six VQA tasks on the \textbf{TLD-3} dataset.}
\label{tab:detailed_gen_tld_3}
\centering
\resizebox{\textwidth}{!}{
\begin{tabular}{lcccccccccccccc}
\toprule
\multirow{2}{*}{\textbf{Model}} & \multicolumn{2}{c}{\textbf{HDC}} & \multicolumn{2}{c}{\textbf{DC}} & \multicolumn{2}{c}{\textbf{SNC}} & \multicolumn{2}{c}{\textbf{PC}} & \multicolumn{2}{c}{\textbf{SI}} & \multicolumn{2}{c}{\textbf{LC}} & \multicolumn{2}{c}{\textbf{Average}} \\ \cmidrule(lr){2-3} \cmidrule(lr){4-5} \cmidrule(lr){6-7} \cmidrule(lr){8-9} \cmidrule(lr){10-11} \cmidrule(lr){12-13} \cmidrule(lr){14-15}
 & $ROUGE_L$ & $GPT$ & $ROUGE_L$ & $GPT$ & $ROUGE_L$ & $GPT$ & $ROUGE_L$ & $GPT$ & $ROUGE_L$ & $GPT$ & $ROUGE_L$ & $GPT$ & $ROUGE_L$ & $GPT$ \\ \midrule
InternVL3 1B & 0.987 & 4.595 & 0.871 & 2.474 & 0.840 & 2.230 & 0.965 & 3.033 & \textbf{0.897} & 2.782 & 0.993 & 4.749 & 0.926 & 3.311 \\
InternVL3 2B & 0.978 & 4.285 & 0.883 & 2.739 & 0.876 & 2.726 & 0.965 & 3.058 & 0.889 & 2.634 & \textbf{0.994} & 4.775 & 0.931 & 3.370 \\
Qwen2.5 3B   & 0.980 & 4.354 & 0.875 & 2.630 & 0.885 & 2.937 & 0.965 & 3.041 & 0.888 & 2.620 & 0.992 & 4.709 & 0.931 & 3.382 \\
Qwen3 2B     & 0.990 & 4.684 & 0.895 & 2.923 & 0.852 & 2.311 & 0.969 & 3.235 & 0.891 & 2.676 & 0.993 & 4.747 & 0.932 & 3.429 \\
LLaVA-OV     & 0.987 & 4.595 & 0.866 & 2.413 & 0.837 & 2.014 & 0.961 & 2.807 & 0.883 & 2.358 & 0.993 & 4.729 & 0.921 & 3.153 \\
LFM-VL 2.5   & 0.989 & 4.637 & 0.905 & 3.089 & 0.862 & 2.530 & 0.967 & 3.162 & 0.891 & 2.621 & \textbf{0.994} & \textbf{4.789} & 0.935 & 3.471 \\
\midrule
\rowcolor[HTML]{E8F4FC} \textbf{SOLAR} & \textbf{0.996} & \textbf{4.874} & \textbf{0.979} & \textbf{4.596} & \textbf{0.985} & \textbf{4.658} & \textbf{0.998} & \textbf{4.872} & 0.886 & \textbf{3.511} & \textbf{0.994} & 4.785 & \textbf{0.973} & \textbf{4.549} \\
\bottomrule
\end{tabular}
}
\end{table}

\begin{table}[H]
\caption{Performance comparison ($ROUGE_L$ and $GPT$) across six VQA tasks on the \textbf{TomaAD} dataset.}
\label{tab:detailed_gen_tomaad}
\centering
\resizebox{\textwidth}{!}{
\begin{tabular}{lcccccccccccccc}
\toprule
\multirow{2}{*}{\textbf{Model}} & \multicolumn{2}{c}{\textbf{HDC}} & \multicolumn{2}{c}{\textbf{DC}} & \multicolumn{2}{c}{\textbf{SNC}} & \multicolumn{2}{c}{\textbf{PC}} & \multicolumn{2}{c}{\textbf{SI}} & \multicolumn{2}{c}{\textbf{LC}} & \multicolumn{2}{c}{\textbf{Average}} \\ \cmidrule(lr){2-3} \cmidrule(lr){4-5} \cmidrule(lr){6-7} \cmidrule(lr){8-9} \cmidrule(lr){10-11} \cmidrule(lr){12-13} \cmidrule(lr){14-15}
 & $ROUGE_L$ & $GPT$ & $ROUGE_L$ & $GPT$ & $ROUGE_L$ & $GPT$ & $ROUGE_L$ & $GPT$ & $ROUGE_L$ & $GPT$ & $ROUGE_L$ & $GPT$ & $ROUGE_L$ & $GPT$ \\ \midrule
InternVL3 1B & 0.985 & 4.505 & 0.825 & 1.524 & 0.791 & 1.249 & 0.928 & 1.771 & \textbf{0.886} & 1.785 & 0.915 & 2.014 & 0.888 & 2.141 \\
InternVL3 2B & \textbf{0.988} & \textbf{4.609} & 0.823 & 1.947 & 0.783 & 1.191 & 0.940 & 1.741 & 0.868 & 2.710 & 0.939 & 2.900 & 0.890 & 2.516 \\
Qwen2.5 3B   & 0.981 & 4.385 & 0.804 & 1.794 & 0.769 & 1.106 & 0.930 & 1.337 & 0.876 & 2.265 & 0.932 & 2.625 & 0.882 & 2.252 \\
Qwen3 2B     & 0.982 & 4.425 & 0.859 & 1.723 & 0.798 & 1.287 & 0.927 & 2.096 & 0.875 & 2.013 & 0.915 & 2.062 & 0.893 & 2.268 \\
LLaVA-OV     & 0.901 & 1.830 & 0.842 & 2.063 & 0.791 & 1.806 & 0.936 & 1.685 & 0.880 & 1.973 & \textbf{0.961} & \textbf{4.723} & 0.885 & 2.347 \\
LFM-VL 2.5   & 0.901 & 1.827 & 0.823 & 1.559 & 0.806 & 1.642 & 0.909 & 1.356 & 0.816 & 1.994 & 0.915 & 2.634 & 0.862 & 1.835 \\
\midrule
\rowcolor[HTML]{E8F4FC} \textbf{SOLAR} & 0.980 & 4.345 & \textbf{0.903} & \textbf{3.098} & \textbf{0.911} & \textbf{3.282} & \textbf{0.970} & \textbf{3.390} & 0.817 & \textbf{3.156} & 0.937 & 2.732 & \textbf{0.920} & \textbf{3.334} \\
\bottomrule
\end{tabular}
}
\end{table}

\begin{table}[H]
\caption{Performance comparison of different vision encoders across six VQA tasks on the \textbf{TomaMMU} dataset.}
\label{tab:detailed_vision_encoder_tomammu}
\centering
\resizebox{\textwidth}{!}{
\begin{tabular}{lcccccccccccccc}
\toprule
\multirow{2}{*}{\textbf{Model}} & \multicolumn{2}{c}{\textbf{HDC}} & \multicolumn{2}{c}{\textbf{DC}} & \multicolumn{2}{c}{\textbf{SNC}} & \multicolumn{2}{c}{\textbf{PC}} & \multicolumn{2}{c}{\textbf{SI}} & \multicolumn{2}{c}{\textbf{LC}} & \multicolumn{2}{c}{\textbf{Average}} \\ \cmidrule(lr){2-3} \cmidrule(lr){4-5} \cmidrule(lr){6-7} \cmidrule(lr){8-9} \cmidrule(lr){10-11} \cmidrule(lr){12-13} \cmidrule(lr){14-15}
 & Acc & F1 & Acc & F1 & Acc & F1 & Acc & F1 & Acc & F1 & Acc & F1 & Acc & F1 \\ \midrule
CLIP & 98.59 & 0.986 & 96.97 & 0.970 & 96.45 & 0.965 & 98.73 & 0.987 & 89.35 & 0.849 & 99.11 & 0.991 & 96.53 & 0.958 \\
BioCLIP & 98.59 & 0.986 & 98.01 & 0.980 & 97.08 & 0.971 & 98.76 & 0.988 & 91.61 & \textbf{0.940} & 99.19 & \textbf{0.992} & 97.21 & \textbf{0.976} \\
AgriCLIP & 98.68 & 0.987 & 98.01 & 0.979 & 96.16 & 0.962 & 98.67 & 0.987 & 90.86 & 0.897 & \textbf{99.23} & \textbf{0.992} & 96.94 & 0.967 \\
\rowcolor[HTML]{E8F4FC} \textbf{SCOLD} & \textbf{98.75} & \textbf{0.988} & \textbf{98.24} & \textbf{0.982} & \textbf{98.40} & \textbf{0.984} & \textbf{99.50} & \textbf{0.995} & \textbf{92.15} & 0.903 & 99.07 & 0.991 & \textbf{97.68} & 0.974 \\
\bottomrule
\end{tabular}
}

\end{table}

\begin{table}[H]
\caption{Performance comparison of different vision encoders across six VQA tasks on the \textbf{TLID} dataset.}
\label{tab:detailed_vision_encoder_tlid}
\centering
\resizebox{\textwidth}{!}{
\begin{tabular}{lcccccccccccccc}
\toprule
\multirow{2}{*}{\textbf{Model}} & \multicolumn{2}{c}{\textbf{HDC}} & \multicolumn{2}{c}{\textbf{DC}} & \multicolumn{2}{c}{\textbf{SNC}} & \multicolumn{2}{c}{\textbf{PC}} & \multicolumn{2}{c}{\textbf{SI}} & \multicolumn{2}{c}{\textbf{LC}} & \multicolumn{2}{c}{\textbf{Average}} \\ \cmidrule(lr){2-3} \cmidrule(lr){4-5} \cmidrule(lr){6-7} \cmidrule(lr){8-9} \cmidrule(lr){10-11} \cmidrule(lr){12-13} \cmidrule(lr){14-15}
 & Acc & F1 & Acc & F1 & Acc & F1 & Acc & F1 & Acc & F1 & Acc & F1 & Acc & F1 \\ \midrule
CLIP & 93.74 & 0.937 & 8.73 & 0.087 & 7.85 & 0.079 & 4.05 & 0.041 & 14.69 & 0.094 & \textbf{99.94} & \textbf{0.999} & 38.17 & 0.373 \\
BioCLIP & \textbf{96.10} & \textbf{0.961} & 0.08 & 0.001 & 0.04 & 0.000 & 0.92 & 0.009 & 3.53 & 0.035 & 97.07 & 0.971 & 32.96 & 0.329 \\
AgriCLIP & 81.68 & 0.817 & \textbf{16.41} & \textbf{0.164} & \textbf{29.81} & \textbf{0.298} & \textbf{30.44} & \textbf{0.304} & \textbf{19.33} & \textbf{0.131} & 99.13 & 0.991 & \textbf{46.13} & \textbf{0.451} \\
\rowcolor[HTML]{E8F4FC} \textbf{SCOLD} & 95.18 & 0.952 & 0.79 & 0.008 & 0.67 & 0.007 & 2.17 & 0.022 & 4.49 & 0.032 & 98.92 & 0.989 & 33.70 & 0.335 \\
\bottomrule
\end{tabular}
}

\end{table}

\begin{table}[H]
\caption{Performance comparison of different vision encoders across six VQA tasks on the \textbf{TLD-3} dataset.}
\label{tab:detailed_vision_encoder_tld_3}
\centering
\resizebox{\textwidth}{!}{
\begin{tabular}{lcccccccccccccc}
\toprule
\multirow{2}{*}{\textbf{Model}} & \multicolumn{2}{c}{\textbf{HDC}} & \multicolumn{2}{c}{\textbf{DC}} & \multicolumn{2}{c}{\textbf{SNC}} & \multicolumn{2}{c}{\textbf{PC}} & \multicolumn{2}{c}{\textbf{SI}} & \multicolumn{2}{c}{\textbf{LC}} & \multicolumn{2}{c}{\textbf{Average}} \\ \cmidrule(lr){2-3} \cmidrule(lr){4-5} \cmidrule(lr){6-7} \cmidrule(lr){8-9} \cmidrule(lr){10-11} \cmidrule(lr){12-13} \cmidrule(lr){14-15}
 & Acc & F1 & Acc & F1 & Acc & F1 & Acc & F1 & Acc & F1 & Acc & F1 & Acc & F1 \\ \midrule
CLIP & 96.10 & 0.961 & 84.01 & \textbf{0.840} & 81.97 & 0.820 & 91.68 & 0.917 & 71.01 & 0.435 & \textbf{96.00} & \textbf{0.960} & 86.80 & 0.822 \\
BioCLIP & \textbf{97.79} & \textbf{0.978} & 80.13 & 0.801 & 80.83 & 0.808 & 91.74 & 0.917 & 68.51 & \textbf{0.685} & 92.24 & 0.922 & 85.21 & 0.852 \\
AgriCLIP & 95.49 & 0.955 & 80.09 & 0.731 & 76.76 & 0.768 & 92.89 & 0.929 & 69.52 & 0.447 & 95.78 & 0.958 & 85.09 & 0.798 \\
\rowcolor[HTML]{E8F4FC} \textbf{SCOLD} & 96.86 & 0.969 & \textbf{91.19} & 0.823 & \textbf{91.35} & \textbf{0.913} & \textbf{96.79} & \textbf{0.968} & \textbf{79.22} & 0.510 & 94.62 & 0.946 & \textbf{91.67} & \textbf{0.855} \\
\bottomrule
\end{tabular}
}

\end{table}

\begin{table}[H]
\caption{Performance comparison of different vision encoders across six VQA tasks on the \textbf{TomaAD} dataset.}
\label{tab:detailed_vision_encoder_tomaad}
\centering
\resizebox{\textwidth}{!}{
\begin{tabular}{lcccccccccccccc}
\toprule
\multirow{2}{*}{\textbf{Model}} & \multicolumn{2}{c}{\textbf{HDC}} & \multicolumn{2}{c}{\textbf{DC}} & \multicolumn{2}{c}{\textbf{SNC}} & \multicolumn{2}{c}{\textbf{PC}} & \multicolumn{2}{c}{\textbf{SI}} & \multicolumn{2}{c}{\textbf{LC}} & \multicolumn{2}{c}{\textbf{Average}} \\ \cmidrule(lr){2-3} \cmidrule(lr){4-5} \cmidrule(lr){6-7} \cmidrule(lr){8-9} \cmidrule(lr){10-11} \cmidrule(lr){12-13} \cmidrule(lr){14-15}
 & Acc & F1 & Acc & F1 & Acc & F1 & Acc & F1 & Acc & F1 & Acc & F1 & Acc & F1 \\ \midrule
CLIP & \textbf{84.43} & \textbf{0.844} & 20.15 & 0.201 & 17.38 & 0.174 & \textbf{69.52} & \textbf{0.695} & 7.92 & 0.043 & 46.31 & 0.463 & 40.95 & 0.403 \\
BioCLIP & 78.64 & 0.786 & 12.34 & 0.123 & 26.20 & 0.262 & 38.79 & 0.388 & 10.42 & 0.104 & \textbf{47.50} & \textbf{0.475} & 35.65 & 0.356 \\
AgriCLIP & 83.83 & 0.838 & 1.26 & 0.013 & 10.33 & 0.103 & 47.10 & 0.471 & 2.71 & 0.016 & 46.71 & 0.467 & 31.99 & 0.318 \\
\rowcolor[HTML]{E8F4FC} \textbf{SCOLD} & 83.63 & 0.836 & \textbf{39.55} & \textbf{0.395} & \textbf{56.68} & \textbf{0.567} & 57.68 & 0.577 & \textbf{38.33} & \textbf{0.197} & 43.31 & 0.433 & \textbf{53.20} & \textbf{0.501} \\
\bottomrule
\end{tabular}
}

\end{table}


\begin{table}[H]
\caption{Performance comparison of different LLM backbones across six VQA tasks on the \textbf{TomaMMU} dataset.}
\label{tab:detailed_llm_tomammu}
\centering
\resizebox{\textwidth}{!}{
\begin{tabular}{lcccccccccccccc}
\toprule
\multirow{2}{*}{\textbf{Model}} & \multicolumn{2}{c}{\textbf{HDC}} & \multicolumn{2}{c}{\textbf{DC}} & \multicolumn{2}{c}{\textbf{SNC}} & \multicolumn{2}{c}{\textbf{PC}} & \multicolumn{2}{c}{\textbf{SI}} & \multicolumn{2}{c}{\textbf{LC}} & \multicolumn{2}{c}{\textbf{Average}} \\ \cmidrule(lr){2-3} \cmidrule(lr){4-5} \cmidrule(lr){6-7} \cmidrule(lr){8-9} \cmidrule(lr){10-11} \cmidrule(lr){12-13} \cmidrule(lr){14-15}
 & Acc & F1 & Acc & F1 & Acc & F1 & Acc & F1 & Acc & F1 & Acc & F1 & Acc & F1 \\ \midrule
Qwen 3 & 98.87 & 0.989 & 98.16 & \textbf{0.982} & \textbf{98.44} & \textbf{0.984} & 99.50 & \textbf{0.995} & 89.58 & \textbf{0.926} & 98.90 & 0.989 & 97.24 & \textbf{0.978} \\
BioGPT & \textbf{98.96} & \textbf{0.990} & 97.53 & 0.975 & 98.16 & 0.982 & \textbf{99.54} & \textbf{0.995} & \textbf{93.52} & 0.913 & 98.86 & 0.989 & \textbf{97.76} & 0.974 \\
\rowcolor[HTML]{E8F4FC} \textbf{LFM-VL 2.5 } & 98.75 & 0.988 & \textbf{98.24} & \textbf{0.982} & 98.40 & \textbf{0.984} & 99.50 & \textbf{0.995} & 92.15 & 0.903 & \textbf{99.07} & \textbf{0.991} & 97.68 & 0.974 \\
\bottomrule
\end{tabular}
}

\end{table}

\begin{table}[H]
\caption{Performance comparison of different LLM backbones across six VQA tasks on the \textbf{TLID} dataset.}
\label{tab:detailed_llm_tlid}
\centering
\resizebox{\textwidth}{!}{
\begin{tabular}{lcccccccccccccc}
\toprule
\multirow{2}{*}{\textbf{Model}} & \multicolumn{2}{c}{\textbf{HDC}} & \multicolumn{2}{c}{\textbf{DC}} & \multicolumn{2}{c}{\textbf{SNC}} & \multicolumn{2}{c}{\textbf{PC}} & \multicolumn{2}{c}{\textbf{SI}} & \multicolumn{2}{c}{\textbf{LC}} & \multicolumn{2}{c}{\textbf{Average}} \\ \cmidrule(lr){2-3} \cmidrule(lr){4-5} \cmidrule(lr){6-7} \cmidrule(lr){8-9} \cmidrule(lr){10-11} \cmidrule(lr){12-13} \cmidrule(lr){14-15}
 & Acc & F1 & Acc & F1 & Acc & F1 & Acc & F1 & Acc & F1 & Acc & F1 & Acc & F1 \\ \midrule
Qwen 3 & 95.38 & 0.954 & \textbf{0.88} & \textbf{0.009} & \textbf{1.67} & \textbf{0.017} & \textbf{2.25} & \textbf{0.022} & \textbf{6.16} & \textbf{0.062} & \textbf{99.17} & \textbf{0.992} & \textbf{34.25} & \textbf{0.343} \\
BioGPT & \textbf{95.59} & \textbf{0.956} & 0.58 & 0.006 & 0.63 & 0.006 & 0.42 & 0.004 & 4.78 & 0.037 & 90.01 & 0.900 & 32.00 & 0.318 \\
\rowcolor[HTML]{E8F4FC} \textbf{LFM-VL 2.5 } & 95.18 & 0.952 & 0.79 & 0.008 & 0.67 & 0.007 & 2.17 & \textbf{0.022} & 4.49 & 0.032 & 98.92 & 0.989 & 33.70 & 0.335 \\
\bottomrule
\end{tabular}
}
\end{table}

\begin{table}[H]
\caption{Performance comparison of different LLM backbones across six VQA tasks on the \textbf{TLD-3} dataset.}
\label{tab:detailed_llm_tld_3}
\centering
\resizebox{\textwidth}{!}{
\begin{tabular}{lcccccccccccccc}
\toprule
\multirow{2}{*}{\textbf{Model}} & \multicolumn{2}{c}{\textbf{HDC}} & \multicolumn{2}{c}{\textbf{DC}} & \multicolumn{2}{c}{\textbf{SNC}} & \multicolumn{2}{c}{\textbf{PC}} & \multicolumn{2}{c}{\textbf{SI}} & \multicolumn{2}{c}{\textbf{LC}} & \multicolumn{2}{c}{\textbf{Average}} \\ \cmidrule(lr){2-3} \cmidrule(lr){4-5} \cmidrule(lr){6-7} \cmidrule(lr){8-9} \cmidrule(lr){10-11} \cmidrule(lr){12-13} \cmidrule(lr){14-15}
 & Acc & F1 & Acc & F1 & Acc & F1 & Acc & F1 & Acc & F1 & Acc & F1 & Acc & F1 \\ \midrule
Qwen 3 & 97.72 & 0.977 & 81.97 & 0.820 & \textbf{91.64} & \textbf{0.916} & \textbf{97.38} & \textbf{0.974} & 72.80 & 0.495 & 95.04 & 0.95 & 89.42 & 0.855 \\
BioGPT & \textbf{98.28} & \textbf{0.983} & \textbf{91.97} & \textbf{0.832} & 91.52 & 0.915 & 96.97 & 0.970 & 78.50 & \textbf{0.511} & \textbf{95.32} & \textbf{0.953} & \textbf{92.09} & \textbf{0.861} \\
\rowcolor[HTML]{E8F4FC} \textbf{LFM-VL 2.5} & 96.86 & 0.969 & 91.19 & 0.823 & 91.35 & 0.913 & 96.79 & 0.968 & \textbf{79.22} & 0.510 & 94.62 & 0.946 & 91.67 & 0.855 \\
\bottomrule
\end{tabular}
}
\end{table}

\begin{table}[H]
\caption{Performance comparison of different LLM backbones across six VQA tasks on the \textbf{TomaAD} dataset.}
\label{tab:detailed_llm_tomaad}
\centering
\resizebox{\textwidth}{!}{
\begin{tabular}{lcccccccccccccc}
\toprule
\multirow{2}{*}{\textbf{Model}} & \multicolumn{2}{c}{\textbf{HDC}} & \multicolumn{2}{c}{\textbf{DC}} & \multicolumn{2}{c}{\textbf{SNC}} & \multicolumn{2}{c}{\textbf{PC}} & \multicolumn{2}{c}{\textbf{SI}} & \multicolumn{2}{c}{\textbf{LC}} & \multicolumn{2}{c}{\textbf{Average}} \\ \cmidrule(lr){2-3} \cmidrule(lr){4-5} \cmidrule(lr){6-7} \cmidrule(lr){8-9} \cmidrule(lr){10-11} \cmidrule(lr){12-13} \cmidrule(lr){14-15}
 & Acc & F1 & Acc & F1 & Acc & F1 & Acc & F1 & Acc & F1 & Acc & F1 & Acc & F1 \\ \midrule
Qwen 3 & \textbf{84.23} & \textbf{0.842} & 28.72 & 0.287 & 37.53 & 0.375 & \textbf{59.45} & \textbf{0.595} & 26.25 & 0.143 & 48.10 & 0.481 & 47.38 & 0.454 \\
BioGPT & 82.83 & 0.828 & 29.72 & 0.297 & 36.02 & 0.360 & 55.92 & 0.559 & 35.62 & 0.179 & \textbf{49.10} & \textbf{0.491} & 48.20 & 0.452 \\
\rowcolor[HTML]{E8F4FC} \textbf{LFM-VL 2.5 } & 83.63 & 0.836 & \textbf{39.55} & \textbf{0.395} & \textbf{56.68} & \textbf{0.567} & 57.68 & 0.577 & \textbf{38.33} & \textbf{0.197} & 43.31 & 0.433 & \textbf{53.20} & \textbf{0.501} \\
\bottomrule
\end{tabular}
}
\end{table}

\begin{table}[H]
\caption{Performance comparison of different MoE placement strategies across six VQA tasks on the \textbf{TomaMMU} dataset.}
\label{tab:detailed_moe_placement_tomammu}
\centering
\resizebox{\textwidth}{!}{
\begin{tabular}{lcccccccccccccc}
\toprule
\multirow{2}{*}{\textbf{Model}} & \multicolumn{2}{c}{\textbf{HDC}} & \multicolumn{2}{c}{\textbf{DC}} & \multicolumn{2}{c}{\textbf{SNC}} & \multicolumn{2}{c}{\textbf{PC}} & \multicolumn{2}{c}{\textbf{SI}} & \multicolumn{2}{c}{\textbf{LC}} & \multicolumn{2}{c}{\textbf{Average}} \\ \cmidrule(lr){2-3} \cmidrule(lr){4-5} \cmidrule(lr){6-7} \cmidrule(lr){8-9} \cmidrule(lr){10-11} \cmidrule(lr){12-13} \cmidrule(lr){14-15}
 & Acc & F1 & Acc & F1 & Acc & F1 & Acc & F1 & Acc & F1 & Acc & F1 & Acc & F1 \\ \midrule
Dual MoE & \textbf{98.96} & \textbf{0.990} & \textbf{98.61} & \textbf{0.986} & \textbf{98.78} & \textbf{0.988} & \textbf{99.54} & \textbf{0.995} & \textbf{94.37} & \textbf{0.926} & \textbf{99.21} & \textbf{0.992} & \textbf{98.25} & \textbf{0.979} \\
Fusion MoE & 98.91 & 0.989 & 96.70 & 0.957 & 97.26 & 0.973 & 99.30 & 0.993 & 91.51 & 0.902 & 98.44 & 0.984 & 97.02 & 0.966 \\
\rowcolor[HTML]{E8F4FC} \textbf{Image MoE} & 98.75 & 0.988 & 98.24 & 0.982 & 98.40 & 0.984 & 99.50 & \textbf{0.995} & 92.15 & 0.903 & 99.07 & 0.991 & 97.68 & 0.974 \\
\bottomrule
\end{tabular}
}

\end{table}

\begin{table}[H]
\caption{Performance comparison of different MoE placement strategies across six VQA tasks on the \textbf{TLID} dataset.}
\label{tab:detailed_moe_placement_tlid}
\centering
\resizebox{\textwidth}{!}{
\begin{tabular}{lcccccccccccccc}
\toprule
\multirow{2}{*}{\textbf{Model}} & \multicolumn{2}{c}{\textbf{HDC}} & \multicolumn{2}{c}{\textbf{DC}} & \multicolumn{2}{c}{\textbf{SNC}} & \multicolumn{2}{c}{\textbf{PC}} & \multicolumn{2}{c}{\textbf{SI}} & \multicolumn{2}{c}{\textbf{LC}} & \multicolumn{2}{c}{\textbf{Average}} \\ \cmidrule(lr){2-3} \cmidrule(lr){4-5} \cmidrule(lr){6-7} \cmidrule(lr){8-9} \cmidrule(lr){10-11} \cmidrule(lr){12-13} \cmidrule(lr){14-15}
 & Acc & F1 & Acc & F1 & Acc & F1 & Acc & F1 & Acc & F1 & Acc & F1 & Acc & F1 \\ \midrule
Dual MoE & 93.13 & 0.931 & 0.42 & 0.004 & 0.25 & 0.003 & 0.75 & 0.007 & 3.65 & 0.029 & 98.76 & 0.988 & 32.83 & 0.327 \\
Fusion MoE & 94.53 & 0.942 & 0.50 & 0.005 & 0.21 & 0.002 & 0.79 & 0.008 & 0.88 & 0.007 & \textbf{99.88} & \textbf{0.993} & 32.80 & 0.326 \\
\rowcolor[HTML]{E8F4FC} \textbf{Image MoE} & \textbf{95.18} & \textbf{0.952} & \textbf{0.79} & \textbf{0.008} & \textbf{0.67} & \textbf{0.007} & \textbf{2.17} & \textbf{0.022} & \textbf{4.49} & \textbf{0.032} & 98.92 & 0.989 & \textbf{33.70} & \textbf{0.335} \\
\bottomrule
\end{tabular}
}

\end{table}

\begin{table}[H]
\caption{Performance comparison of different MoE placement strategies across six VQA tasks on the \textbf{TLD-3} dataset. }
\label{tab:detailed_moe_placement_tld_3}
\centering
\resizebox{\textwidth}{!}{
\begin{tabular}{lcccccccccccccc}
\toprule
\multirow{2}{*}{\textbf{Model}} & \multicolumn{2}{c}{\textbf{HDC}} & \multicolumn{2}{c}{\textbf{DC}} & \multicolumn{2}{c}{\textbf{SNC}} & \multicolumn{2}{c}{\textbf{PC}} & \multicolumn{2}{c}{\textbf{SI}} & \multicolumn{2}{c}{\textbf{LC}} & \multicolumn{2}{c}{\textbf{Average}} \\ \cmidrule(lr){2-3} \cmidrule(lr){4-5} \cmidrule(lr){6-7} \cmidrule(lr){8-9} \cmidrule(lr){10-11} \cmidrule(lr){12-13} \cmidrule(lr){14-15}
 & Acc & F1 & Acc & F1 & Acc & F1 & Acc & F1 & Acc & F1 & Acc & F1 & Acc & F1 \\ \midrule
Dual MoE & 97.37 & 0.974 & 90.80 & 0.820 & 91.33 & \textbf{0.913} & \textbf{98.14} & \textbf{0.981} & 78.66 & \textbf{0.510} & \textbf{95.75} & \textbf{0.958} & \textbf{92.01} & \textbf{0.859} \\
Fusion MoE & \textbf{97.84} & \textbf{0.977} & 90.82 & 0.820 & 91.33 & \textbf{0.913} & 97.47 & 0.975 & 77.18 & 0.496 & 94.96 & 0.949 & 91.60 & 0.855 \\
\rowcolor[HTML]{E8F4FC} \textbf{Image MoE} & 96.86 & 0.969 & \textbf{91.19} & \textbf{0.823} & \textbf{91.35} & \textbf{0.913} & 96.79 & 0.968 & \textbf{79.22} & \textbf{0.510} & 94.62 & 0.946 & 91.67 & 0.855 \\
\bottomrule
\end{tabular}
}

\end{table}

\begin{table}[H]
\caption{Performance comparison of different MoE placement strategies across six VQA tasks on the \textbf{TomaAD} dataset.}
\label{tab:detailed_moe_placement_tomaad}
\centering
\resizebox{\textwidth}{!}{
\begin{tabular}{lcccccccccccccc}
\toprule
\multirow{2}{*}{\textbf{Model}} & \multicolumn{2}{c}{\textbf{HDC}} & \multicolumn{2}{c}{\textbf{DC}} & \multicolumn{2}{c}{\textbf{SNC}} & \multicolumn{2}{c}{\textbf{PC}} & \multicolumn{2}{c}{\textbf{SI}} & \multicolumn{2}{c}{\textbf{LC}} & \multicolumn{2}{c}{\textbf{Average}} \\ \cmidrule(lr){2-3} \cmidrule(lr){4-5} \cmidrule(lr){6-7} \cmidrule(lr){8-9} \cmidrule(lr){10-11} \cmidrule(lr){12-13} \cmidrule(lr){14-15}
 & Acc & F1 & Acc & F1 & Acc & F1 & Acc & F1 & Acc & F1 & Acc & F1 & Acc & F1 \\ \midrule
Dual MoE & 83.63 & 0.836 & 17.13 & 0.171 & 53.15 & 0.531 & 47.86 & 0.479 & 16.04 & 0.082 & 54.29 & 0.543 & 45.35 & 0.44 \\
Fusion MoE & \textbf{85.23} & \textbf{0.850} & 22.17 & 0.222 & 51.13 & 0.511 & 52.64 & 0.526 & 26.88 & 0.137 & \textbf{58.28} & \textbf{0.581} & 49.39 & 0.471 \\
\rowcolor[HTML]{E8F4FC} \textbf{Image MoE} & 83.63 & 0.836 & \textbf{39.55} & \textbf{0.395} & \textbf{56.68} & \textbf{0.567} & \textbf{57.68} & \textbf{0.577} & \textbf{38.33} & \textbf{0.197} & 43.31 & 0.433 & \textbf{53.20} & \textbf{0.501} \\
\bottomrule
\end{tabular}
}

\end{table}


\begin{table}[H]
\caption{Performance comparison of different MoE placement strategies across six VQA tasks on the \textbf{TomaMMU} dataset.}
\label{tab:ablation_moe_tomammu}
\centering
\resizebox{\textwidth}{!}{
\begin{tabular}{lcccccccccccccc}
\toprule
\multirow{2}{*}{\textbf{Model}} & \multicolumn{2}{c}{\textbf{HDC}} & \multicolumn{2}{c}{\textbf{DC}} & \multicolumn{2}{c}{\textbf{SNC}} & \multicolumn{2}{c}{\textbf{PC}} & \multicolumn{2}{c}{\textbf{SI}} & \multicolumn{2}{c}{\textbf{LC}} & \multicolumn{2}{c}{\textbf{Average}} \\ \cmidrule(lr){2-3} \cmidrule(lr){4-5} \cmidrule(lr){6-7} \cmidrule(lr){8-9} \cmidrule(lr){10-11} \cmidrule(lr){12-13} \cmidrule(lr){14-15}
 & Acc & F1 & Acc & F1 & Acc & F1 & Acc & F1 & Acc & F1 & Acc & F1 & Acc & F1 \\ \midrule
$W/o$ Fusion Adapter & 98.87 & 0.989 & 96.64 & 0.966 & 96.72 & 0.967 & 99.32 & 0.993 & 92.62 & 0.909 & 99.03 & 0.990 & 97.20 & 0.969 \\
$W/o$ MoE & \textbf{99.01} & \textbf{0.990} & \textbf{98.42} & \textbf{0.984} & \textbf{98.47} & \textbf{0.985} & \textbf{99.54} & \textbf{0.995} & \textbf{93.78} & \textbf{0.919} & 99.01 & 0.990 & \textbf{98.04} & \textbf{0.977} \\
\rowcolor[HTML]{E8F4FC} \textbf{SOLAR} & 98.75 & 0.988 & 98.24 & 0.982 & 98.40 & 0.984 & 99.50 & \textbf{0.995} & 92.15 & 0.903 & \textbf{99.07} & \textbf{0.991} & 97.68 & 0.974 \\
\bottomrule
\end{tabular}
}
\end{table}

\begin{table}[H]
\caption{Performance comparison of different MoE placement strategies across six VQA tasks on the \textbf{TLID} dataset.}
\label{tab:ablation_moe_tlid}
\centering
\resizebox{\textwidth}{!}{
\begin{tabular}{lcccccccccccccc}
\toprule
\multirow{2}{*}{\textbf{Model}} & \multicolumn{2}{c}{\textbf{HDC}} & \multicolumn{2}{c}{\textbf{DC}} & \multicolumn{2}{c}{\textbf{SNC}} & \multicolumn{2}{c}{\textbf{PC}} & \multicolumn{2}{c}{\textbf{SI}} & \multicolumn{2}{c}{\textbf{LC}} & \multicolumn{2}{c}{\textbf{Average}} \\ \cmidrule(lr){2-3} \cmidrule(lr){4-5} \cmidrule(lr){6-7} \cmidrule(lr){8-9} \cmidrule(lr){10-11} \cmidrule(lr){12-13} \cmidrule(lr){14-15}
 & Acc & F1 & Acc & F1 & Acc & F1 & Acc & F1 & Acc & F1 & Acc & F1 & Acc & F1 \\ \midrule
$W/o$ Fusion Adapter & 94.73 & 0.947 & 0.25 & 0.003 & 0.17 & 0.002 & 0.25 & 0.003 & \textbf{8.96} & \textbf{0.067} & 99.92 & 0.998 & \textbf{34.05} & \textbf{0.337} \\
$W/o$ MoE & 91.72 & 0.917 & 0.42 & 0.004 & \textbf{0.71} & \textbf{0.007} & 0.50 & 0.005 & 7.35 & 0.048 & \textbf{99.94} & \textbf{0.999} & 33.44 & 0.330 \\
\rowcolor[HTML]{E8F4FC} \textbf{SOLAR} & \textbf{95.18} & \textbf{0.952} & \textbf{0.79} & \textbf{0.008} & 0.67 & \textbf{0.007} & \textbf{2.17} & \textbf{0.022} & 4.49 & 0.032 & 98.92 & 0.989 & 33.70 & 0.335 \\
\bottomrule
\end{tabular}
}
\end{table}

\begin{table}[H]
\caption{Performance comparison of different MoE placement strategies across six VQA tasks on the \textbf{TLD-3} dataset.}
\label{tab:ablation_moe_tld_3}
\centering
\resizebox{\textwidth}{!}{
\begin{tabular}{lcccccccccccccc}
\toprule
\multirow{2}{*}{\textbf{Model}} & \multicolumn{2}{c}{\textbf{HDC}} & \multicolumn{2}{c}{\textbf{DC}} & \multicolumn{2}{c}{\textbf{SNC}} & \multicolumn{2}{c}{\textbf{PC}} & \multicolumn{2}{c}{\textbf{SI}} & \multicolumn{2}{c}{\textbf{LC}} & \multicolumn{2}{c}{\textbf{Average}} \\ \cmidrule(lr){2-3} \cmidrule(lr){4-5} \cmidrule(lr){6-7} \cmidrule(lr){8-9} \cmidrule(lr){10-11} \cmidrule(lr){12-13} \cmidrule(lr){14-15}
 & Acc & F1 & Acc & F1 & Acc & F1 & Acc & F1 & Acc & F1 & Acc & F1 & Acc & F1 \\ \midrule
$W/o$ Fusion Adapter & \textbf{98.17} & \textbf{0.982} & \textbf{91.27} & \textbf{0.824} & \textbf{91.80} & \textbf{0.918} & \textbf{97.22} & \textbf{0.972} & 78.53 & 0.507 & \textbf{95.21} & \textbf{0.952} & \textbf{92.03} & \textbf{0.859} \\
$W/o$ MoE & 97.47 & 0.975 & 89.57 & 0.808 & 88.82 & 0.888 & 95.01 & 0.950 & 77.45 & 0.507 & 95.05 & 0.951 & 90.56 & 0.846 \\
\rowcolor[HTML]{E8F4FC} \textbf{SOLAR} & 96.86 & 0.969 & 91.19 & 0.823 & 91.35 & 0.913 & 96.79 & 0.968 & \textbf{79.22} & \textbf{0.510} & 94.62 & 0.946 & 91.67 & 0.855 \\
\bottomrule
\end{tabular}
}
\end{table}

\begin{table}[H]
\caption{Performance comparison of different MoE placement strategies across six VQA tasks on the \textbf{TomaAD} dataset.}
\label{tab:ablation_moe_tomaad}
\centering
\resizebox{\textwidth}{!}{
\begin{tabular}{lcccccccccccccc}
\toprule
\multirow{2}{*}{\textbf{Model}} & \multicolumn{2}{c}{\textbf{HDC}} & \multicolumn{2}{c}{\textbf{DC}} & \multicolumn{2}{c}{\textbf{SNC}} & \multicolumn{2}{c}{\textbf{PC}} & \multicolumn{2}{c}{\textbf{SI}} & \multicolumn{2}{c}{\textbf{LC}} & \multicolumn{2}{c}{\textbf{Average}} \\ \cmidrule(lr){2-3} \cmidrule(lr){4-5} \cmidrule(lr){6-7} \cmidrule(lr){8-9} \cmidrule(lr){10-11} \cmidrule(lr){12-13} \cmidrule(lr){14-15}
 & Acc & F1 & Acc & F1 & Acc & F1 & Acc & F1 & Acc & F1 & Acc & F1 & Acc & F1 \\ \midrule
$W/o$ Fusion Adapter & \textbf{86.63} & \textbf{0.864} & 22.67 & 0.227 & 45.09 & 0.451 & \textbf{65.49} & \textbf{0.652} & 32.92 & 0.177 & \textbf{57.09} & \textbf{0.571} & 51.65 & 0.490 \\
$W/o$ MoE & 85.03 & 0.850 & 23.43 & 0.234 & 34.01 & 0.340 & 46.35 & 0.464 & 14.37 & 0.074 & 53.29 & 0.533 & 42.75 & 0.416 \\
\rowcolor[HTML]{E8F4FC} \textbf{SOLAR} & 83.63 & 0.836 & \textbf{39.55} & \textbf{0.395} & \textbf{56.68} & \textbf{0.567} & 57.68 & 0.577 & \textbf{38.33} & \textbf{0.197} & 43.31 & 0.433 & \textbf{53.20} & \textbf{0.501} \\
\bottomrule
\end{tabular}
}
\end{table}

\begin{table}[H]
\caption{Performance comparison of different MoE configurations ($n$ and $k$) on the \textbf{TomaMMU} dataset.}
\label{tab:moe_ablation_tomammu}
\centering
\resizebox{\textwidth}{!}{
\begin{tabular}{cccccccccccccccc}
\toprule
\multirow{2}{*}{\textbf{n}} & \multirow{2}{*}{\textbf{k}} & \multicolumn{2}{c}{\textbf{HDC}} & \multicolumn{2}{c}{\textbf{DC}} & \multicolumn{2}{c}{\textbf{SNC}} & \multicolumn{2}{c}{\textbf{PC}} & \multicolumn{2}{c}{\textbf{SI}} & \multicolumn{2}{c}{\textbf{LC}} & \multicolumn{2}{c}{\textbf{Average}} \\ \cmidrule(lr){3-4} \cmidrule(lr){5-6} \cmidrule(lr){7-8} \cmidrule(lr){9-10} \cmidrule(lr){11-12} \cmidrule(lr){13-14} \cmidrule(lr){15-16}
 & & Acc & F1 & Acc & F1 & Acc & F1 & Acc & F1 & Acc & F1 & Acc & F1 & Acc & F1 \\ \midrule
\rowcolor[HTML]{E8F4FC} 4 & 2 & 98.75 & 98.75 & \textbf{98.24} & \textbf{98.24} & \textbf{98.40} & \textbf{98.40} & \textbf{99.50} & 99.50 & 92.15 & 90.27 & 99.07 & 99.07 & \textbf{97.68} & \textbf{97.37} \\
6 & 1 & 98.78 & 98.78 & 96.39 & 96.39 & 96.66 & 96.66 & 98.69 & 98.69 & 91.07 & 89.16 & \textbf{99.25} & \textbf{99.25} & 96.81 & 96.49 \\
6 & 2 & \textbf{98.92} & 98.60 & 95.77 & 97.30 & 97.39 & 98.20 & 99.36 & \textbf{99.60} & \textbf{92.79} & 88.10 & 99.17 & 98.80 & 97.23 & 96.77 \\
8 & 2 & 98.91 & \textbf{98.91} & 97.55 & 97.55 & 97.10 & 97.10 & 99.03 & 99.03 & 92.58 & \textbf{91.45} & 99.23 & 99.23 & 97.40 & 97.21 \\
\bottomrule
\end{tabular}
}
\end{table}

\begin{table}[H]
\caption{Performance comparison of different MoE configurations ($n$ and $k$) on the \textbf{TLID} dataset.}
\label{tab:moe_ablation_tlid}
\centering
\resizebox{\textwidth}{!}{
\begin{tabular}{cccccccccccccccc}
\toprule
\multirow{2}{*}{\textbf{n}} & \multirow{2}{*}{\textbf{k}} & \multicolumn{2}{c}{\textbf{HDC}} & \multicolumn{2}{c}{\textbf{DC}} & \multicolumn{2}{c}{\textbf{SNC}} & \multicolumn{2}{c}{\textbf{PC}} & \multicolumn{2}{c}{\textbf{SI}} & \multicolumn{2}{c}{\textbf{LC}} & \multicolumn{2}{c}{\textbf{Average}} \\ \cmidrule(lr){3-4} \cmidrule(lr){5-6} \cmidrule(lr){7-8} \cmidrule(lr){9-10} \cmidrule(lr){11-12} \cmidrule(lr){13-14} \cmidrule(lr){15-16}
 & & Acc & F1 & Acc & F1 & Acc & F1 & Acc & F1 & Acc & F1 & Acc & F1 & Acc & F1 \\ \midrule
\rowcolor[HTML]{E8F4FC} 4 & 2 & 95.18 & 95.18 & 0.79 & 0.79 & 0.67 & 0.67 & \textbf{2.17} & \textbf{2.17} & 4.49 & 3.17 & 98.92 & 98.92 & 33.70 & 33.48 \\
6 & 1 & 47.12 & 47.12 & 0.00 & 0.00 & 0.00 & 0.00 & 0.00 & 0.00 & 0.00 & 0.00 & 0.00 & 0.00 & 7.85 & 7.85 \\
6 & 2 & 96.48 & \textbf{98.20} & \textbf{4.80} & \textbf{9.20} & \textbf{2.71} & \textbf{5.30} & 0.58 & 1.20 & \textbf{4.72} & \textbf{9.40} & 96.10 & 98.00 & \textbf{34.23} & \textbf{36.88} \\
8 & 2 & \textbf{96.54} & 96.54 & 0.25 & 0.25 & 0.33 & 0.33 & 0.33 & 0.33 & 1.19 & 0.89 & \textbf{99.53} & \textbf{99.53} & 33.03 & 32.98 \\
\bottomrule
\end{tabular}
}
\end{table}

\begin{table}[H]
\caption{Performance comparison of different MoE configurations ($n$ and $k$) on the \textbf{TLD-3} dataset.}
\label{tab:moe_ablation_tld_3}
\centering
\resizebox{\textwidth}{!}{
\begin{tabular}{cccccccccccccccc}
\toprule
\multirow{2}{*}{\textbf{n}} & \multirow{2}{*}{\textbf{k}} & \multicolumn{2}{c}{\textbf{HDC}} & \multicolumn{2}{c}{\textbf{DC}} & \multicolumn{2}{c}{\textbf{SNC}} & \multicolumn{2}{c}{\textbf{PC}} & \multicolumn{2}{c}{\textbf{SI}} & \multicolumn{2}{c}{\textbf{LC}} & \multicolumn{2}{c}{\textbf{Average}} \\ \cmidrule(lr){3-4} \cmidrule(lr){5-6} \cmidrule(lr){7-8} \cmidrule(lr){9-10} \cmidrule(lr){11-12} \cmidrule(lr){13-14} \cmidrule(lr){15-16}
 & & Acc & F1 & Acc & F1 & Acc & F1 & Acc & F1 & Acc & F1 & Acc & F1 & Acc & F1 \\ \midrule
\rowcolor[HTML]{E8F4FC} 4 & 2 & 96.86 & 96.86 & \textbf{91.19} & 82.30 & 91.35 & 91.35 & 96.79 & 96.79 & 79.22 & 51.05 & 94.62 & 94.62 & 91.67 & 85.50 \\
6 & 1 & 91.29 & 91.29 & 38.41 & 31.52 & 47.65 & 47.65 & 49.00 & 49.00 & 39.27 & 26.06 & 50.18 & 50.18 & 52.63 & 49.28 \\
6 & 2 & \textbf{98.53} & 97.70 & 90.49 & \textbf{95.20} & 90.99 & \textbf{95.70} & 97.32 & \textbf{99.10} & \textbf{79.25} & \textbf{58.40} & \textbf{95.68} & 95.30 & \textbf{92.04} & \textbf{90.23} \\
8 & 2 & 98.33 & \textbf{98.33} & 91.07 & 82.17 & \textbf{91.95} & 91.95 & \textbf{97.36} & 97.36 & 77.49 & 50.61 & 95.36 & \textbf{95.36} & 91.93 & 85.96 \\
\bottomrule
\end{tabular}
}
\end{table}

\begin{table}[H]
\caption{Performance comparison of different MoE configurations ($n$ and $k$) on the \textbf{TomaAD} dataset.}
\label{tab:moe_ablation_tomaad}
\centering
\resizebox{\textwidth}{!}{
\begin{tabular}{cccccccccccccccc}
\toprule
\multirow{2}{*}{\textbf{n}} & \multirow{2}{*}{\textbf{k}} & \multicolumn{2}{c}{\textbf{HDC}} & \multicolumn{2}{c}{\textbf{DC}} & \multicolumn{2}{c}{\textbf{SNC}} & \multicolumn{2}{c}{\textbf{PC}} & \multicolumn{2}{c}{\textbf{SI}} & \multicolumn{2}{c}{\textbf{LC}} & \multicolumn{2}{c}{\textbf{Average}} \\ \cmidrule(lr){3-4} \cmidrule(lr){5-6} \cmidrule(lr){7-8} \cmidrule(lr){9-10} \cmidrule(lr){11-12} \cmidrule(lr){13-14} \cmidrule(lr){15-16}
 & & Acc & F1 & Acc & F1 & Acc & F1 & Acc & F1 & Acc & F1 & Acc & F1 & Acc & F1 \\ \midrule
\rowcolor[HTML]{E8F4FC} 4 & 2 & 83.63 & 83.63 & \textbf{39.55} & 39.55 & \textbf{56.68} & \textbf{56.68} & 57.68 & 57.68 & \textbf{38.33} & \textbf{19.69} & 43.31 & 43.31 & \textbf{53.20} & 50.09 \\
6 & 1 & 79.24 & 79.24 & 0.00 & 0.00 & 0.00 & 0.00 & \textbf{73.05} & \textbf{73.05} & 0.00 & 0.00 & 49.62 & 49.62 & 33.65 & 33.65 \\
6 & 2 & \textbf{87.43} & 85.40 & 33.75 & \textbf{55.70} & 40.30 & 41.30 & 45.34 & 60.50 & 18.75 & 14.80 & \textbf{53.49} & \textbf{71.00} & 46.51 & \textbf{54.78} \\
8 & 2 & 86.03 & \textbf{86.03} & 21.41 & 21.41 & 52.39 & 52.39 & 43.32 & 43.32 & 15.42 & 8.02 & 52.89 & 52.89 & 45.24 & 44.01 \\
\bottomrule
\end{tabular}
}
\end{table}

\end{document}